\documentclass[sigconf]{acmart}

\makeatletter
\@ifundefined{Bbbk}{}{%
  
}
\makeatother
\usepackage{amssymb}
\usepackage{bm}  
\usepackage{etoc}
\usepackage{microtype}
\usepackage{graphicx}
\usepackage{subfig}
\usepackage{float} 
\usepackage{algorithm}
\usepackage{enumitem}   
\usepackage{tabularx,booktabs}
\usepackage{xspace}
\makeatletter
\@ifundefined{theHalgorithm}
  {}
  {}
\makeatother
\usepackage{hyperref}
\usepackage{algorithmic}
\usepackage{url}

\usepackage{tabularx,array,xcolor,makecell}
\newcommand{\hlred}[1]{\colorbox{red!15}{\strut #1}}

\newcolumntype{Y}{>{\centering\arraybackslash}X}
\newcolumntype{Z}{>{\Centering\arraybackslash}X} 

\usepackage{framed}

\usepackage{amsthm}
\usepackage{amssymb}
\usepackage{amsfonts}
\usepackage{mathtools}

\usepackage{hyperref}
\usepackage[nameinlink,capitalize]{cleveref}
\hypersetup{colorlinks=true,linkcolor=blue,citecolor=blue,urlcolor=blue,pdfborder={0 0 0}}
\usepackage[normalem]{ulem}

\usepackage[utf8]{inputenc}
\usepackage[T1]{fontenc}
\usepackage{url}
\usepackage{booktabs}
\usepackage{nicefrac}
\usepackage{microtype}
\usepackage{multirow}
\usepackage{lscape}

\usepackage{graphicx}
\usepackage{enumitem}

\newtheorem{theorem}{Theorem}[section]
\newtheorem{proposition}{Proposition}
\newtheorem{corollary}{Corollary}[section]

\theoremstyle{definition}

\theoremstyle{remark}

\newcommand{\bc}{\begin{center}}
\newcommand{\ec}{\end{center}}

\newcommand{\bdm}{\begin{displaymath}}
\newcommand{\edm}{\end{displaymath}}

\newcommand{\beq}{\begin{equation}}
\newcommand{\eeq}{\end{equation}}

\newcommand{\bfl}{\begin{flushleft}}
\newcommand{\efl}{\end{flushleft}}

\newcommand{\bt}{\begin{tabbing}}
\newcommand{\et}{\end{tabbing}}

\newcommand{\beqn}{\begin{align}}
\newcommand{\eeqn}{\end{align}}

\newcommand{\beqs}{\begin{align*}}
\newcommand{\eeqs}{\end{align*}}

\renewcommand{\algorithmicrequire}{\textbf{Input:}}

\usepackage{framed}

\usepackage{siunitx}
\theoremstyle{plain}
\usepackage{tikz}
\usetikzlibrary{shadows}
\definecolor{mine}{RGB}{205, 232, 248}%
\definecolor{minedark}{RGB}{160, 190, 210}%
\definecolor{LegendBlue}{HTML}{1F77B4}

\newcommand{\llm}[1][]{%
  \if\relax\detokenize{#1}\relax
    \textbf{LM}\xspace
  \else
    \ensuremath{\mathbf{LM}_{\mathrm{#1}}}\xspace
  \fi
}

\newcommand{\llmforward}{\llm[forward]}
\newcommand{\llmbackward}{\llm[backward]}

\newcommand{\dataset}{D}
\newcommand{\metric}{\texttt{Perf}}
\newcommand{\dataDist}{\mathcal{D}}
\newcommand{\vocab}{\mathcal{V}}
\newcommand{\Update}{\texttt{Update}}

\newcommand{\prompt}{\pi}
\newcommand{\tokenblock}{\text{TokenBlock}}
\newcommand{\textgrad}{\text{TextGrad}}
\newcommand{\adalflow}{\text{AdalFlow}}
\newcommand{\tg}{\text{TG}}
\newcommand{\alg}{\text{TSGD-M}}
\newcommand{\dspy}{\text{COPRO}}
\newcommand{\adal}{\text{Adalflow}}

\definecolor{valgray}{HTML}{BDBDBD}
\definecolor{testgreen}{HTML}{2CA02C}

\usepackage{siunitx}
\usepackage{colortbl}

\renewcommand{\arraystretch}{1.15}

\usepackage{soul}          
\sethlcolor{blue!12}
\newcommand{\hlblue}[1]{{\sethlcolor{blue!12}\hl{#1}}}

\definecolor{BaseTemp}{HTML}{1F77B4}  
\definecolor{TGRev}{HTML}{FF7F0E}     
\definecolor{dspy}{HTML}{FF7F0E}
\definecolor{TGNoRev}{HTML}{2CA02C}   
\definecolor{TSGDpw}{HTML}{9467BD}    

\definecolor{TSGDbl}{HTML}{D62728}    
\definecolor{AFALMbl}{HTML}{FFD580}
\definecolor{GEPAcol}{HTML}{008080} 
\newcommand{\hlc}[2]{%
  {\sethlcolor{#1}\hl{#2}}%
}

\newcommand{\Base}[1]{\hlc{BaseTemp!18}{#1}}
\newcommand{\TGwith}[1]{\hlc{TGRev!22}{#1}}
\newcommand{\TGwithout}[1]{\hlc{TGNoRev!22}{#1}}
\newcommand{\TSGDPromptwise}[1]{\hlc{TSGDpw!22}{#1}}
\newcommand{\TSGDBlockwise}[1]{\hlc{TSGDbl!18}{#1}}

\newcommand{\GEPA}[1]{\hlc{GEPAcol!20}{#1}}

\newcommand{\secref}[1]{\S\ref{#1}}
\AtBeginDocument{%
  }

\begin{document}

\title{Scaling Textual Gradients via Sampling-Based Momentum}


\author{Zixin Ding}
\authornote{Both authors contributed equally to this research.}
\orcid{0009-0007-2227-1046}
\affiliation{%
  \institution{University of Chicago}
  \city{Chicago}
  \state{IL}
  \country{USA}
}
\email{zixin@uchicago.edu}

\author{Junyuan Hong}
\authornotemark[1]
\affiliation{%
  \institution{University of Texas at Austin}
  \city{Austin}
  \country{USA}}
\email{mr.junyuan.hong@gmail.com}

\author{Zhan Shi}
\affiliation{%
  \institution{Santa Clara University}
  \city{Santa Clara}
  \country{USA}
}
\email{ashi2@scu.edu}

\author{Jiachen T. Wang}
\affiliation{%
 \institution{Princeton University}
 \city{Princeton}
 \state{NJ}
 \country{USA}}
 \email{tianhaowang@princeton.edu}

\author{Zinan Lin}
\affiliation{%
  \institution{Microsoft}
  \city{Redmond}
  \state{WA}
  \country{USA}}
  \email{linzinan1995@gmail.com}

\author{Li Yin}
\affiliation{%
  \institution{SylphAI}
  \city{San Francisco}
  \state{CA}
  \country{USA}}
  \email{li@sylph.ai}

\author{Meng Liu}
\affiliation{%
  \institution{SylphAI}
  \city{San Francisco}
  \state{CA}
  \country{USA}}
  \email{meng@sylph.ai}

\author{Zhangyang Wang}
\affiliation{%
  \institution{University of Texas at Austin}
  \city{Austin}
  \state{TX}
  \country{USA}}
\email{atlaswang@utexas.edu}

\author{Yuxin Chen}
\affiliation{%
  \institution{University of Chicago}
  \city{Chicago}
  \state{IL}
  \country{USA}}
\email{chenyuxin@uchicago.edu}

\renewcommand{\shortauthors}{Ding et al.}

\begin{abstract}
  LLM-based prompt optimization, which uses LLM-provided ``textual gradients'' (feedback) to refine prompts, has emerged as an effective method for automatic prompt engineering. However, its scalability and stability are unclear when using more data in training. We systematically investigate the potential and challenges of scaling training data in textual gradient descent. We show that naively scaling training examples is infeasible due to both explicit context-length limits and an implicit context wall, where long-context degradation yields diminishing returns. Inspired by prior wisdom in stochastic gradient descent, we propose Textual Stochastic Gradient Descent with Momentum (TSGD-M), which reweights updates through momentum sampling, using bootstrapped minibatch validation accuracy as importance weights over historical prompts. To stabilize TSGD and enable effective scaling within a limited context window, TSGD-M carries prior prompts information by \textit{dynamically} exploring the past top performing prompts without expanding input context length. TSGD-M integrates seamlessly into existing prompt optimization frameworks, including TextGrad, DSPy-COPRO, and AdalFlow, and achieves consistent gains across 6 benchmarks. 
\end{abstract}



\acmYear{2026}\copyrightyear{2026}
\setcopyright{cc}
\setcctype[4.0]{by}
\acmConference[ACM CAIS '26]{ACM Conference on AI and Agentic Systems}{May 26--29, 2026}{San Jose, CA, USA}
\acmBooktitle{ACM Conference on AI and Agentic Systems (ACM CAIS '26), May 26--29, 2026, San Jose, CA, USA}
\acmDOI{10.1145/3786335.3813168}
\acmISBN{979-8-4007-2415-2/26/05}


\maketitle

\section{Introduction}

With the scaling of pre-training data, Large Language Models (LLMs) have demonstrated impressive capabilities in understanding human language and executing human instructions or prompts~\citep{brown2020language,ouyang2022training}. Meanwhile, the capability of LLMs tends to be confined to the quality of prompts -- a well-written prompt can significantly boost the performance of LLMs, and vice versa, in semantic classification, programming, semantic understanding, and sophisticated agentic tasks~\citep{ying2024internlm, zhou2022large, sclarquantifying, anagnostidis2024susceptible}. 
The importance of prompt to LLM performance has motivated extensive work on Automatic Prompt Engineering, i.e., the method of leveraging LLMs to reflect on the prompts and iteratively refining them~\citep{zhou2022large}.

Recent work~\citep{yuksekgonul2024textgrad,yin2025llm} systematically reformulated these ideas into the \textbf{Textual Gradient Descent} (TGD) framework, which iteratively updates prompts using ``textual gradients'' -- feedback generated by LLMs.
The method closely mirrors numerical gradient descent (GD) in optimization. Just as GD iteratively refines parameters along the gradient direction to minimize a loss function, TGD
extends gradients and parameters into text space, allowing the backpropogated gradients to optimize individual elements of compound AI systems, such as agentic~\citep{yin2025llm} or multi-agent system~\citep{zhou2025multi}.

\begin{figure*}
    \centering
\includegraphics[width=0.99\linewidth,height=3.26cm]{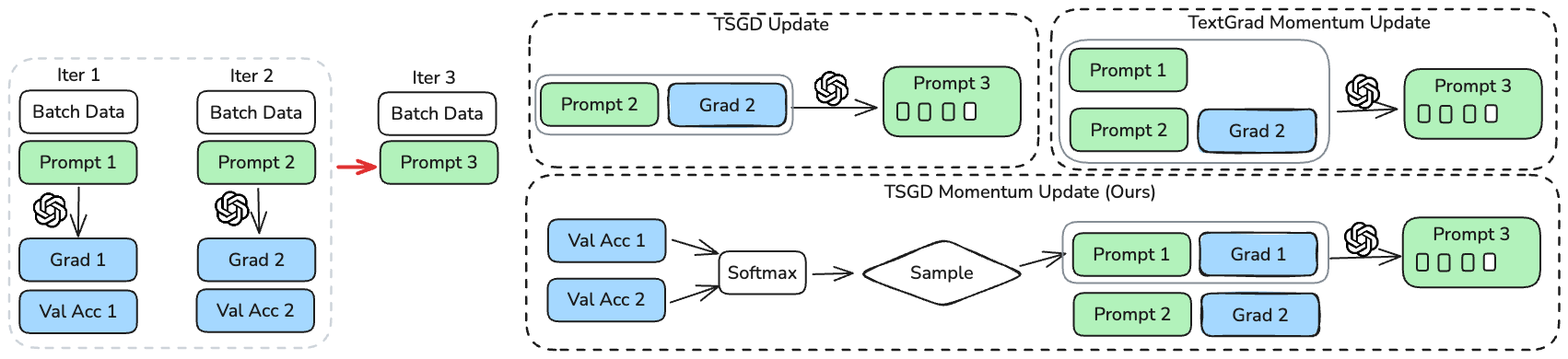}
    \caption{Comparison of TSGD variants. Standard TSGD updates the prompt using the previous prompt and gradient. TextGrad adds momentum by concatenating past prompts in context. Our method upweights historic prompts in sampling that are of higher validation accuracy and only uses a single past prompt–gradient pair to generate the next block of tokens.
    }
    \label{fig:tsgd-m}
\end{figure*}

While TextGrad performs well in low-data settings (e.g., molecular optimization, test-time problem solving), a natural question is : Does its performance improve with more training data, given that data scaling is a central driver of modern AI progress~\citep{hoffmann2022training}?
Despite the analogy between TGD- and GD-based learning, their learning mechanisms are distinct and therefore the scaling law of GD cannot simply extend to TGD. In particular, TGD updates prompts by discrete sampling from LLMs, instead of relying on continual updates with numerical gradients.
Therefore, its scalability hinges on the model’s long-context capability~\citep{agarwal2024many}. In practice, scaling is constrained by the LLM’s finite context window, which limits the number of examples processed during training.

In this paper, we first empirically revisit the effect of scaling training data in TGD with \textit{fixed full batch} per iteration.
We vary the number of samples in the MATH tasks~\citep{hendrycks2measuring} and and evaluate the prompt checkpoint selected by validation accuracy. Surprisingly, we observe two coupled bottlenecks that limit scaling.
~\textbf{First, explicit context windows of LLMs impose a hard cap on in-context examples.} Explicit context limits of LLMs prevent scaling in-context examples \citep{hooper2024squeezed, tang2024quest, kuratov2024babilong} as inference costs grow quadratically with sequence length \citep{Vaswani+2017,chencore}. 
~\textbf{Second, performance encounters an implicit context wall}. Performance increases at the beginning but drops sharply after peaking at small number of samples with generated rationales, even though the input remaining below the LLM's explicit context length limit. When more examples are fed in the context, LLMs struggle to reliably extract actionable “gradient” signal from very long contexts \citep{liu2024lost, peng2023does, du-etal-2025-context}.
Motivated by these constraints, we explore Textual Stochastic Gradient Descent (TSGD) with \textit{stochastically sampled minibatch} per iteration. TSGD shall extend the scaling and achieve better performance by effectively exploring diverse prompts. 

However, actually finding the optimal minibatch is \textit{difficult}. There is a tradeoff in minibatch sizing: smaller minibatches reduce inference cost, yielding overly terse textual gradients with limited suggestion; larger minibatches could produce longer but contradictory editing directions across iterations (See Appendix~\ref{appendix:minibatch_size_textual_gradients}). To stabilize TSGD and make scaling effective, we propose Text Stochastic Gradient Descent with momentum (TSGD-M). TSGD-M is inspired by momentum in classical SGD~\citep{rumelhart1986learning, polyak1964some, liu2020improved}, but adapts it to textual gradients, where gradients are produced via discrete sampling rather than numerical differentiation.
 
\alg \, samples new prompts from a mixture distribution conditioned on past prompts and their associated textual gradients, preserving the same context length as vanilla \textgrad\ without an extra context window (\cref{fig:tsgd-m}). As discrete sampling is non-smooth and classical step-wise decay offers limited stabilization, we replace iteration decay with performance-based decay: higher performing prompts receive larger sampling weights with higher chance for next iteration generation. To enable effective \textit{scaling}, we use minibatch validation performance as an unbiased estimator of full validation accuracy. 
We summarize our contributions as follows:

\noindent$\bullet$ We examine scaling with TSGD and TGD. We identify the key challenges are explicit context-length limits and an implicit context wall 
    where long-context yields diminishing returns (\secref{sec:scaling}).

\noindent$\bullet$ We propose TSGD-M, a textual-gradient method with sampling-based momentum, stabilizing and scaling TSGD by \textit{dynamically} exploring the past top performing prompts (\secref{Section:TSGD-M}). Rather than concatenating historical prompts, we define momentum as a weighted combination of high-performing past prompts, with weights balancing prompt quality against context-length constraints. TSGD-M is framework-agnostic and can be integrated into existing stacks such as TextGrad \citep{yuksekgonul2024textgrad},  DSPy \citep{khattab2024dspy} and AdalFlow \citep{yin2025llm} (\secref{sec:extensions_of_TSGD_M}).

\noindent$\bullet$ Empirical studies show that our method can improve the generalization of prompts generated by baseline methods with statistical significance in multiple benchmarks. With momentum, we further improve upon the TextGrad baseline \citep{yuksekgonul2024textgrad} in HotPotQA \citep{yang2018hotpotqa} by 2.16\% and TREC \citep{lu2022fantastically} by 4\% averaged over 5 independent trials (\secref{sec:experiments}).

\section{Related Work}
\label{sec:prelim}

\begin{table*}[t]
\centering
\caption{Comparison of APE methods. \emph{val} stands for validation set. ``Textual-Gradient-free'' refers to
applying $\Update$ in one step \textbf{without} generating a textual gradient from $\llmbackward$. 
\textbf{Our method} can be extended to other prompt optimization frameworks even without textual gradients, and works for single and multi-prompt sampling with Gumbel-Top-$k$ selection on \emph{val}.
}
\label{tab:baseline-comparison}
\scriptsize
\begin{tabular}{llll}
\toprule
\textbf{Methods} & \textbf{Update Rule} & \textbf{Prompt Generation Context} & \textbf{Selection Criterion} \\
\midrule
APE~\cite{zhou2022large} & Textual-Gradient-free & Sampling multiple prompts (mutation/crossover) & Evolutionary selection \\
OPRO~\cite{yang2024large} & Textual-Gradient-free & Past prompts + demo concatenation & Greedy/ top-$k$ select \\
GEPA~\cite{agrawal2025gepa} & Textual-Gradient-free & Sampling single prompt (mutation/crossover) & Evolutionary selection \\
PromptAgent~\cite{wang2024promptagent} & Textual Gradient & Sampling multiple prompts & Tree search \\
ProTeGi~\cite{pryzant2023automatic} & Textual Gradient & Sampling multiple prompts & Greedy / top-$k$ \\
\midrule
COPRO~\cite{dspy2025optimizers} & Textual-Gradient-free & Sampling multiple prompts & Greedy (val) \\
TextGrad~\cite{yuksekgonul2024textgrad} & Textual Gradient & Sampling single prompt & Greedy (val)\,/\,Revert \\
AdalFlow~\cite{yin2025llm} & Textual Gradient & Past prompts concatenation & Greedy (val) /Revert \\
\alg \, \textbf{(Our method)} & Textual Gradient or Textual-Gradient-free & Sampling multiple prompts & Gumbel-Top-$k$ (val) \\
\bottomrule
\end{tabular}
\end{table*}

\textbf{Automatic Prompt Engineering (APE)} optimizes LLM instructions through iterative self-improvement inspired by evolutionary algorithms \citep{zhou2022large}.
Early methods such as OPRO \citep{yang2024large} and DLN1 \citep{sordoni2023joint} frame LLMs as optimizers that refine prompts via performance feedback or learned proposal distributions.
ProTeGi \citep{pryzant2023automatic} first interprets LLM feedback as textual gradients, while TextGrad \citep{yuksekgonul2024textgrad} formalizes a computation graph for textual gradient descent and momentum integration.
DSPy \citep{khattab2024dspy} extends this idea with structured optimization (COPRO) but without explicit textual gradients.
Recent approaches such as GEPA \citep{agrawal2025gepa} incorporate hierarchical evolution but remain \textit{exploitative} by relying on prompts that have the highest validation accuracy so far.
Our work differs by introducing Gumbel-Top-k momentum, which adaptively reweights historical prompts via a probabilistic categorical distribution, balancing exploration and exploitation without requiring the highest-performing prompt to persist when a newly generated prompt has a temporary validation drop. 
This formulation generalizes TextGrad-style optimization and integrates with frameworks such as DSPy-COPRO and AdalFlow (See ~\cref{tab:baseline-comparison} for comparison). A comprehensive comparison with prior prompt optimization methods is provided in Appendix~\ref{Appendix: extended related work}.

\textbf{Scaling Law for LLMs.}
Scaling laws offer a principled recipe for systematically improving LLMs by trading off parameters, data, and compute~\citep{hoffmann2022training,kaplan2020scaling}. Recent work on test-time scaling shows allocating more inference compute, e.g., generating additional candidates, revising answers, or invoking verifiers, can substantially boost reasoning without changing model weights~\citep{snell2024scaling,muennighoff2025s1,li2025s,brown2024large,ning2024can}.
However, the scaling law for APE is \textit{underexplored}: we do not yet know how performance scales with overall number of training examples, iterations, or evaluations per step. This gap is critical because APE is effectively a test-time optimization loop whose data and compute budgets are bounded by context length and noisy scoring metrics in selecting the best prompt(s) to roll out. Prior observations that prompt optimization helps most in low-data regimes~\citep{yuksekgonul2024textgrad, yang2025ad, wang2024promptagent} raise a central question: what limits data scaling in APE, and how should we allocate additional samples or iterations for predictable gains?
This paper addresses that question by formulating a scaling-aware APE procedure. We analyze where naive scaling breaks (context walls, variance in scoring function estimate, and overly exploitative selection) and introduce a probabilistic sampler that reallocates test-time compute to balance exploration and exploitation over past prompts, enabling more stable scaling in APE.
\vspace{-1.5mm}
\section{Preliminaries}
\label{sec:tgd}

Consider a LLM formally defined as $\llm: \vocab^* \rightarrow \vocab^*$, 
where $\vocab$ denotes the vocabulary set and $\vocab^*$ represents the space of all possible sequences over $\vocab$. 
For a given prompt $\prompt \in \vocab^*$ and input $x \in \vocab^*$, the LLM processes their concatenation $[\prompt, x] \in \vocab^*$ to produce an output sequence. Let $\dataDist$ be a distribution over the input-output pairs $(x, y) \in \vocab^* \times \vocab^*$. 
Suppose we have a dataset $D = \{(x_{i}, y_{i})\}_{i=1}^N$, sampled i.i.d from $\dataDist$. 
The goal of prompt engineering is to identify an optimal prompt $\pi^* \in \vocab^*$ that optimizes the model's expected performance on data drawn from a certain distribution. Formally,
\begin{align}
\pi^* = \arg\max_{\pi \in \vocab^*} \mathbb{E}_{(x,y) \sim \dataDist}\left[\metric\left(\llm([\prompt, x]), y\right)\right],
\end{align}
where $\metric: \vocab^* \times \vocab^* \to \mathbb{R}$ is a metric function evaluating the quality of the model's output against the ground truth, and $[\pi, x]$ denotes the concatenation operation. For simplicity, we do not consider training randomness here.

\begin{figure*}
    \centering
    \subfloat[\label{fig:scaling_left}]{
        \includegraphics[
        height=0.2\textheight,keepaspectratio]{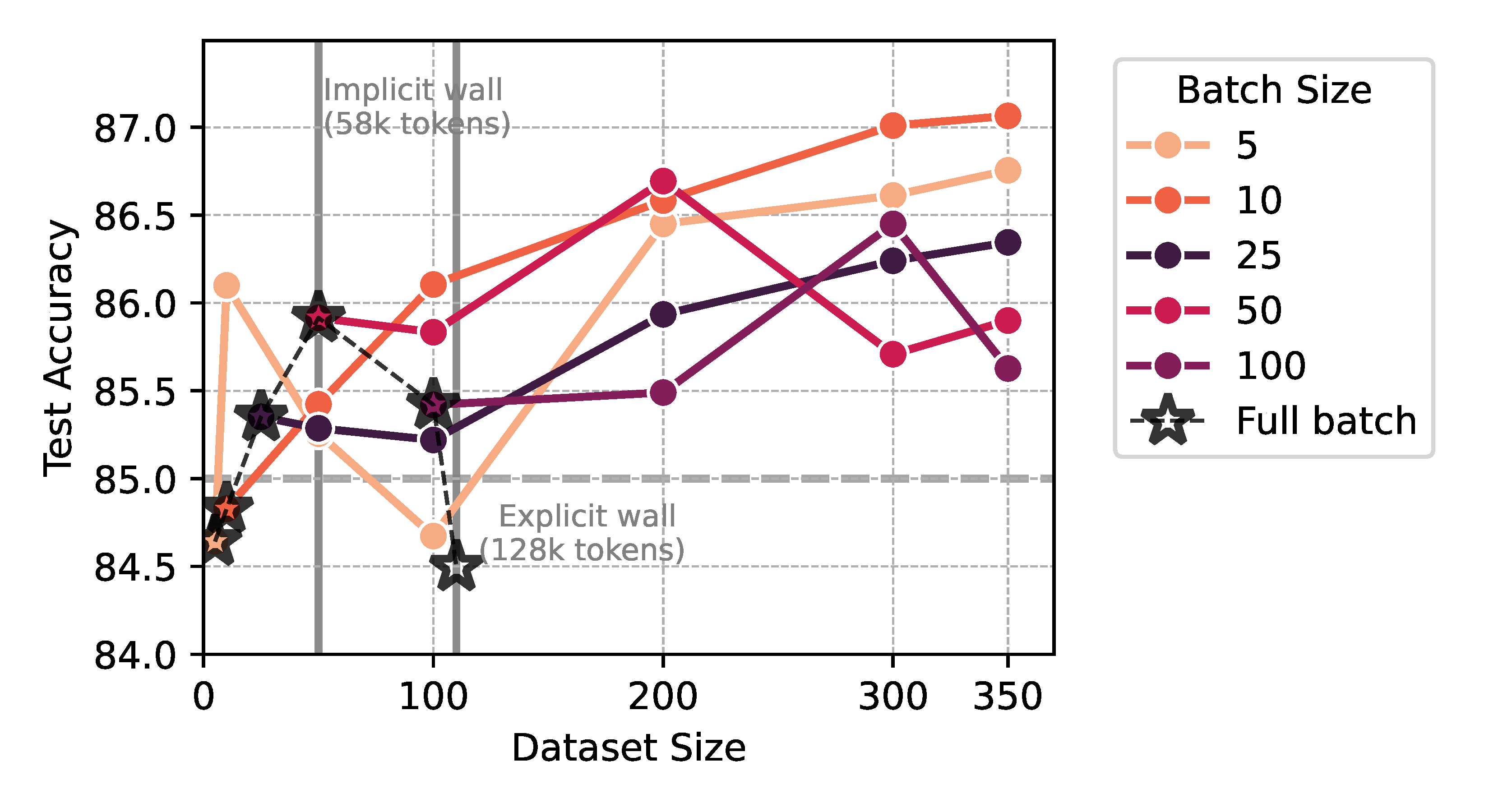}
    }
    \subfloat[\label{fig:scaling_right}]{
\includegraphics[
height=0.2\textheight,keepaspectratio]{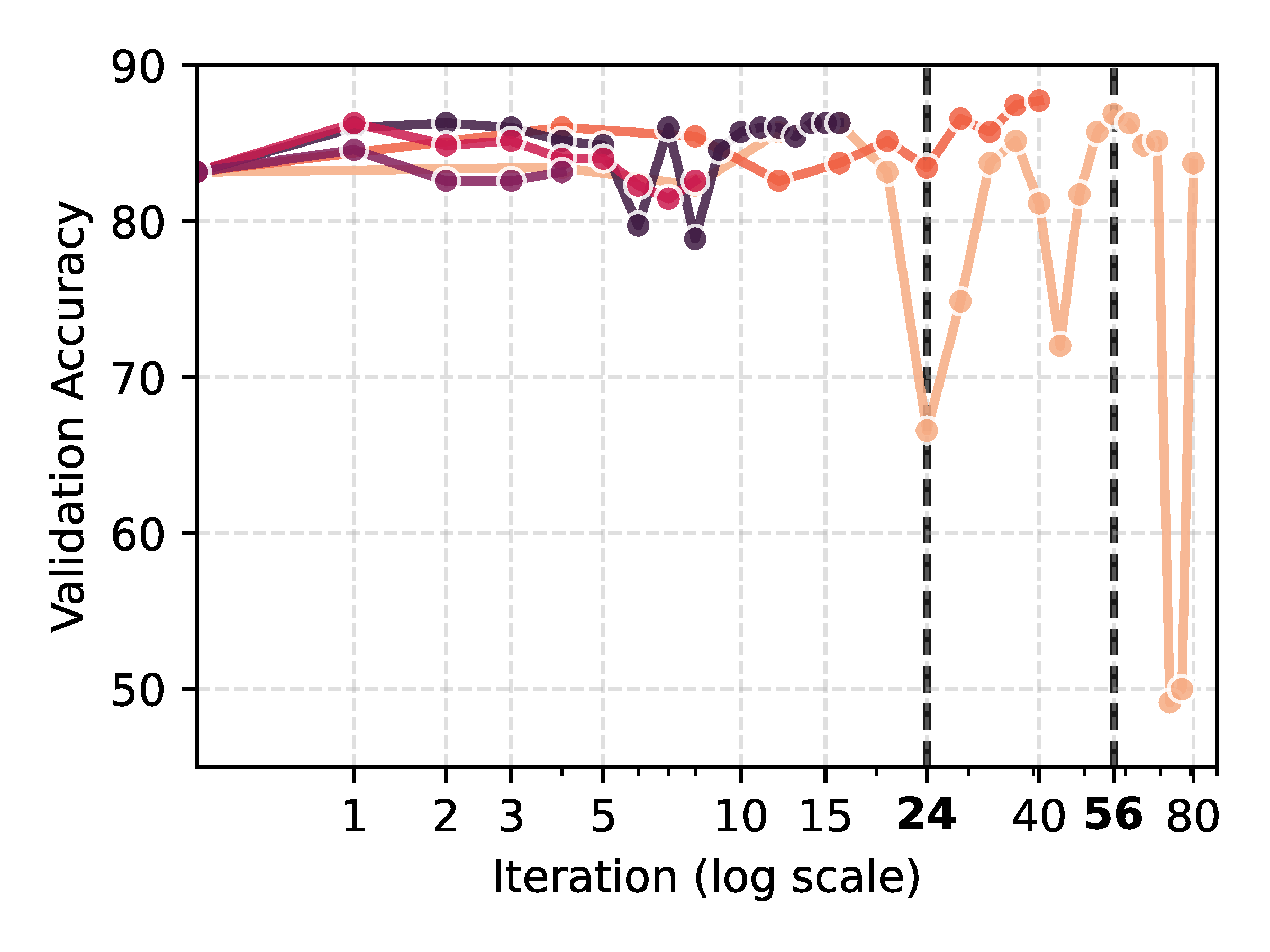}
    }
    \vspace{-3mm}
    \caption{Scaling of TGD/TSGD on MATH. The dashed line indicates average initial test accuracy. \textbf{(a)} Minibatch TSGD enables scaling to larger datasets, whereas full-batch TGD is hindered by an implicit wall (where additional examples reduce accuracy) and an explicit wall (context length limits on $\llmbackward$). \textbf{(b)} Smaller batch sizes exhibit higher validation oscillations, while larger batches iterate more smoothly but may plateau in prompt quality. Extended scaling experiments are in Appendix~\ref{appendix:extended_scaling_experiment}.
    }
    \label{fig:scaling}
    \vspace{-2mm}
\end{figure*}

\textbf{Textual Gradient Descent.}
\emph{Textual Gradient Descent} (TGD) is defined as iterative updates with LLM reflection on quality~\citep{yuksekgonul2024textgrad} and other APE methods can be viewed as variants with different (implicit) reflections. 
To clarify, when stochastically sampled minibatches of training data instead of full dataset are used in each iteration, we call it \emph{Textual Stochastic Gradient Descent} (TSGD).
By default, TSGD refers to the instantiation by \textbf{TextGrad} \cite{yuksekgonul2024textgrad}.
Accordingly, one \textbf{epoch} is defined as the process when all samples are visited by the inference language model \llmforward and we consider stochastic sampling via shuffling in an epoch using different random seeds. Within each epoch, examples are shuffled; different random seeds induce different permutations, so each seed yields a distinct visitation order.
The prompt $\prompt$ serves as the parameter being optimized. For iteration $t$, we denote the current prompt as $\prompt_t$ and sample a minibatch of data $\{(x^{(t)}_i,y^{(t)}_i)\}_{i=1}^m$ with size $m$ uniformly at random from $D$. 
For $x_i^{(t)}$, we obtain the inference LLM prediction $\widehat{y}_i^{(t)} = \llmforward([\prompt_t, x_i^{(t)}])$. The prompt update rule is written as:
\begin{align*}
\prompt_{t+1} = \Update(\llmbackward, \prompt_t, \{(x_i^{(t)}, y_i^{(t)}, \widehat{y}_i^{(t)})\}_{i=1}^m)
\end{align*}
where $\Update$ is an algorithm that leverages backward LLMs' capabilities to analyze discrepancies between predictions $\{\widehat{y}_i^{(t)}\}_{i=1}^m$ and ground truth labels $\{y_i^{(t)}\}_{i=1}^m$ and yield an improved prompt $\prompt_{t+1}$.
In TGD, the process is analogous to the use of numerical gradients in backpropagation to optimize neural network weights.

In TGD, the $\Update$ is formalized as a three-stage process \citep{pryzant2023automatic,yuksekgonul2024textgrad}.
Given the current prompt $\prompt_t$ and a batch of example triplets $\{(x_i, y_i^{(t)}, \widehat{y}_i^{(t)})\}_{i=1}^m$, the $\Update$ algorithm extends to:
\begin{align*}
\frac{\partial L}{\partial \widehat{y}^{(t)}} &= \llmbackward([\prompt_{\text{analyze}}, \{ (x_{i}, y_{i}^{(t)}, \widehat y_{i}^{(t)}) \}_{i = 1}^{m}]) \\
    g_{t} \coloneqq \frac{\partial L}{\partial \prompt_{t}} &= \llmbackward ([\prompt_{\text{generating\_gradient}},\prompt_{t},  \{x_{i}, y_{i}^{(t)} \}_{i=1}^{m}, \frac{\partial L}{\partial \widehat y^{(t)}}])
  \\
  \prompt_{t+1} &= \llmbackward \bigl([\prompt_{\text{refine}},\,\prompt_t,\, g_{t}]\bigr).
\end{align*}
where $\frac{\partial L}{\partial \widehat y^{(t)}}$ denotes the feedback obtained from $\llmbackward$ and $g_{t}$ denotes an error analysis that captures systematic discrepancies between predictions and ground truth resulted by the input prompt $\prompt_{t}$, analogous to a gradient in traditional optimization \citep{pryzant2023automatic}. This textual gradient indicates the direction for prompt improvement, with $\prompt_{\text{analyze}}$ directing the LLM to identify error patterns (i.e. generating structured criticism), $\prompt_{\text{generating\_gradient}}$ creating editing direction (i.e. forming the textual gradient) and $\prompt_{\text{refine}}$ instructing the LLM to update the prompt accordingly (i.e. performing the \textit{descent} step). This three-stage approach creates a conceptual parallel to how gradient descent uses the numerical gradient to update parameters.

\vspace{-2mm}
\section{Scaling Textual Gradient Descent}
\label{sec:scaling}
\textbf{TGD} generates each prompt update using all available examplars (``full‐batch'') with larger input context, whereas \textbf{TSGD} samples only a \textit{subset} per iteration, yielding noisier but more scalable textual gradients due to shorter input context. Prior work shows increasing the number of in-context examples can improve accuracy and fluency while reducing variance \citep{hao2022structured,zhou2024enhancing}. However, excessively long contexts may degrade reasoning performance, particularly in single-pass settings \citep{levy2024same, du-etal-2025-context}. In reality, TGD and TSGD often require multiple update steps. For instance, original TSGD requires 12 iterations with 3 examples per batch~\citep{yuksekgonul2024textgrad}. This raises a key question under the context of \textit{scaling}: how—and how much—should we \textit{scale} data in these iterative prompt‐optimization frameworks?

\textbf{Setup.} We adopt \textgrad~\citep{yuksekgonul2024textgrad} as a representative method as it is conceptually clean: it performs pure textual-gradient updates without auxiliary selection heuristics, making it one of the most intuitive APE methods. We follow the DSPy math tutorial setup and use the MATH Algebra subset, keeping the original split sizes: 350 training examples and 350 validation examples~\citep{DSPyMathReasoning2025}. We explore sampling training dataset size ranging from 5 to 350 with batch sizes scaled from 5 to 110, up to the maximum context limit for both \llmforward and \llmbackward. We use GPT-4o-mini as inference (\llmforward) and GPT-4o as feedback (\llmbackward) that generate textual gradients and refine the prompts. In our runs, the canonical TextGrad “validation-revert” (always restoring the prompt with the highest validation score) did not improve validation performance (see Fig.~\ref{fig:MATH} in \secref{sec:momentum_improves_scaling}); accordingly, we disable revert and generate each new prompt from the immediately previous one. To reduce stochasticity, we set \textit{epochs} 2, suggesting that each example would be seen \textit{twice} during \textgrad \, optimization. We repeat each experiment five times using 5 random seeds throughout this paper. Extended scaling experiments (including runtime) are deferred to Appendix~\ref{appendix:extended_scaling_experiment}.

\textbf{Results.}
In \cref{fig:scaling}, we present the scaling effects by TSGD with varying batch sizes.
We observe that full-batch TSGD (or TGD) is limited by the context length and the implicit context length.
Only 110 samples can be used in TGD. 
Although both GPT-4o-mini and GPT-4o support context windows up to $C=128{,}000$ tokens, the number of TextGrad \emph{accepted} examples that can be processed per update is constrained by the backward pass. In particular, $\llmbackward$ receives the full tuple $(x,r,y,\hat r,\hat y)$, where $\hat r$ is the reasoning trace generated by $\llmforward$ and $r$ is the "gold" reasoning chain provided by $\dataset$ if it exists; it uses $\hat{r}$ to analyze error patterns and synthesize a gradient-like update signal. Let $\bar{\ell}(\cdot)$ denote average token length; then the number of accepted batch size $m$ that fit in $\llmbackward$ per step is bounded by
\begin{align*}
m \;<\;
\frac{C}{ \bar{\ell}(x)+\bar{\ell}(r)+\bar{\ell}(y)+ \bar{\ell}(\hat r)+ \bar{\ell}(\hat y)}
\end{align*}

On \textsc{MATH}, the predicted reasoning trace typically dominates the budget: $\ell(\hat r)\approx 700$ tokens on average, compared to $\ell(x)\approx 150$ tokens for the input and $\ell(r)\approx 60$ tokens for the gold rationale, while iterative prompt growth can reach $\ell(\prompt)\lesssim 2000$ tokens per iteration. Consequently, long reasoning traces (and new prompts with growing length) substantially reduce the effective batch size despite the large nominal context window. 

We observe an implicit context wall at dataset size $N = 50$: beyond this point, the performance of full-batch TGD degrades as more examples are added. In contrast, TSGD with smaller minibatches per iteration shall keep the per-step context shorter and enables scaling to larger dataset sizes. TSGD exhibits extended scaling behavior, and with $N = 350$ and batch size $m = 10$, it outperforms the best TGD configuration by more than 1 \% in test accuracy.
However, minibatch examples introduces instability over iterations, and poorly performing prompts often lag far behind the best ones. For instance, Figure~\ref{fig:scaling} (b) shows even with the same batch size as 5, stopping at iteration 56 yields more than 20\% validation accuracy than stopping at iteration 24, which shall confuse practitioners when deciding when to stop.
Due to the instability, more iterations are required to find a good prompt. Increasing the batch size from 5 to 10 slightly stabilizes training with smoother validation accuracy over iterations (Fig~\ref{fig:scaling} (b)), whereas further increasing the batch size reduce test performance compared to smaller batch size (Fig~\ref{fig:scaling} (a)).

Unlike standard SGD, TSGD does not exhibit steady convergence at either large or small batch sizes, highlighting the importance of \textit{adaptively} selecting prompts based on validation performance (Figure~\ref{fig:scaling} (b)) rather than simply taking the final prompt at the end of iterations.
Overall, the key challenge for scaling TSGD is to maintain stability across iterations while reducing batch size. We also provide qualitative analysis for textual gradients in Appendix~\ref{appendix:minibatch_size_textual_gradients}.
\section{TSGD-M}
\label{Section:TSGD-M}

In this section, we introduce a method to improve the stability of TSGD without increasing the minibatch. The core primitive is to retain small batches while reusing signals from \textit{preceding} batches, analogous to momentum in SGD, so updates aggregate information over time rather than relying on a single, noisy textual gradient. We summarize our algorithm in 
\cref{alg:tsgd-m}, with an elaborated version provided in the Appendix~\ref{section:extended_algorithm} (\cref{alg:gumbel-top-k}).

For brevity, we unify the notation of prompt generation from either $\llmforward$ or $\llmbackward$ to $P$ since all generation can be viewed as sampling from the LLM.
Following the notations in \secref{sec:tgd}, we define some simplified terms for the ease of discussion.
With the gradient $g_t$, TSGD updates the prompt by sampling, $\pi_{t+1} \sim P(\pi |\pi_{t}, g_t)$.
In practice, the generation is done token by token:
\begin{align*}
    \pi_{t+1}^j &\sim P(\pi_{t+1}^{j} | \pi_{t+1}^{<j}, \pi_{t}, g_t).%
\end{align*}
$\pi_{t+1}^{<j}$ is the prefix up to token $j$; $\pi_{t+1}^j$ is token $j$. 

\begin{algorithm}[t]
\caption{Textual Stochastic Gradient Descent with Momentum (TSGD-M)}\label{alg:tsgd-momentum}
\label{alg:tsgd-m}
\begin{algorithmic}[1]
\REQUIRE Initial prompt $\prompt_{0}$, training set $\mathcal{D}_{\text{train}}$, validation set $\mathcal{D}_{\text{val}}$, total steps $T$, momentum window size $K$, scoring function on accuracy $S$
\STATE Initialize cache $\Phi \leftarrow \varnothing$
\FOR{$\tau = 0 \dots T-1$}
  \STATE Draw a training minibatch $B_{\text{train}} \sim \mathcal{D}_{\text{train}}$ and a validation minibatch $B_{\text{val}} \sim \mathcal{D}_{\text{val}}$
  \STATE $v_\tau \gets S(P(\prompt_\tau, B_{\text{train}}), y)$
  \STATE Sample $K$ prompts $\Pi_\tau^*$ from all past iterations based on their running mean over all cached val acc.
  \STATE Evaluate validation accuracy of $\Pi_\tau^*$ on $B_{\text{val}}$, denoted as $V$. Add $\{ \Pi_{\tau}^{*}, V \}$ to cache.
  \STATE Select $\pi_\tau$ based on $\{ \Pi_{\tau}^{*}, V \}$ using \cref{eq:promptwise_generation} or \cref{eq:blockwise_generation}
  \STATE Generate next prompt: $\prompt_{\tau+1} = P(\pi | g_\tau, \pi_\tau)$
  \STATE Compute gradient $g_{\tau+1}$ on the $\pi_{\tau+1}$ and add $(\tau+1, \pi_{\tau+1}, g_{\tau+1}, [v_\tau])$ to $\Phi$
\ENDFOR
\ENSURE Optimized prompt $\pi_{T}$
\end{algorithmic}
\end{algorithm}

\subsection{Momentum from the First Principle}
\label{sec:Momentum_from_the_first_principle}
\textbf{Token-wise Momentum Avoids Forgetting.}
Vanilla TSGD selects the top-performing validation prompt and ignores the rest. Instead, we leverage momentum to combine previous prompts by
\vspace{-0.05in}
\begin{align}
    P_{\tau}(\cdot) := P(\pi^{j} | \pi^{<j}, \pi_{\tau}, g_\tau),\quad \pi_{t+1}^j \sim \sum \nolimits_{\tau=1}^t \alpha_\tau P_{\tau}(\cdot) \label{eq:mix_token_pdf}%
\end{align}
\vspace{-0.02in}
where the next token is generated based on the ensemble of all prior distributions.
$\alpha_\tau$ is a normalization factor such that $\sum_{\tau=1}^t \alpha_\tau=1$.

\textbf{Memory Decay.}
A standard momentum-style ensemble assigns higher weight to more recent prompts/gradients via exponential recency weighting $\alpha_\tau \propto q^{t - \tau}$, $0<q<1$ so that older trajectories are exponentially downweighted. 
In reality, TSGD trajectories
can be non-monotone. Therefore, we use performance-based weighting over each previously generated prompt rather than
recency weighting, i.e., $\alpha_\tau \propto \text{SoftMax}(s_\tau)$ where $s_\tau$ is the validation accuracy.

In optimization, the Gumbel-Top-$k$ trick refers to obtain distinct $k$ samples without replacement from a categorical distribution by adding i.i.d. Gumbel(0,1) noise to logits and selecting the top-$k$ perturbed scores \citep{kirschstochastic,kool2019stochastic, maddison2014sampling}. With a slight abuse of notation, we use $K$ to denote the number of selected prompts produced by Gumbel-Top-$k$ trick. In APE, we view each generated prompt $\prompt_{\tau}$ (for $\tau = 1, ... t$) as a class in a categorical distribution with logits defined by their validation scores $s_{\tau}$. We then sample $K$ distinct prompts without replacement by computing $S_{\tau} = s_{\tau} + \epsilon_{\tau}$, $\epsilon_{\tau} \sim \text{Gumbel}(0,1)$ and select the indices corresponding to the $K$ largest perturbed scores $\{ S_{\tau} \}$ (\cref{proposition:gumbel_top_k} in Appendix~\ref{Appendix:theoretic_justification}).
This procedure produces samples without replacement according to the Softmax weights while preserving ranking consistency and enabling effiecient sampling in one pass. It encourages stochastic yet structured \textit{exploration}-e.g., picking multiple top-performing prompts, using only additive noise and sorting, making it attractive for large-scale ML systems where scalability and controlled diversity are key.

\textbf{Efficiency Bottleneck.}
However, the complexity could be high.
\textbf{1) Inference is expensive.}
If generating on one GPU, we are not able to cache KV data to speed up the computation resulting in slow inference.
If each individual distribution is estimated on one GPU, the space complexity multiplies linearly with $t$. Soon, the number of GPUs will be insufficient.
\textbf{2) Evaluating validation accuracy $s_\tau$ is expensive.}
To obtain a nearly \textit{unbiased} estimator of the prompt quality, validation set has to be large enough, resulting large cost in forwarding.
In particular, smaller validation set increases the number of iterations and corresponding number of validations.
We address the two challenges in the following two sections.
\vspace{-1.5mm}
\subsection{Inference via Block-wise Generation}
\label{section:efficient_inference_via_blockwise_generation}
As \cref{eq:mix_token_pdf} defines a mixture model, we do not need to evaluate the marginal $\pi_{t+1}^{j}$ explicitly.
Instead, we can \textbf{1)} first sample from the categorical distribution $P(t=\tau) = \alpha_\tau$, and \textbf{2)} sample the token $\prompt_{t+1}^{j}$ from the conditional distribution $P(\pi_{t+1}^{j} | \pi_{t+1}^{<j}, g_\tau, \pi_{\tau})$.

Prior approaches frequently \emph{reload} prompts during decoding, which prevents effective KV-cache reuse. Furthermore, token-by-token prompt switching is often infeasible in production LLM endpoints: modern serving stacks commonly use continuous batching and speculative decoding, where the backend may draft and validate \textit{multiple} tokens per step (rather than strictly generating \textit{one} token, appending it, an re-entering decoding) \citep{liu2024optimizing}.
In such settings, changing the prompt between adjacent tokens would require restarting the decode loop, effectively discarding (or recomputing) the cached prefix states and defeating prefix/KV-cache reuse~\citep{vllm_prefix_caching_docs, zheng2024sglang}. In practice, adjacent tokens are largely determined by short-range context and grammar rather than by per-token gradient signals or prompt changes~\citep{xiaoefficient, queipo-de-llano2026attention}. We therefore avoid token-by-token prompt switching and propose two efficient alternatives that preserve the KV cache:
\textbf{1) Blockwise Generation.}
Let $b$ be a fixed block size and index blocks by $i=1,\ldots,\lfloor T_{\max}/b \rfloor$. At iteration $\tau$, after shortlisting $K$ candidates $\Pi_{\tau}^{\star}=\{\prompt_{\tau,1},\dots,\prompt_{\tau,K}\}$,
we first sample a single prompt $\prompt_{\tau,k^\star} \in \Pi_{\tau}^{\star}$ uniformly at random, and then generate the entire block $\tokenblock_{i+1}$ using $\prompt_{\tau,k^\star}$ while keeping the prompt fixed within the block:
\begin{align}
    \tokenblock_{i+1} \sim P(\tokenblock_{i+1} 
\mid \tokenblock_{1:i}, \prompt_{\tau,k^\star}) \label{eq:blockwise_generation}
\end{align}
This design balances the theoretical benefit of tokenwise momentum with the practicalities of KV-cache efficiency. By setting $b > 1$, we reduce the frequency of momentum sampling and prompt switching by a factor of $b$. This allows the system to utilize modern serving optimizations like continuous batching and speculative deocding within each block, significantly reducing the overhead of KV-cache invalidation at the cost of a slightly coarser approximation of the first-principle momentum. Blockwise generation amortizes attention computation by reusing the KV-cache across blocks. Instead of regenerating the entire prefix, we retain the cached KV tensors for previously generated blocks. When generating the next block, only the new tokens are processed and appended to the cache. The computation scales linearly with the incremental block length.
\textbf{2) Promptwise Generation.}
Promptwise Generation shall be viewed as a \emph{special} case of the blockwise approach where the blockwise $b$ is equal to the total generation length $T_{\max}$. In this regime, momentum is applied exactly once at the beginning of the generation.
At iteration $\tau$, after shortlisting $K$ prompts as $\Pi_{\tau}^{*}$, we draw a fresh minibatch and evaluate $\prompt_{\tau, k} \in \Pi_{\tau}^{*}$ on the minibatch and obtain scores $V = \{ v_{\tau,k} \}_{\pi_{\tau,k} \in \Pi_\tau^*}$:
\begin{align}
\prompt_{\tau}^{\star}
= \operatorname*{arg\,max} \; V,
\quad
 \prompt_{\tau + 1} = P(\prompt_{\tau}^{\star}).
 \label{eq:promptwise_generation}
\end{align}
While this minimizes inference costs and maximizes KV-cache reuse--making it appealing for most closed-source LLM APIs while sacrificing the \emph{adaptivity} to the validation accuracy feedback compared against blockwise momentum sampling.
\subsection{Efficient Validation via
Minibatches}
\label{section:efficient_validation_via_stochastic_minibatches}
\textbf{Minibatch Validation with Momentum.}
Our decayed sampling strategy requires reliable performance estimates, yet evaluating on the full validation set per iteration is computationally prohibitive. 
Instead, at iteration $\tau$, we draw a random validation minibatch and evaluate only the shortlisted prompts. To mitigate minibatch noise and bias, a prompt is re-evaluated only when selected; its validation estimate is then updated via a running mean. We maintain a cache $\Phi$ that stores, for each prompt $\prompt_{i}$, its evaluation count $n_i$ and the sequence of minibatch accuracies collected whenever $i$ appears among the top $K$ candidates. 
The running estimate $\mu_{i}'$ for $\prompt_{i}$ is then computed as the average of its accumulated minibatch validation accuracies divided by $n_{i}$.
Each iteration proceeds in two stages:
\textbf{(1)~Screening:} sample a minibatch and evaluate a Top-$K$ set of prompts; 
\textbf{(2)~Refinement:} select one candidate from this set, balancing exploration and exploitation, and generate its refinement. 
Choosing the next prompt thus forms a sequential decision-making problem that explicitly trades off exploring novel prompts against exploiting high-scoring ones~\citep{lattimore2020bandit, kirschstochastic}.
\\
\textbf{Exploration vs. Exploitation.}
When validation relies on a small set, greedy selection of the prompt with the highest validation accuracy can overfit to sampling noise and deviate from the true best prompt under full validation. 
This creates a classical exploration--exploitation tradeoff: the algorithm must balance exploiting the current best estimate against exploring uncertain but promising alternatives \citep{kool2019stochastic,maddison2014sampling}.
Our method follows a two-stage decision rule inspired by batched active learning \citep{kirschstochastic}. \\
\textbf{(1) Exploration.}
We apply Gumbel–Top-$k$ sampling to the prompt pool, injecting stochasticity into the ranking
(Proposition~\ref{proposition:gumbel_top_k}).
This ensures strong (though not necessarily optimal) prompts retain a nonzero selection probability, thereby encouraging exploration of the upper tail rather than \emph{pointwise} greedy exploitation.
Repeated resampling across iterations provides diverse coverage of the validation set and adaptive reweighting from fresh minibatch feedback, forming a reduced-bias bootstrap estimate of prompt quality. \\
\textbf{(2) Generation.}
After obtaining $\Pi_\tau^{\star}$ per iteration, we perform exploitation through two modes of generation.
In the \textbf{Promptwise} mode, we evaluate all $K$ shortlisted prompts on the \emph{same} minibatch and select the prompt with the highest validation accuracy. Under equal-precision posteriors, the Bayes-optimal prompt is the argmax of posterior means; with a shared fresh minibatch this coincides with the best prompt evaluated over that batch (\cref{thm:equal-precision}). In contrast, applying the Gumbel–max trick to the $K$ running means introduces unnecessary randomization and is Bayes-suboptimal for this objective.
In the \textbf{Blockwise} mode, we instead generate tokens in contiguous blocks and uniformly sample among $K$ shortlisted prompts for each block.
Blockwise generation has variance no larger than
TextGrad without validation revert, and becomes strictly smaller under mild
conditions (\cref{thm:tsgdm-variance-short}). System-wise, uniform block sampling regularizes the inter-arrival pattern of prompts, improving KV-cache reuse and reducing tail latency.
\vspace{-1.5mm}
\subsection{Complexity Analysis}
\label{sec:complexity_analysis}
Let $T$ be the number of iterations, $K$ the Gumbel–Top-$k$ size, $|v|$ the minibatch validation size, $m$ minibatch size per iteration and $|D_{\text{train}}|$ the full training set.
Let $c_f$ be the (constant) tokens per example for \llmforward\ during validation,
$c_m$ the tokens to run \llmbackward\ once to produce a new prompt, $c_g$ the tokens to produce textual gradients per iteration (batch-size effects are absorbed into $c_g$), and $c_{g}^{\text{train}}(s)$ tokens to run the training-side inferencing over $s$ training examples ($s \in \{m, |\mathcal{D}_{\text{train}}| \}$).
Per-iteration on a minibatch validation costs $O(|v|)$. Per-iteration on full training set evaluation costs $O(|\mathcal{D}_{\text{train}}|)$ and batch training set $O(m)$. We use $O(\cdot)$ notation and suppress constant factors and additive lower-order terms.

For clarity, we denote the concatenation of past prompts with \textgrad\ as \textbf{\textgrad-Momentum} and our \alg\ module as \textbf{\textgrad-M}. We also analyze full batch \textgrad\ as \textbf{TGD} with fixed full-training set per iteration and minibatch TextGrad as \textbf{TSGD} with stochastically sampled minibatch per iteration.

Let $L_p$ denote the average token length contributed by one prior prompt when \textgrad-Momentum concatenates the last $K$ prompts into \llmbackward. 
Thus \textgrad-Momentum’s backward pass scales as $c_m + K\,L_p$, whereas \textgrad-M (ours) keeps the backward prompt length fixed (no concatenation), so its backward cost remains $c_m$ and momentum is realized via Gumbel-Top-$k$.

We can compare the per-iteration costs as follows:
\begin{itemize}[leftmargin=1.5em, itemsep=1pt, topsep=1pt, parsep=0pt]
  \item \textbf{TSGD}: 
  $O\!\big(c_m + c_g^{\text{train}} \cdot m \big)$
  \item \textbf{TGD}:
  $O\!\big( c_{m} + c_{g}^{\text{train}}(|\mathcal{D}_{\text{train}}|) ) $
  \item \textbf{\textgrad-M (Ours)}: 
  $K$ candidates evaluated on minibatches 
  $O\!\big(c_m + c_g^{\text{train}}\cdot m + K\,|v|\,c_f\big)$
  \item \textbf{TextGrad-Momentum (concat $K$ prompts into \llmbackward)}:
  Backward context grows with $K$:
  $O\!\big(c_m + K\,L_p + c_g^{\text{train}}\cdot m 
  \big)$
\end{itemize}
\textgrad-M pays $K$ validation minibatches and \textgrad-Momentum has backward prompt growing by $K \cdot L_{p}$. TGD is the most computationally expensive as $c_{g}^{\text{train}}(|\mathcal{D}_{\text{train}}|) \gg c_{g}^{\text{train}} \cdot m$ due to full training data pass per iteration. TSGD is the cheapest module, but exhibits instability for small $b$ (\cref{fig:scaling} (b)).
As our ablation study shows the minibatch size $|v|$ shall be small compared to full validation set (See ~\cref{tab:sensitivity_analysis} in Appendix~\ref{appendix:sensitivity_analysis}) and $L_{p}$ is generally large (as denoted by the final optimized prompt length in Appendix~\ref{subsection:best_prompts}), \textgrad-M shall be cheaper than \textgrad-Momentum. Additionally, concatenation-based momentum inflates the backward context by $K \cdot L_{p}$, increasing context and KV-cache load. Therefore, TextGrad-Momentum raises the backward prompt cost per iteration to $c_{m} + K L_{p}$. In contrast, our TextGrad-M keeps the backward prompt fixed (only $c_m$) while directing exploration into minibatch validation, reducing context length and KV-cache load. Finally, we have the practical ordering that TSGD $\ll$ \textgrad-M $<$ \textgrad-Momentum $\ll$ TGD in terms of tokens cost. In~\secref{sec:minibatch_vs_full}, we show TextGrad-M is robust regardless of whether the final reported prompt is chosen by the running mean of minibatch validation or by a full validation. We present full computational complexity of generation in Appendix~\ref{appendix:complexity_analysis}.
\vspace{-1mm}
\subsection{Extensions of TSGD-M}
\label{sec:extensions_of_TSGD_M}
Our method can be naturally extended to other methods that share principles with TSGD. We highlight two representative frameworks: (1) DSPy (COPRO)~\citep{dspy2025optimizers}: This method can be adapted to sample multiple prompts and remains robust under Textual-Gradient-free settings. We include it to demonstrate that our method integrate seamlessly into prompt optimization frameworks that do not rely on textual gradients. (2) AdalFlow~\citep{yin2025llm}: As a representative method (with 4k+ github stars) capable of sampling single or multiple prompts, while using concatenations of past prompts as input, AdalFlow offers broader flexibility than many existing baselines. 

\textbf{Extension to DSPy COPRO.} \dspy\ is a declarative programming framework for compiling LLM pipelines and optimizing their prompts without "textual gradients".
COPRO exposes two key hyperparameters: \emph{depth} (number of optimization iterations) and \emph{breadth} (number of candidates generated per iteration). 
Because COPRO assumes \(\text{breadth} > 1\), it naturally extends \textgrad's single-candidate update to a multi-candidate regime. 
In our implementation, we set \(\text{breadth}=2\) but it can be generalized to any breadth number.
At iteration \(\tau\), we first sample a Top-\(K\) set \(\Pi_{\tau}^{\star}\) using the Gumbel–Top-\(k\) trick; we then select the top-scoring prompt in \(\Pi_{\tau}^{\star}\) (under the current minibatch estimate) and use it to instantiate two descendants for evaluation, rather than producing a single descendant as in vanilla \textgrad\ (See generation templates in Appendix~\ref{Appendix:Templates}).  
This coupling preserves exploration via stochastic sampling while retaining exploitation through breadth-wise refinement.

\textbf{Extension to Adalflow.} 
AdalFlow is a PyTorch-like library that enables seamless prompt optimization across textual, numerical, and functional components, supporting both simple and complex computation graphs without textual gradients. AdalFlow-M applies momentum-based prompt selection before each gradient update. 
At iteration \(\tau\), we draw two Top-$K$ sets from the history cache using the Gumbel-Top-$k$ trick: $\Pi_{\tau}^{*}$, weighted by average validation scores, and $\Pi_{\tau}^{*'}$ weighted by the negative scores of average validation scores (worst performing prompts).
AdalFlow maintains dual historical tracking per iteration $\tau$: \emph{best\_prompt\_history} records top-performing prompts $\Pi_{\tau}^{*}$ with averaged scores across multiple mini-batch evaluations, while \emph{failed\_proposal\_history} logs rejected candidates $\Pi_{\tau}^{*'}$ to prevent redundant exploration. 
The selected prompt candidates are subsequently passed to the optimizer model along with $\Pi_{\tau}^{*}$ and $\Pi_{\tau}^{*'}$ to guide refinement.  
In each iteration, AdalFlow generates \(K\) proposals for gradient updates.
Only prompts that show improved performance on the mini-batch evaluation are accepted, ensuring that optimization strictly favors performance gains while leveraging historical momentum for structured exploration. We do not apply block-wise uniform sampling because AdalFlow-M maintains two distinct $K$ sized candidate sets-one for top and one for failed prompts, making uniform averaging ill-defined and potentially doubling compute overhead.

\begin{table*}[ht]
\centering
\caption{Test accuracy (\%) $\pm$ std dev. $\llmforward$: GPT-4o-mini; $\llmbackward$: GPT-4o. \alg\ improves TSGD variants.}
\label{tab:benchmark}
\scriptsize
\setlength{\tabcolsep}{4pt}
\begin{tabular}{lcccccccc}
\toprule
\textbf{Method} & \textbf{TREC} & \textbf{ARC-Challenge} & \textbf{GSM8K} & \textbf{MATH} & \textbf{HotPotQA} &\textbf{IFBench} & \textbf{Aggregate} & \textbf{Improvement} \\
\midrule
\textgrad\, w/o val revert & 81.92(1.53) & 91.35(0.30) & 93.15(0.55) & 84.67(0.58) & 49.46(0.87) & 37.20(0.76) & 72.96 & -- \\
\textgrad\, w/ val revert & 77.40(0.55) & 91.20(0.0) & 93.74(0.27) & 85.42(0.00) & 49.06(0.30) & 37.50(0.05) & 72.39 & -- \\
\textbf{\textgrad-M (Promptwise)} & 83.36(2.02) & 91.96(1.01) & \textbf{94.04(0.24)} & \textbf{86.78(0.68)} & 50.53(1.02)  & 38.02(0.28) & 74.12 & +1.73 \\
\textbf{\textgrad-M (Blockwise)} & \textbf{85.44(0.78)} & \textbf{92.64(0.28)} & 93.98(0.20) & 86.45(0.38) & \textbf{50.66(1.93)} & \textbf{38.35(0.32)} & 74.59 & \textbf{+2.20} \\
\midrule
COPRO & 80.28(3.38) & 94.05(0.27) & 87.99(1.87) & \textbf{70.84(0.48)} & 39.80(1.68) & 34.12(0.68) & 67.84 & -- \\
\textbf{COPRO-M (Promptwise)} & \textbf{83.36(1.99)} & 94.40(0.43) & \textbf{88.74(1.63)} & 68.75(1.10) & 41.23(1.05) & 36.25(0.21) & 68.79 & +0.94 \\
\textbf{COPRO-M (Blockwise)} & 82.52(0.96) & \textbf{94.49(0.48)} & 86.90(0.71) & 69.98(2.51) & \textbf{41.96(1.22)} & \textbf{37.52 (0.28)} & 68.90 & \textbf{+1.05} \\
\midrule
AdalFlow & \textbf{85.00(0.41)} & 91.00(0.71) & 89.00(0.55) & 81.90(0.14) & 48.73(0.13) & 35.33(0.004) & 71.83 & -- \\
\textbf{AdalFlow-M (Promptwise)} & 85.00(1.47) & \textbf{91.67(0.24)} & \textbf{90.78(1.03)} & \textbf{82.77(0.56)} & \textbf{48.91(0.69)} & \textbf{36.67(0.13)} & 72.63 & \textbf{+0.80} \\
\midrule
GEPA & 84.33(0.58) & 91.67(0.58) & 91.55(0.39) & 81.5(0.09) & 47.6(0.04) & 36.00(0.82) & 72.11 & -- \\
\bottomrule
\end{tabular}
\end{table*}
\vspace{-1mm}
\section{Experiments}
\label{sec:experiments}

\subsection{Momentum Improves Scaling}
\label{sec:momentum_improves_scaling}

\begin{figure}[ht]
\centering
\includegraphics[width=0.8\linewidth]
{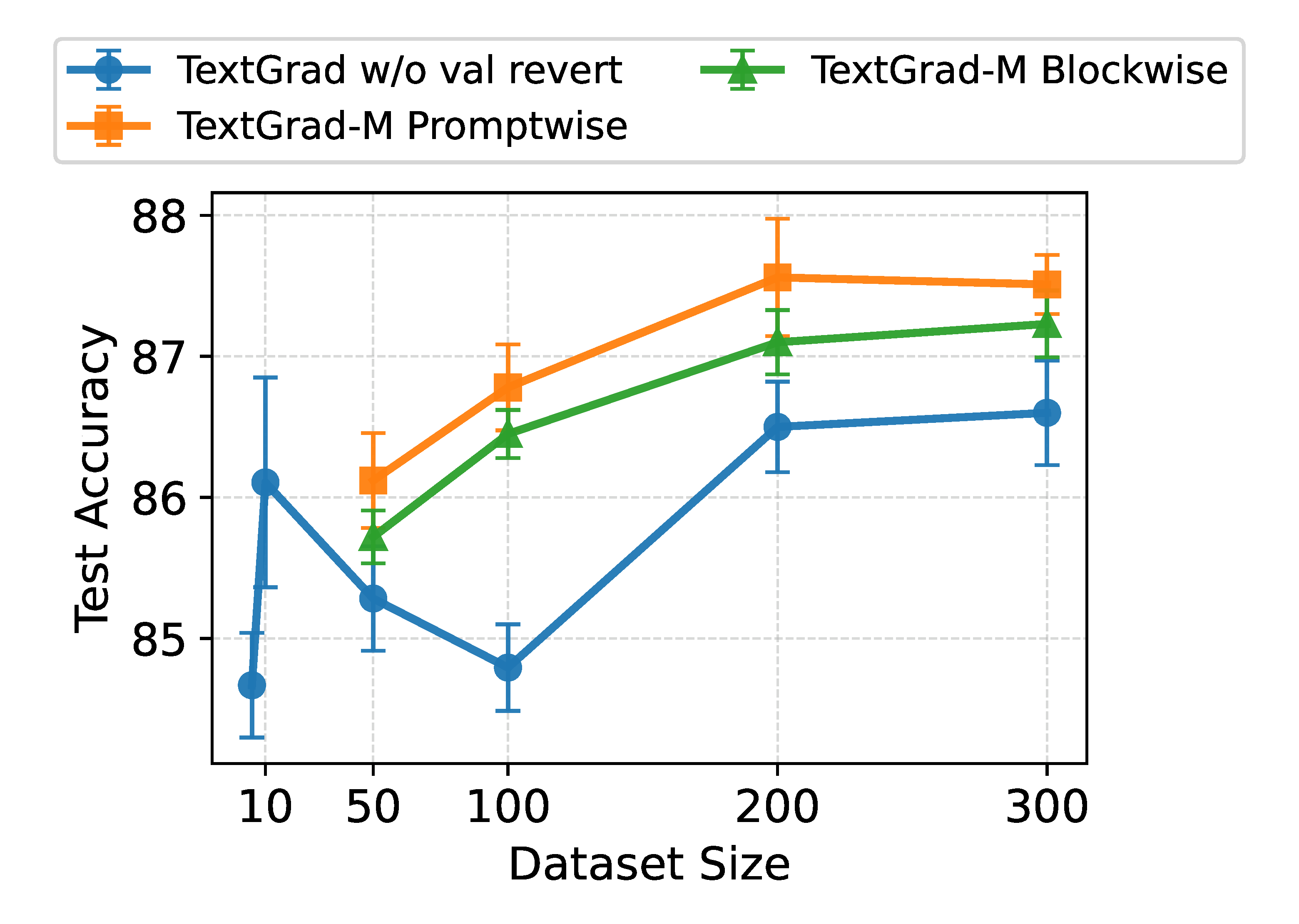}
\vspace{-3mm}
\caption{Upon scaling training data, \textgrad-M outperforms \textgrad\ on the MATH task with a batch size of 5.}
\label{fig:scale}
\end{figure}

Following \secref{sec:scaling}, we conduct scaling experiments on MATH, by increasing the training set size while keeping the batch size fixed at 5 and the number of epochs at 2. \cref{fig:scale} demonstrates our momentum-based module achieves consistently higher test accuracy than vanilla \textgrad\ across a wide range of training dataset sizes, as \llmforward\ benefits from a larger pool of training set and \llmbackward\ adapts to more diverse generated predictions.
Our method exhibits a smoother curve under scaling. Note  we do not plot out training dataset size less than 50 to highlight the generalization capability of our momentum method under the setting of scaling training dataset size.
The two variants of momentum achieve similar performance, while promptwise is slightly better.

Except for the average accuracy, the variance of test accuracy is remarkably reduced with the proposed momentum method, compared to \textgrad, which corroborates our theoretical intuition. For the \textit{Promptwise} generation, the variance drop compared to TextGrad is most significant as $0.37\%$, at 300 samples.
This reduction mirrors the principle behind bootstrap aggregation \cite{breiman1996bagging}, where repeated resampling stabilizes estimators and lowers variance. 

\vspace{-0.5em}

\subsection{Momentum Improves TSGD Variants}

Our method can be plugged into TSGD and its variants. We benchmark three representative TSGD variants and one state-of-the-art baseline, \emph{GEPA}~\citep{agrawal2025gepa}.
\emph{TextGrad} \citep{yuksekgonul2024textgrad} is TSGD by textual-gradient driven learning from examples. We use two setups: one with validation revert and one without.
\emph{COPRO} from DSPy \citep{dspy2025optimizers} iteratively sample new prompts based on validation accuracy. \emph{AdalFlow}~\citep{yin2025llm} refines prompts via \llmbackward\ applied only to error examples. We use \textbf{-M} to denote the momentum extension to these TSGD variants. We also include \emph{GEPA}~\citep{agrawal2025gepa} as a baseline, which similarly searches over a set of Pareto-front candidates.

\textbf{Tasks.}
Following prior work, \citep{sordoni2023joint, yuksekgonul2024textgrad, khattab2024dspy, clark2018think}, we assemble a diverse set of tasks. 
\textbf{Text Classification}: TREC~\cite{lu2022fantastically} is a natural-language understanding benchmark for question-type classification.
\textbf{Math}: \emph{GSM8K}~\cite{khattab2024dspy, yuksekgonul2024textgrad} is a set of grade school math problems for evaluating LLM capabilities in reasoning. We use the same evaluation metric as TextGrad, a string-based exact match metric to quantify accuracy.
\emph{MATH} (algebra) \cite{hendrycks2measuring} is a harder set of math problems.
Here, we use an algebra subset according to DSPy~\cite{dspy_math_tutorial}.
\textbf{Reasoning}:
\emph{ARC-Challenge}~\cite{clark2018think} is a benchmark consisting of grade-school-level multiple-choice science QA. 
\textbf{Long-Form Generation and Retrieval}: \emph{HotPotQA}~\cite{yang2018hotpotqa} is a large-scale Wikipedia QA benchmark that requires LLMs to understand context materials before answering questions. We adopt the full-wiki setting. \textbf{Instruction Following Benchmark}: IFBench~\citep{pyatkingeneralizing} is designed to assess LLMs’
ability to follow precise human instructions.
Data splitting, evaluation metrics, and baseline details are deferred in Appendix~\ref{app:exp_details}.

\textbf{Training Setups.}
If not otherwise specified, the accuracy over the full validation set is evaluated every four iterations, and we report the test accuracy on the prompts with the highest full validation accuracy.
Both TextGrad and Adalflow use a shuffled mini-batch with a batch size of $5$ and $2$ epochs.
Since COPRO does not use training data to obtain textual gradients, we apply the same mini-batch strategy for the validation.
Unless otherwise noted, the momentum window size is $5$.

\textbf{Momentum Boosts TSGD.} In \cref{tab:benchmark}, we report results on 5 tasks (with runtime analysis in Appendix~\ref{appendix:cost_analysis}). Across all TSGD variants, momentum is able to improve the performance for most tasks.
The improvement is most significant and consistent on TextGrad with gradients.
On hard tasks like MATH and HotPotQA, TextGrad-M has non-trivial improvements varying from 1.36\% to 1.60\%.
TextGrad with validation revert is purely \textit{exploitative}: with temperature 0, it collapses the action distribution to a single prompt, reducing variance between different seeds of trials but increasing bias.
In contrast, our method strikes a balance in exploration and exploitation over \textit{historical} prompts, thereby improving TextGrad.

The boosting is also observed for framework handling different components in LLM systems (AdalFlow). Even though AdalFlow introduces prior top prompts and failed prompts history as a flavor of exploration in the LLM's input context, AdalFlow-M introduces further boost in downstream performance by additional exploration and exploitation via momentum sampling (as observed in performance improvement on 4 over 5 tasks). Among all TSGD variants, COPRO does not have explicit ``textual gradients'' but directly sample next prompts given prior prompts. Even in such a gradient-free setting, our momentum method still improves the method for most tasks, showcasing the generality of our method. An exception occurs in MATH, where the initial DSPy signature prompt is already near-optimal, leaving minimal space for exploration, indicating task-specific saturation rather than our method's failure.

\begin{figure}[h]
\centering
\includegraphics[width=0.8\linewidth]{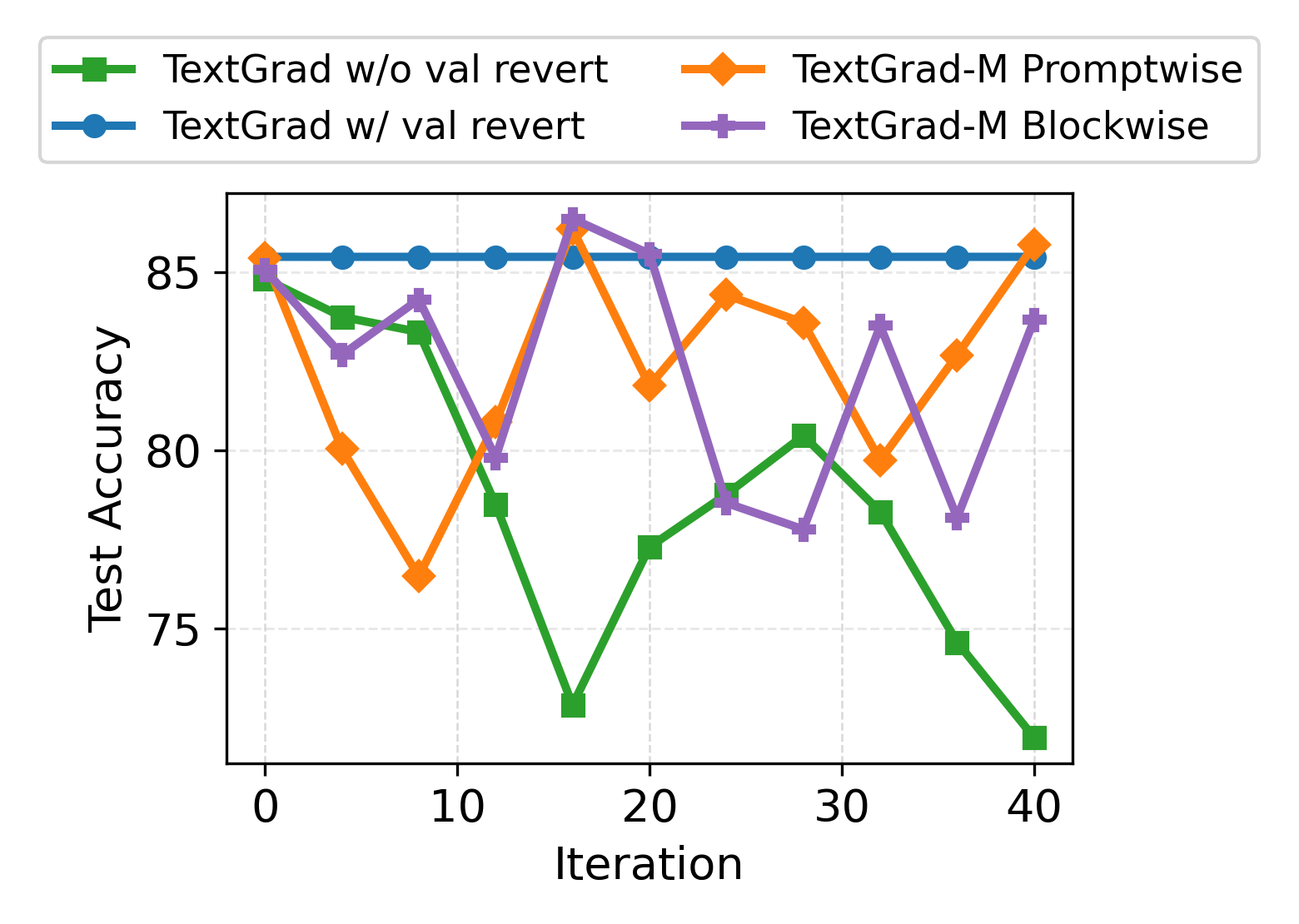}
\vspace{-3mm}
\caption{Average test accuracy over iterations on MATH. 
Without validation revert, TextGrad exhibits a declining trend; with validation revert, performance stagnates due to insufficient exploration of distinct prompts.}
\label{fig:MATH}
\end{figure}

In \cref{fig:MATH}, we analyze training dynamics over iterations.
Validation revert confines TextGrad to a narrow search region and prevents discovery of better prompts.
In contrast, \textgrad-M
avoids this collapse, allowing both \textit{Promptwise} and \textit{Blockwise} variants to attain higher accuracy with a more stable optimization trajectory.

\vspace{-0.5em}

\begin{figure*}[h]
  \centering
  \subfloat[ARC]{
    \includegraphics[width=0.26\textwidth]{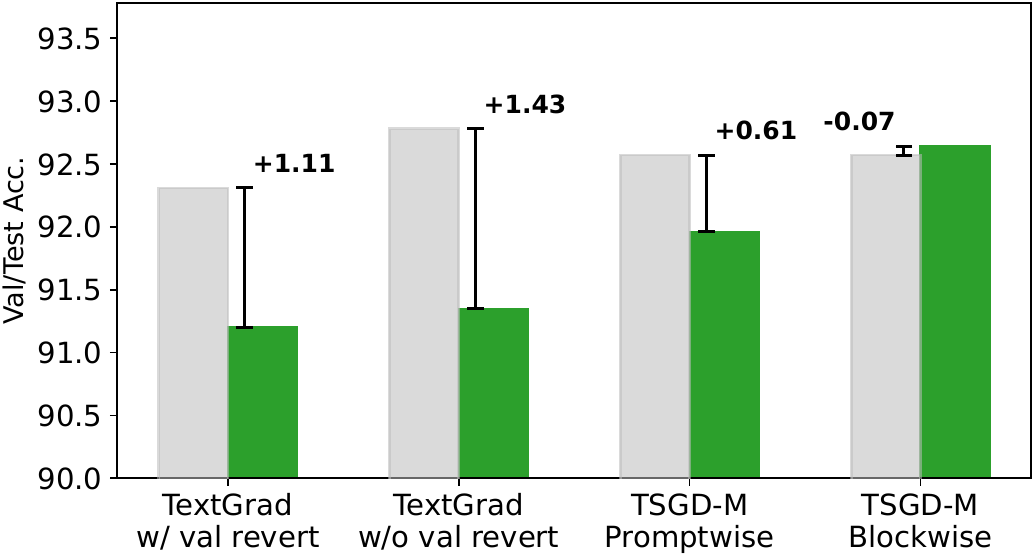}
  }\qquad
  \subfloat[GSM8K]{
    \includegraphics[width=0.26\textwidth]{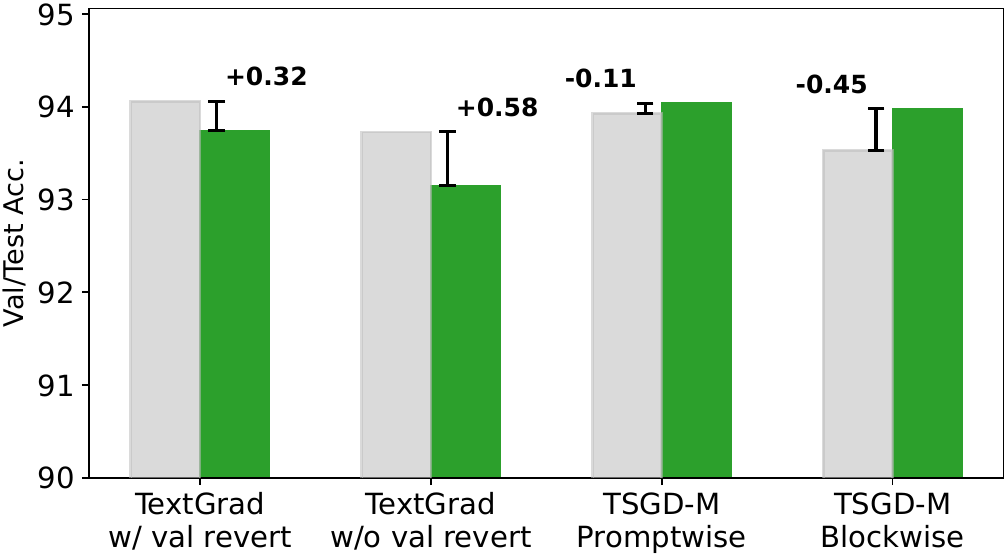}
  }\qquad
  \subfloat[MATH]{
    \includegraphics[width=0.26\textwidth]{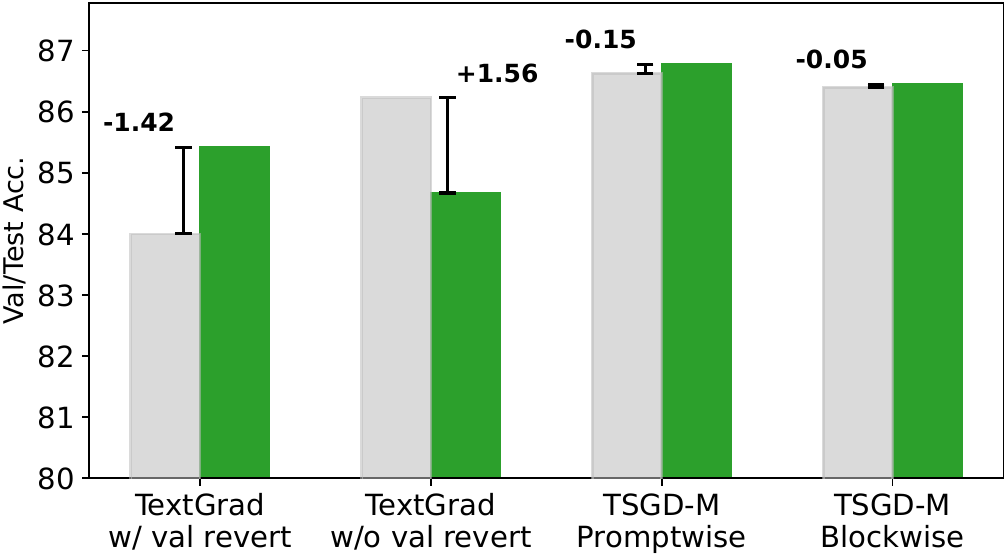}
  }
  \vspace{-3mm}
  \caption{Performance of \textgrad\ (w/ and w/o validation-revert) vs. \alg\ (Promptwise, Blockwise). All runs use training size 100. We report \textcolor{valgray}{validation} and \textcolor{testgreen}{test} accuracy for each module and calculate out the difference between them respectively.}
  \label{fig:generalization_gap}
\end{figure*}

\begin{table*}[h]
\centering
\caption{Ablations on evaluation scope, validation estimator, selection rule and exploration setting for \textsc{MATH}. Mini denotes minibatch validation. Val denotes validation set. Gumbel-Max denotes Gumbel-$k$ when $k=1$. Mini (mean) denotes running mean of minibatch validation accuracy.}
\label{tab:ablation_on_modules}

\scriptsize
\setlength{\tabcolsep}{4.0pt}
\renewcommand{\arraystretch}{0.9}

\begin{tabularx}{\linewidth}{l c c c c >{\raggedleft\arraybackslash}X}
\toprule
\textbf{Method} &
\textbf{Val Evaluation Scope} &
\textbf{Val Estimator} &
\textbf{Selection Rule} &
\textbf{Exploration Setting} &
\textbf{Val/Test (\%)} \\
\midrule
\tg\ w/o validation revert
    & Full
    & Argmax (Deterministic)
    & Argmax
    & None (Pure Exploitation)
    & 86.00 / 84.67 \\

Exploration
    & Full
    & Full (Gumbel-Max)
    & Argmax
    & \textbf{Exploration in Gumbel-Max}
    & 83.71 / 83.37 \\

Efficient evaluation
    & Mini
    & \textbf{Single mini val}
    & Gumbel-Max
    & Exploration in Gumbel-Max
    & 86.29 / 85.63 \\

Stable Evaluation
    & Mini
    & \textbf{Running mean}
    & Gumbel-Max
    & Exploration in Gumbel-Max
    & 85.00 / 84.86 \\

Explorative Evaluation
    & Mini
    & Running mean
    & Gumbel-$k$
    & \textbf{Gumbel-Max over Gumbel-$k$($K{=}5$)}
    & 85.43 / 85.30 \\

\textbf{Ours (exploration and exploitation)}
    & Mini
    & Running mean
    & Gumbel-$k$
    & \textbf{Argmax over Gumbel-$k$($K{=}5$)}
    & \textbf{86.63 / 86.78} \\
\bottomrule
\end{tabularx}

\end{table*}
\vspace{-1.mm}
\subsection{Ablation Studies}
\label{sec:ablation_study}
\looseness -1 \textbf{\textgrad-M \, outperforms TextGrad-Momentum with varying window sizes.} In vanilla \textgrad, momentum concatenates the prompts from the previous $K$ \emph{consecutive} iterations and feeds this window to \llmbackward to propose the next update. The impact of $K$ was not systematically examined in the original work \cite{yuksekgonul2024textgrad}. In contrast, our momentum variant samples the top $K$ prompts via momentum sampling, so the selected prompts need not be contiguous in time. 
\cref{tab:window_selection} and \cref{tab:window_selection_validation} show that \textgrad-M consistently improves both validation and test accuracy over TextGrad-Momentum.
\textbf{(i) Robustness to window size.}
Our method is relatively insensitive to $K$, though larger values generally improve validation accuracy. 
$K{=}40$ with \emph{Promptwise} yields the best validation accuracy, while $K{=}12$ attains the best test accuracy, though longer windows increase overhead (cost scales roughly with evaluation frequency and minibatch size, $\tilde{O}(|v|\cdot T)$, and larger $K$ adds cost of inferencing). 
For main experiments we adopt $K{=}5$ as a strong accuracy–cost trade-off. 
\textbf{(ii) Scaling beyond context-length limitations.} For vanilla \textgrad, increasing the \emph{concatenation} window does not
yield monotonic gains and can even degrade performance (e.g., Window $=5$ to $12$).
We hypothesize that longer input context degrade summarization capabilities due to long-context effects~\citep{liu2024lost, du-etal-2025-context}, suggesting that simply concatenating past prompts does not necessarily improve performance and the optimal context length varies by tasks. 

\textbf{Explore-then-exploit outperforms pure exploitation.}
Using \textgrad-M as a representative instance of \alg, we ablate its key components (\cref{alg:tsgd-momentum}) under the Promptwise setting. We vary: 
(i) the \emph{val evaluation scope} (full vs.\ minibatch of validation set), 
(ii) the \emph{val estimator} (single minibatch vs.\ a running mean over minibatches),
and (iii) the \emph{selection rule} (Gumbel-Max vs.\ Gumbel-$k$) together with the exploitation step (argmax within the chosen set on the \emph{same} minibatch for fairness), (iv) the \emph{exploration setting} (exploration vs.\ exploitation within the whole module).
\cref{tab:ablation_on_modules} shows: 
(1) Replacing full-validation evaluation (as in the second row, with validation and test accuracies of 83.71 \% and 83.37 \%) with repeated minibatch validation (rows four to six, averaging above 85 \% on validation and 84 \% on test) leads to improved generalization. Evaluating on small, randomly drawn minibatches introduces stochasticity analogous to bootstrap resampling, which effectively reduces estimator variance and mitigates overfitting to a single validation partition.
(2) With the running-mean estimator, Gumbel-$k$ outperforms Gumbel-Max by enabling exploration within the top $K$ performing set of prompts as visualized by average test accuracy improved from 84.86 to 85.30.
(3) After selecting the top $K$ candidates, exploiting by taking the argmax within the same minibatch yields an additional performance gain of approximately $\gtrsim\!1\%$ on both validation and test set. This improvement confirms our theoretical proof that Gumbel-Max over $K$ shortlisted candidates is Bayes-suboptimal to the empirical argmax on a shared minibatch (See \secref{Appendix:theoretic_justification}).

\textbf{Momentum improves generalization.} In \cref{fig:generalization_gap}, \textgrad-M ranks top 2 on validation and best on test. Momentum improves both metrics by balancing exploration (via Gumbel-$K$) and exploitation (Promptwise: generate from the top 1 on the same minibatch for fair comparison; Blockwise: uniformly sample from the top $K$). In contrast, \textgrad\ progressively picks the argmax over the full validation set. The sensitivity analysis in Appendix~\ref{appendix:sensitivity_analysis} showcases TSGD-M is also robust to minibatch validation size.
\section{Conclusion}
We present a systematic approach to scalable, effective automatic prompt tuning. 
We empirically visit the scaling law for textual gradient stochastic descent (TSGD) and identify the key challenge is to efficiently use minibatch while maintaining high stability.
Therefore, we propose a novel method, TSGD-Momentum, that reuses past prompts (and gradients) as a mixed distribution to sample new prompts.
Our method is modular and integrates seamlessly with existing frameworks (e.g., \textgrad, COPRO, AdalFlow) and empirically improves their performance on multiple tasks, including classification, reasoning and long-form generation.

\begin{acks}
This work was supported in part by the National Science Foundation under Grant No. IIS 2313131 and CMMI 2037026.
\end{acks}

\bibliographystyle{ACM-Reference-Format}
\bibliography{main}

\clearpage
\appendix

\section{Extended Related Works}
\label{Appendix: extended related work}
\subsection{Automatic Prompt Engineering Workflow}
\label{Appendix: APE Related Work}
We will revisit several \textbf{Automatic Prompt Engineering} frameworks below.

1. \textbf{APE} \citep{zhou2022large} is a seminal work in leveraging LLMs for instruction optimization. In each iteration, a set of instructions is evaluated on a validation set, and the optimizer generates a new set by paraphrasing the highest-performing instructions. This iterative process continues until convergence. However, we argue that APE does not fall under the category of Textual Gradient Descent (TGD) but instead aligns more closely with evolutionary algorithms \citep{yu2010introduction}, as it is inherently gradient-free. Rather than utilizing textual gradients for optimization, APE explicitly prompts LLMs to generate variations of instructions while preserving their semantic meaning, replacing lower-performing prompts through mechanisms akin to random variation (e.g., mutation or crossover), a hallmark of evolutionary strategies. Therefore, we exclude it for our evaluation.
\\
2. \textbf{DLN1} \citep{sordoni2023joint} views prompt optimization as learning a distribution $p_{LM}(y|x, \pi)$ in which $x, y$ the inputs or outputs are learned separately, and $\pi$ is learnable prompt. The iterative process is similar to APE but can include a verbalization of difficult examples from the task: the final prompts shall combine both instructions and task examples, which mimic a mix of zero-shot learning and in-context learning.
\\
3. \textbf{OPRO} \citep{yang2024large} optimizes instructions by showing the LLM a meta-prompt that includes the trajectory of previously generated prompts with their training accuracies, plus randomly sampled demonstrations that specify the task. To respect context-length limits, the meta-prompt retains only the highest-scoring instructions. Each iteration then asks the LLM to propose one additional candidate prompt. Compared to DLN1 \citep{sordoni2023joint}, OPRO typically runs for a substantially longer optimization horizon (approximately 100 steps per task); DLN1 can be viewed as a shorter iterations variant of \citet{yang2024large}. Moreover, OPRO requires hand-crafted refinement prompts (
$\prompt_{\text{refine}}$) tailored to different LLM optimizers, even for the same task (e.g., distinct templates for PaLM-2-L and GPT models on GSM8K; see Appendix C.2 of \citealt{yang2024large}). Template design can materially change the produced prompts and the optimization dynamics \citep{anthropic2025context}. Because constructing and comparing meta-prompt templates is beyond our scope for our paper, and given OPRO’s procedural similarity to DLN1 but with longer iterations and more specific $\prompt_{\text{refine}}$ templates design, we fix $\prompt_{\text{refine}}$ for all three baselines (TextGrad, DSPy-COPRO, DLN1) and adopt DLN1 as the representative instruction-optimization baseline when evaluating our TSGD-M algorithm.
\\
4. \textbf{TextGrad} \citep{yuksekgonul2024textgrad} backpropogates textual feedback provided by the proposal and view the textual feedback as gradients to perform descent or improve upon. For every iteration, they randomly extract several demonstrations and generate only one new prompt. They also present a momentum version by simply concatenating previously generated past gradients within certain window length.
\\
5. \textbf{DSPy} \citep{khattab2024dspy}. As we limit our study into zero-shot prompt optimization, in which we solely focus on \textit{instruction optimization} rather than \textit{example optimization} or jointly optimize both of them \citep{wan2024teach, opsahl2024optimizing}, we only discuss COPRO module in \citet{khattab2024dspy}. As our tasks are APE with zero-shot demonstrations needed to optimize, we use COPRO for automatic instruction optimization and exclude MIPROv2 as our baselines do not involve optimizing the set of few shots demonstrations and do not treat prompt generation temperature as a hyperparameter to optimize during iterative prompt optimization. Similar to DLN-1, COPRO leverages Signatures (structured prompts) to optimize Signatures themselves. We refer readers for further discussions on different optimizers \citep{dspy2025optimizers}.
\\
6. \textbf{PromptAgent} \citep{wang2024promptagent} views prompt optimization as a more advanced planning agent using Monte Carlo Tree Search (MCTS). We argue that \citet{wang2024promptagent} does not fall under TGD framework also. The MCTS algorithm itself is not a gradient-based algorithm as it relies on a search-based approach rather than differentiable optimization techniques, and MCTS does not compute or apply gradients. Even though MCTS shall be combined with gradient-based learning, where a policy network is trained using policy gradients and used to guide tree search, it is beyond our paper's research scope. Thus, we exclude this method.
\\
7. \textbf{GEPA} \citep{agrawal2025gepa} shares a similar spirit in exploring the top-performing prompts for each problem instance stochastically rather than optimizing a single global prompt as in TextGrad \cite{yuksekgonul2024textgrad}; it frames search over a set of Pareto-front candidates. \emph{However, our method differs in three key ways}:

\textbf{Who teaches the next generation.}
GEPA proposes new candidates by mutating existing ones using learning signals from their \emph{parents} and from the current rollout along a genetic tree. In contrast, we \emph{look back over the entire search history}: at each iteration we select supervision from the top–$K$ prompts across all previously generated candidates (sampled via the Gumbel-Top-K trick), not just from an immediate parent of the previous prompt. This history-wide selection provides stronger and more stable learning signals.

\textbf{How candidates are evaluated.}
\textsc{GEPA} scores every newly proposed prompt on the \emph{full} validation set each iteration, incurring high inference cost. We instead use lightweight \emph{mini-batch} evaluations to guide selection and generation, eliminating full-set scoring for every candidate. The full-validation sweep is run only for choosing the final best prompt.

\textbf{When a candidate is accepted.}
GEPA retains the TextGrad-style gate: a newly generated prompt is run on the full validation set and accepted only if it improves validation accuracy; otherwise it is rejected. Our approach \emph{removes} this full-set acceptance check. Even when evolving from a temporarily non-optimal prompt, we observe that sampling next-iteration seeds from the historical top–$K$ reliably yields improved prompts, as reflected by validation accuracy over time, while substantially reducing inference cost.

Taken together, these choices let us (i) exploit stronger signals than parent-only modification, (ii) avoid costly full-validation sweeps during search per iteration, and (iii) maintain progression of optimizing prompts even when the current incumbent is not on the estimated Pareto front.

\subsection{Relation to Prior Work}

Our method bears certain similarities to prior approaches, yet their formulations exhibit inherent limitations.

\textbf{Validation Revert.}
In the limit $K = 1$ and zero Gumbel noise, \alg \, reduces exactly to validation-revert.
\emph{Validation-revert} policy was used in prior work like TextGrad~\cite{yuksekgonul2024textgrad} or GEPA~\cite{agrawal2025gepa}, where the proposed prompt will be reverted to a prior iteration if its validation accuracy falls behind.
We argue that the validation reverting purely exploits the prompt with the highest (noisy) validation score but lacks exploring top-performing variants.
As a result, it can be easily trapped in local optima and does not necessarily yield the best generalization. Aggressively exploiting local optima tends to increase bias as the best prompt in the validation set is not necessarily the prompt generalizing best in the test set (See Fig~\ref{fig:val_test_3dataset} in Appendix~\ref{section:validation_vs._test accuracy}).
In contrast, \alg \, uses sampling-based methods to balance exploration and exploitation. (i) Sampling a Top-$k$ set injects stochastic exploration while retaining non-zero probability toward high-performing rather than  argmax prompts on the validation set. 
 (ii) By evaluating $k$ candidates per step, we obtain multiple, independent minibatch estimates of validation performance, improving the robustness of the selection signal and reduces estimator bias. 
This echoes the bootstrap idea~\cite{breiman1996bagging}: repeated evaluation over subsampled validation sets stabilizes estimates of a prompt’s full-validation accuracy. 

\textbf{TextGrad Momentum.}
\alg \, performs a \emph{dynamic}, probabilistic selection of $K$ prompts from the history: candidates are sampled according to weights derived from minibatch validation estimates and then used as the input context to \llmbackward. TextGrad \cite{yuksekgonul2024textgrad} first introduces momentum idea by concatenating the previous $K$ prompts and feeding them to \llmbackward\ to generate the next prompt. Both algorithms share the same name momentum and are exploring the history of past prompts. However, \alg \, avoids the rigidity of a fixed, most-recent $K$ window as vanilla \textgrad\, which might not contain the most informative prompts for the next iteration. 
By stochastically favoring historically high-value prompts (as indicated by their running mean) while still allowing exploration, \alg \, adapts to non-stationarity and mitigates stale-history effects. 
We study the choice of $K$ in our \alg \, and compare it with \textgrad \, momentum in our ablation study Table~\ref{tab:window_selection} (Section~\ref{sec:ablation_study}).

\section{How Minibatch Size $m$ Shapes Textual Gradient Quality}
\label{appendix:minibatch_size_textual_gradients}
Here, we examine how minibatch size $m$ affects the \emph{quality} of textual gradients on the MATH (algebra) subset, using \(\llmforward\) = GPT-4o-mini and $\llmbackward$ = GPT-4o. For the purpose of investigating textual gradients quality under the \emph{scaling} setting, we fix the underlying training set to \(N=350\) examples so that the optimizer’s candidate prompts are evaluated against the full pool of available supervision, and vary the per-iteration minibatch size \(m \in \{5,10,100\}\) used to generate gradient feedback. 

Figures~\ref{fig:smaller_minibatch_textual_gradient}--\ref{fig:larger_minibatch_textual_gradient} compare representative textual gradients across settings, illustrating how \(m\) controls (i) the \textbf{concision vs.\ coverage} trade-off (small \(m\) yields shorter, more localized feedback in Fig~\ref{fig:smaller_minibatch_textual_gradient}), (ii) the probability of \textbf{degenerate/empty gradients} when a sampled minibatch is already fully correct (See Fig~\ref{fig:smaller_minibatch_textual_gradient}), and (iii) the tendency of large \(m\) to produce \textbf{longer, multi-issue, and occasionally inconsistent} feedback that can be harder to apply reliably in downstream APE updates (See Fig~\ref{fig:larger_minibatch_textual_gradient}). Figure~\ref{fig:middle_minibatch_textual_gradient} illustrates that a medium minibatch ($m=10$) yields a more balanced textual gradient, providing actionable and specific prompt-editing suggestions. In our MATH (algebra) setting with $\llmforward$= GPT-4o-mini and 
$\llmbackward$= GPT-4o, $m = 10$ is the best-performing choice. However, the optimal minibatch size is highly setting-dependent and generally cannot be predicted \emph{a priori}; identifying it typically requires extensive sweeps, which can be computationally prohibitive without large-scale experimentation.

\newcommand{\cmt}[1]{
  \par\smallskip
  {\footnotesize\textcolor{blue!70}{\textbf{Comment:} #1}}\par\smallskip
}

\begin{figure*}[htbp]
\begin{framed}
\noindent\textbf{Sample Textual Gradients when minibatch size \(m=5\)}\\[1ex]

\small
Textual Gradient Sample \#1: The current prompt is commendably strict in its instructions regarding output format and the exclusion of extraneous text. However, the language model's output still includes a redundant natural language statement of the answer ("The number of vertical asymptotes is 1.") immediately preceding the strictly formatted boxed answer. \hlblue{This indicates that while the prompt's negative constraints are present, they are not sufficiently robust or salient to prevent.} \cmt{With only one error example, the gradient is necessarily generic and lacks actionable, instance-specific edits.}
 
Textual Gradient Sample \#2:
The current variable already perfectly satisfies the objective function, as indicated by the `1` output from the evaluation function. \hlblue{No feedback is needed to improve the metric.}
\cmt{All sampled examples in the minibatch are correct, yielding an (almost) empty gradient signal.}

\end{framed}
\caption{Smaller minibatches tend to yield more succinct textual gradients; when a sampled minibatch is already fully correct, the gradient becomes minimal for potential improvement (or empty).}
\label{fig:smaller_minibatch_textual_gradient}
\end{figure*}

\begin{figure*}[htbp]
\begin{framed}
\noindent\textbf{Sample Textual Gradients when minibatch size \(m=10\)}\\[1ex]

\small
Textual Gradient Sample \#1:
The current system prompt is highly effective at enforcing output formatting and conciseness, as evidenced by the LM's bare numerical output. However, the core issue is a mathematical error in the calculation itself (outputting `-1` instead of `1`). The prompt's extensive focus on *how* to output might inadvertently de-emphasize the paramount importance of *mathematical correctness*.

Here are strategies to improve the variable:

1.  **Explicitly Elevate Mathematical Accuracy as the Foremost Priority:** While the prompt mentions "expert" and "accuracy," the sheer volume and intensity of formatting constraints (bolding, ALL CAPS, repetition) might make them the most salient features for the LM. Re-prioritize by explicitly stating that **mathematical correctness is the absolute highest objective**, even above conciseness or formatting if there's any perceived internal conflict. \hlblue{For example, add a statement like: "Your **ABSOLUTE PRIMARY DIRECTIVE** is
the **UNQUESTIONABLE MATHEMATICAL CORRECTNESS** of the final result.
All formatting and conciseness rules are secondary to ensuring this correctness."}
\cmt{More explicit, instance-specific textual edits.}

2.  **Instruct for Internal Verification and Self-Correction:** Since the LM is forbidden from showing its work, it must perform all verification internally. The prompt should explicitly instruct the LM to internally double-check its calculations, algebraic manipulations, and final simplification steps *before* producing the output. This can be framed as a crucial strategy for an "expert" to guarantee accuracy. For instance: "As an expert, you are expected to **internally verify and rigorously double-check all calculations and algebraic steps** before arriving at your final answer. This internal self-correction mechanism is vital for achieving absolute accuracy."

3.  **Clarify the Meaning of "Expertise":** \hlblue{The prompt states, "Your expertise is demonstrated through the accuracy and conciseness of the final result, not by showing your work." This could be refined to emphasize that *accuracy is the primary component of expertise*. For example: "Your expertise is **first and foremost demonstrated by the absolute mathematical accuracy** of the final result, and then by its conciseness and adherence to output specifications. The absence of shown work necessitates flawless internal computation."}
\cmt{More explicit actionable guidance for textual gradient.}

4.  **Reinforce Rigor in Calculation:** Add a specific directive that emphasizes the rigor required for the mathematical computation itself. This ensures the LM understands that "completing all calculations" means doing so with utmost precision. For example: "You **MUST apply rigorous mathematical principles and meticulous calculation** to derive the definitive, final answer. No approximation or estimation is acceptable; only exact results.
\cmt{More instance-specific prompt editing suggestion.}

\end{framed}
\caption{Example textual gradient produced with minibatch size \(m=10\). Medium minibatches often yield more detailed, actionable feedback, grounded in specific failure cases, than very small minibatches.}
\label{fig:middle_minibatch_textual_gradient}
\end{figure*}

\begin{figure*}[htbp]
\begin{framed}
\noindent\textbf{Sample Textual Gradients when minibatch size \(m=100\)}\\[1ex]

\small
Textual Gradient Sample \#1 at \hlred{iteration 2}:
\\
The current variable, the system prompt, is critically incomplete and lacks specific instructions for the "Final Answer" section, which is the primary reason the LM\_OUTPUT fails to meet the objective function.
Here's how the variable can be improved:
1.  \hlblue{**Complete the "Thought Process" Section Definition:** The variable currently ends abruptly with "This content will be". This needs to be completed to form a grammatically sound and clear instruction. For example, it could be "This content will be detailed enough to show your work and reasoning."}
\cmt{With a large minibatch $m = 100$, \llmbackward aggregates multiple failure modes and produces a long, multi-part edit plan (formatting, section definitions, and evaluation constraints), increasing gradient length and implementation burden.}
2.  **Define the "Final Answer" Section with Precision:** The prompt mentions "Final Answer" but provides no instructions for it. This is the most significant omission. The prompt must clearly specify:
    *   **Content:** What exactly should be included in this section (e.g., "only the final numerical value", "the integer 'm'").
    *   **Format:** How it should be presented (e.g., "just the number", "Final Answer: [number]", "m = [number]"). Given the string-based evaluation, a very specific and consistent format is crucial for easy parsing.
    *   **Purpose:** Emphasize that this section is for the *final, verifiable result* that will be evaluated.
... 
\\
Textual Gradient Sample \# 2 at \hlred{iteration 4}:

The current system prompt, specifically the variable span, has an incomplete sentence: "This content will be". This grammatical error in the prompt itself could be a direct cause of the language model's output terminating prematurely and incompletely (e.g., "The result is a"). Fixing this grammatical error is crucial for the model to understand the full instruction.

To address the objective function's feedback regarding the incompleteness of the thought process and the missing final answer, the prompt should be modified in the following ways:

1.  \hlblue{**Complete the incomplete sentence in the prompt**:} The phrase "This content will be" needs to be completed to form a coherent instruction. For example, it could be \hlblue{"This content will be used to verify your solution."} or "This content will be evaluated for correctness and completeness." This ensures the instruction itself is clear and doesn't inadvertently signal an incomplete task to the model.
\cmt{The same fix resurfaces across iterations, consistent with prior observations that long-context optimization can weaken error retrieval and persistence \citep{du-etal-2025-context}.}

4.  \hlblue{**Reinforce the necessity of *all* calculation steps, including the final one**:} While "detailed enough to show your work" is present, it could be strengthened to explicitly include the *final step* of calculation that yields the numerical value. This would help prevent the model from stopping just before the ultimate numerical result is produced. Consider adding emphasis like: "Ensure *all* intermediate and final calculation steps are shown, leading directly to the final numerical value."
\cmt{Emphasizing the necessity to include both reasoning and answer.}
Textual Gradient Sample \# 3 at \hlred{iteration 6}:
The current prompt \hlblue{encourages verbosity and detailed explanations, which directly conflicts with the objective of reducing the length and complexity of the `LM\_OUTPUT` for faster string-based correctness checking.}
\cmt{Later gradients partially contradict earlier advice (promoting detailed reasoning vs.\ discouraging verbosity), illustrating a reversion mode where large minibatches induce competing objectives and unstable optimization directions.}
...
\end{framed}
\caption{Sample textual gradients with a large minibatch (\(m=100\)). Larger minibatches often yield longer, multi-issue feedback and can introduce instability across iterations, repeating earlier fixes and even reversing optimization direction (e.g., encouraging verbosity at iteration 2 and 4 vs.\ penalizing it) at iteration 6.}
\label{fig:larger_minibatch_textual_gradient}
\end{figure*}
\section{Extended Algorithm}
\label{section:extended_algorithm}
Algorithm~\ref{alg:gumbel-top-k} provides the complete implementation details for Algorithm~\ref{alg:tsgd-momentum} (\secref{Section:TSGD-M}).
\label{Extended_Algorithm}

\setlength{\textfloatsep}{6pt}
\setlength{\floatsep}{6pt}

\newcommand{\TSGDCommonPrefix}{
  \Require Language Model \llm, $p_{0}$, $\dataDist$, iterations $T$, batch size $m$,
  max tokens $T_{\max}$, candidates per iter $c$, score $S$, template $\prompt_{\text{refine}}$,
  window/Gumbel–Top-$K$ parameter $K$, minibatch val set size $a$, block size $b$, full val set $val$.
  \STATE Initialize $\tilde{Z} \gets \{ \prompt_{0}: [], \ldots, \prompt_{T}: [] \}$ \Comment{history of scores}
  \For{$\tau = 0,1,\dots,T-1$}
    \STATE Sample $\{(x_i^{(t)}, y_i^{(t)})\}_{i=1}^m \sim \dataDist$
    \STATE $\hat y_i^{(t)} \gets \llm([\prompt_\tau, x_i^{(t)}])$ for $i=1,\dots,m$
    \STATE Draw minibatch validation set $v_\tau \subset val$ with $|v_\tau|=a$
    \STATE Evaluate $\prompt_\tau$ on $v_\tau$ to get $v_{\tau,\tau}$; append to $\tilde{Z}[\prompt_\tau]$
    \If{$\tau \le K$}
      \STATE Evaluate $\prompt_{0:\tau}$ on $v_\tau$ to get $v_{\tau,0:\tau}$
      \STATE $\prompt_\tau^\star \gets \arg\max_{k\in\{0,\ldots,\tau\}} v_{\tau,k}$;\quad
             append $v_{\tau,k}$ to $\tilde{Z}[\prompt_k]$ for all $k\in[0,\tau]$
    \Else
      \STATE Apply Gumbel–Top-$K$ to averages in $\{\tilde{Z}[\prompt_k]\}_{k=1}^{\tau}$;
             obtain $\prompt_{\tau,1:K}$
      \STATE Evaluate $\prompt_{\tau,1:K}$ on $v_\tau$ to get $v_{\tau,1:K}$
      \STATE $\prompt_\tau^\star \gets \arg\max_{k\in\{1,\ldots,K\}} v_{\tau,k}$;\quad
             append $v_{\tau,k}$ to $\tilde{Z}[\prompt_{\tau,k}]$ for $k=1{:}K$
    \EndIf
    \STATE $Z \gets \emptyset$ \Comment{candidate pool for iteration $\tau$}
}%

\begin{algorithm}[h]
  \caption{Textual Stochastic Gradient Descent with Momentum}
  \label{alg:gumbel-top-k}
  \begin{algorithmic}[1]
    \REQUIRE Language Model \llm, $p_{0}:$ initial prompt, $\dataDist$: Data Distribution, $T:$ Total iterations, $m$: batch size, use Promptwise Generation flag, $T_{\max}$: max tokens, $c:$ number of candidate prompts to generate, $S:$ score function, $\prompt_{\text{refine}}$: Template to generate new prompts, $K:$ window size/Gumbel-Top-K parameter, $v$: minibatch validation set with size $a$, $b:$ Size of sampling token block, $val:$ original validation set.
    \STATE $\bar Z \gets \{\,\pi_\tau : []\,\}_{\tau=0}^{T}$ 
    \FOR{$\tau = 0, 1, \dots, T-1$}
            \STATE Sample batch $\{(x_i^{(t)}, y_i^{(t)})\}_{i=1}^m \sim \dataDist$
            \STATE $\hat y_i^{(t)} \gets \mathrm{LM}([\pi_\tau,x_i^{(t)}])$
            \STATE $\nabla \prompt_t \gets \llmbackward\bigl([p_{\text{analyze}}, \prompt_t,\{(x_i,y_i,\widehat y_i)\}_{i=1}^m]\bigr)$

            \STATE At iteration $\tau$, randomly draw minibatch validation set $v_{\tau}$ with size $a$ from validation set $val$
            \STATE Evaluate $\prompt_{\tau}$ on $v_{\tau}$ and obtain evaluated validation accuracy $v_{\tau, \tau}$ on $v_{\tau}$. 
            \STATE $\tilde{Z}[\prompt_{\tau}]\text{.append}(v_{\tau, \tau})$
            \STATE Apply Gumbel-Top-$k$ Trick on average scores of $\tilde{Z}$ and obtain $\Pi_{\tau}^{*} = \{ \prompt_{\tau, 1}, ..., \prompt_{\tau, K} \}$.
            \STATE Evaluate $ \Pi_{\tau}^{*}$ on $v_{\tau}$ and get $v_{\tau, 1}, ..., v_{\tau, K}$.
            \STATE $\prompt_{\tau}^{*} \leftarrow \operatorname*{arg\,max}_{k\in\{1,\ldots,K\}} \; v_{\tau,k}$ 
            \STATE $\tilde{Z}[\prompt_{k}]\text{.append}(v_{\tau, k})$ for $k \in [1, ..., K]$
            \IF{use Promptwise Generation}
            \STATE $\begin{aligned}
Z \gets \text{PromptwiseGen}(\pi_\tau^{*}, \nabla\pi_\tau^{*}, c, T_{\max}, \pi_{\text{refine}},\\ 
\pi_i, \nabla \pi_{i}, \{x_i^{(t)}, y_i^{(t)}, \widehat y_i^{(t)}\}_{i=1}^{m})
\end{aligned}$
            \ELSE
            \STATE $Z \leftarrow$ BlockwiseGen($\Pi_{\tau}^{*}$, $c$, $b$, $T_{\max}$, $K$, $\prompt_{\text{refine}}$, $\tau$, \\
            $\{ x_{i}^{(t)}, y_{i}^{(t)}, \hat{y}_{i}^{(t)} \}_{i = 1}^{m}$) 
            \ENDIF
            \STATE $\prompt_{t+1} \gets \operatorname*{arg\,max}\limits_{z \in Z} S(z)$
            \ENDFOR
\STATE \textbf{Output:} Optimized prompt $\prompt_{T-1}$
    \end{algorithmic}
\end{algorithm}

\begin{algorithm}[t]
\caption{PromptwiseGen$(\prompt_\tau^\star, \nabla \prompt_\tau^\star, c, T_{\max}, 
\\ \prompt_{\text{refine}}, \prompt_{i}, 
\nabla \prompt_{i}, \{ x_{i}^{(t)}, y_{i}^{(t)}, \hat{y}_{i}^{(t)} \}_{i = 1}^{m})$}
\label{alg:promptwisegen}
\begin{algorithmic}[1]
  \STATE $Z \gets \emptyset$
  \FOR{$j=1$ to $c$}
    \STATE $z_j \gets$ Generate $T_{\max}$ tokens using
       {\color[RGB]{19,99,223}$\llm(\prompt_{\text{refine}}+\prompt_{\tau}^{\star} +  \{ x_{i}^{(t)}, y_{i}^{(t)}, \hat{y}_{i}^{(t)} \}_{i = 1}^{m}$}
       \emph{(direct prompt refinement/textual gradients free)}, 
       \\
       or {\color[RGB]{236,114,114}$\llm(\prompt_{\text{refine}}+\prompt_{i}+\nabla \prompt_{i}+\prompt_{\tau}^{\star}+\nabla \prompt_{\tau}^{\star} + \{ x_{i}^{(t)}, y_{i}^{(t)}, \hat{y}_{i}^{(t)} \}_{i = 1}^{m})$}
       \emph{(textual gradients)}.
    \STATE $Z \gets Z \cup \{z_j\}$
  \ENDFOR
  \STATE \textbf{return} $Z$
\end{algorithmic}
\end{algorithm}

\begin{algorithm}[t]
\caption{BlockwiseGen$(\Pi_{\tau}^{*}, c, b, T_{\max}, K, \prompt_{\text{refine}}, 
\\
\tau, \{ x_{i}^{(t)}, y_{i}^{(t)}, \hat{y}_{i}^{(t)} \}_{i = 1}^{m})$}
\label{alg:blockwisegen}
\begin{algorithmic}[1]
  \STATE $Z \gets \emptyset$
  \FOR{$j=1$ to $c$}
    \STATE $z_j \gets \emptyset$
    \FOR{$i=1$ to $T_{\max}//b$}
      \STATE $\prompt_\tau^\star \sim 
      \mathrm{Unif} \!\big( \Pi_{\tau}^{*} \big)$
      \STATE Generate $b_i$ tokens using
        {\color[RGB]{19,99,223}$\llm(\prompt_{\text{refine}}+\prompt_{\tau}^{\star} + \{ x_{i}^{(t)}, y_{i}^{(t)}, \hat{y}_{i}^{(t)} \}_{i = 1}^{m})$} \emph{(direct prompt refinement/textual gradients free)},
        or {\color[RGB]{236,114,114}$\llm(\prompt_{\text{refine}}+\prompt_{i}+\nabla \prompt_{i}+\prompt_{\tau}^{\star}+\nabla \prompt_{\tau}^{\star} + \{ x_{i}^{(t)}, y_{i}^{(t)}, \hat{y}_{i}^{(t)} \}_{i = 1}^{m})$} \emph{(textual gradients)}.
      \STATE $z_j \gets z_j + [b_i]$
    \ENDFOR
    \STATE $Z \gets Z \cup \{z_j\}$
  \ENDFOR
  \STATE \textbf{return} $Z$
\end{algorithmic}
\end{algorithm}

\section{Theoretical Justification}
\label{Appendix:theoretic_justification}

Here, we show our momentum sampling (via Gumbel–Top-$k$) is not heuristic but theoretically grounded as sampling without replacement from a categorical distribution over validation scores.
\begin{proposition}[Gumbel–Top-$k~$\citep{kirschstochastic}]
\label{proposition:gumbel_top_k}
Given arbitrary real-valued scores $s_i$ ($i\in\{1,\dots,n\}$), $k\le n$, and $\beta>0$, if
$\epsilon_i \sim \mathrm{Gumbel}(0;\beta^{-1})$ i.i.d., then
\[
\operatorname*{arg\,top}_k \{s_i+\epsilon_i\}_i
\]
is an ordered sample without replacement from 
$\mathrm{Categorical}\!\left(\frac{e^{\beta s_i}}{\sum_j e^{\beta s_j}}\right)$.
\end{proposition}
\subsection{Proof for Promptwise Generation}
\label{Appendix:Proof_Promptwise}
We justify why, in Promptwise generation, it is Bayes-optimal to select the
prompt with the highest fresh-batch validation accuracy (``Fresh-Argmax'')
rather than to re-sample using Gumbel noise on the updated running means (Gumbel-Max trick on the k running means).
We formalize the selection step as a one-shot decision problem with linear utility and analyze it under a Bayesian framework.

Let $\Pi_{\tau}^{*}$ denote the size-$k$ shortlist returned by Gumbel-Top-$k$ on scores $\tilde{Z}$ at iteration $\tau$.
For each $\prompt_{\tau, i} \in \Pi_{\tau}^{*}$, let $r_{\tau,i}\in[0,1]$ be the (unknown) true validation accuracy on the downstream task.
On the current iteration we evaluate all $\prompt_{\tau, i} \in \Pi_{\tau}^{*}$ on the \emph{same} fresh mini-batch, obtaining
\[
\begin{aligned}
& v_{\tau,i} \;=\; r_{\tau,i} \,+\, \varepsilon_i, \\
\qquad &\text{with }~\{\varepsilon_i\}_{i\in \Pi_{\tau}^{*}}\text{ independent, mean-zero,}\\
&\text{and from a log-concave location family.}
\end{aligned}
\]
(e.g., Binomial $\to$ sub-Gaussian/Gaussian approximation with variance $r_i(1-r_i)/m$ when batch size is $m$).
Let $B_{i}$ denote all past data used to form the running means $\mu'_i$ (plain averages with sample counts $n_i$ as how many times $\prompt_{i}$ being selected into k shortlist candidates for fresh minibatch evaluation).
Within the $K$ items, we compare two rules for selecting the next prompt:
\begin{align*}
\textsf{A (Fresh-Argmax):}\quad  i_A &\in \arg\max_{k \in \{1, ... K\}} v_{\tau, k},\\
\textsf{B (Gumbel-on-Mean):} \quad i_B &\in \arg\max_{k \in \{1, ... K\}} \{\mu'_i + \xi_i\},
\\
\xi_i & \stackrel{iid}{\sim}\mathrm{Gumbel}(0,1).
\end{align*}
We measure performance by expected reward $\mathbb{E}[r_{i_\bullet}]$ (equivalently, minimizing expected regret).

\begin{theorem}[Bayes optimality of posterior-mean argmax under equal precision]
\label{thm:equal-precision}
Fix an iteration $\tau$ and the realized shortlist $\Pi_{\tau}^{*}$ of size $k$.
Assume
independent Gaussian posteriors with common variance
$r_{\tau,i}\mid B_i \sim \mathcal{N}(\mu_i,\tau^2)$
and a shared fresh mini-batch with homoscedastic Gaussian noise
$v_{\tau,i}\mid r_{\tau,i}\sim \mathcal{N}(r_{\tau,i},\sigma^2)$,
with the same $\tau^2,\sigma^2$ for all $i\in \Pi_{\tau}^{*}$.
Let
\begin{align}
m_i \;:=\; \mathbb{E}\!\left[r_{\tau,i}\,\middle|\,B_i, v_{\tau,i}\right]
\;=\; \alpha\,\mu_i + (1-\alpha)\,v_{\tau,i},
\\
\alpha := \frac{\tau^2}{\tau^2+\sigma^2}\in(0,1).
\end{align}
Then the Bayes-optimal rule for maximizing expected reward is
\[
i^\star \in \arg\max_{i\in \Pi_{\tau}^{*}} m_i.
\]
Moreover, the “Fresh-Argmax” rule $i_A \in \arg\max_{k \in \{1, ... K\}} v_{\tau,k}$ coincides with $i^\star$
\emph{if and only if} its maximizer $i_A$ satisfies the margin condition
\begin{equation}
\label{eq:margin-condition}
(1-\alpha)\,\big(v_{\tau,i_A}-v_{\tau,j}\big)
\;\ge\;
\alpha\,\big(\mu_j-\mu_{i_A}\big)
\qquad \text{for all } j\in \Pi_{\tau}^{*}.
\end{equation}
\end{theorem}

\begin{proof}
By conjugacy,
$m_i=\alpha\,\mu_i+(1-\alpha)\,v_{\tau,i}$ with the same $\alpha$ across candidates of $\Pi_{\tau}^{*}$. 
Thus the Bayes-optimal decision is $\arg\max_i m_i$.

For coincidence with Fresh-Argmax, note that
$i_A = \arg\max_{k \in \{1, .. K \}} v_{\tau,k}$ and $i^\star=\arg\max m_i$ coincide
iff $m_{i_A}\ge m_j$ for all $j$, i.e.
\[
\alpha\,(\mu_{i_A}-\mu_j) + (1-\alpha)\,(v_{\tau,i_A}-v_{\tau,j}) \;\ge\; 0
\quad \forall j.
\]
Rearranging yields~\eqref{eq:margin-condition}, which is both necessary and sufficient.
\end{proof}
\textbf{Note.} The $k$ shortlisted prompts $\Pi_{\tau}^{*}$ are drawn from the global pool via
Gumbel-Top-$k$ on their running means $\tilde{Z}$
, which stochastically perturbs and then ranks the existing performance estimates.
Conditional on selection, the shortlisted prompts therefore represent the
upper tail of the prior distribution of means~$\{\mu_i\}$.
Because Gumbel-Top-$k$ only adds mean-zero perturbations and truncates the top
region, the conditional spread of $\mu_i$ within $\Pi_{\tau}^{*}$ is
narrow compared with both the prior variance~$\tau^{2}$ and the fresh-batch
noise variance~$\sigma^{2}$, i.e.
$\operatorname{Var}(\mu_i\mid i\in\Pi_{\tau}^{*})\ll\tau^{2},\sigma^{2}$.
Hence the selected prompts are approximately exchangeable and may be modeled as
having \emph{near-equal prior means}, allowing the posterior-mean ordering
$m_i=\alpha\mu_i+(1-\alpha)v_{\tau,i}$ to coincide with the fresh-batch
ordering $\arg\max_{k \in \{1, ... K\}} v_{\tau,k}$.

\paragraph{Decision-theoretic setup.}
We model the selection of the next prompt as a single-step decision problem
with a linear utility function
\[
u(a,\omega) \;=\; r_{\tau,a}(\omega),
\]
the (unknown) true reward obtained by choosing action $a\in\mathcal{A}=\Pi_\tau^{*}$
under latent state $\omega$.
The goal is to maximize expected utility
$\mathbb{E}_{\pi,\delta}[u(a,\omega)]$
over all decision rules $\delta$ that map observable statistics
(e.g., $\mathcal X=(B_i,\{v_{\tau,j} \} _{j \in \Pi_{\tau}^{*}})$ or its coarsening $\mu'$)
to probability distributions over actions.
A linear utility corresponds to minimizing a convex loss (up to an affine transform),
and ensures that the Bayes-optimal rule can be taken to be deterministic.

\begin{theorem}[Information dominance of decisions using the fresh batch]
\label{thm:blackwell}
In the general (possibly heteroscedastic) case, let the decision-maker observe either the \emph{full} statistic $(B_{i},\{v_{\tau, j} \}_{j\in \Pi_{\tau}^{*}})$ or a coarsened statistic $T(B_{i},\{v_{\tau, j} \}_{j \in \Pi_{\tau}^{*}})=\mu'$, where $\mu'$ is the collection of updated running means for k shortlisted prompts at iteration $\tau$. In other words, $\mu' := \{ \mu_{i}' : i \in \Pi_{\tau}^{*} \}$ with $\mu_{i}' = \frac{n_{i} \cdot \mu_{i} + v_{\tau, i}}{n_{i} + 1}$. For any bounded linear reward (equivalently, convex loss), the Bayes risk using $(B_{i},\{v_{\tau, j} \}_{j \in \Pi_{\tau}^{*}})$ is no worse than that using $\mu'$.
Consequently, the best deterministic rule based on $(B_{i},\{v_{\tau, j} \}_{j \in \Pi_{\tau}^{*}} )$ weakly dominates any rule that relies only on $\mu'$ and independent randomization (such as Gumbel-Max on $\mu'$).
\end{theorem}

\begin{proof}
Fix an iteration $\tau$ and condition on the realized shortlist $\Pi_{\tau}^{*}$.
Let the action set be $\mathcal{A}:=\Pi_{\tau}^{*}$. Let
\[
\mathcal{X}\;:=\;(B_{i},\{v_{\tau,j}\}_{j\in \Pi_{\tau}^{*}})
\qquad\text{and}\qquad
\mathcal{T}(\mathcal{X})\;:=\;\mu'
\]
denote, respectively, the \emph{full} statistic and its coarsening into the updated running means.

A (possibly randomized) decision rule based on a statistic $Z$ is a Markov kernel
$\delta:\mathsf{Z}\to\Delta(\mathcal{A})$, mapping an observation $z$ to a distribution
$\delta(\cdot\,|z)$ over $\mathcal{A}$. Write $\mathcal{D}(Z)$ for the class of such rules when the observable is $Z$.
Let the (bounded) linear reward be $u:\mathcal{A}\times\Omega\to\mathbb{R}$, where $\Omega$ is the latent outcome space
(e.g., containing $r_{\tau,i}$’s). Let $\pi$ be a prior on $\Omega$ and let $\mathbb{E}_{\pi,\delta}$ denote expectation over
$\omega\sim\pi$, the randomness generating $\mathcal{X}$ (hence $\mu'=\mathcal{T}(\mathcal{X})$), and any internal randomization of $\delta$.

\medskip\noindent
\emph{Step 1 (Simulation/garbling argument).}
Take any $\delta'\in\mathcal{D}(\mu')$, i.e., a rule that observes only $\mu'$.
Define its \emph{lift} $\delta\in\mathcal{D}(\mathcal{X})$ by composition:
\[
\delta(\cdot\,|\,\mathcal{X}) \;:=\; \delta'\big(\cdot\,\big|\,\mathcal{T}(\mathcal{X})\big)
\;=\; \delta'\big(\cdot\,\big|\,\mu'\big).
\]
Thus, given $\mathcal{X}$, the rule $\delta$ first applies the measurable map $\mathcal{T}$ to obtain $\mu'$ and then acts exactly as $\delta'$.
By construction, $\delta$ and $\delta'$ induce the same conditional action law given $\mu'$; hence
\[
\mathbb{E}_{\pi,\delta}\!\big[u(a,\omega)\big]
\;=\;
\mathbb{E}_{\pi,\delta'}\!\big[u(a,\omega)\big].
\]
Therefore, for every rule based on $\mu'$ there exists a rule based on $\mathcal{X}$ achieving \emph{the same} Bayes expected reward.

\medskip\noindent
\emph{Step 2 (Information dominance).}
Taking suprema over admissible rules yields
\[
\sup_{\delta\in\mathcal{D}(\mathcal{X})}\ \mathbb{E}_{\pi,\delta}\!\big[u(a,\omega)\big]
\;\;\ge\;\;
\sup_{\delta'\in\mathcal{D}(\mu')}\ \mathbb{E}_{\pi,\delta'}\!\big[u(a,\omega)\big].
\]
Equivalently, for any convex loss $\ell$ (the negative of a linear reward up to an affine transform),
the minimal Bayes risk based on $\mathcal{X}$ is no larger than that based on $\mu'$.

\medskip\noindent
\emph{Step 3 (Deterministic dominance within the full statistic).}
Because the objective is linear in the action distribution conditional on the observed statistic,
the supremum over the probability simplex is attained at an extreme point; hence there exists an
optimal rule based on $\mathcal{X}$ that is \emph{deterministic} almost surely (selects an argmax of the
posterior expected reward given $\mathcal{X}$). Consequently, any rule that observes only $\mu'$ and
then injects independent randomization (e.g., Gumbel-Max on $\mu'$) is weakly dominated by some deterministic rule that uses $\mathcal{X}$.

Combining the three steps proves the theorem.
\end{proof}

\begin{corollary}[Deterministic exploitation dominates given the fresh batch]
\label{cor:deterministic-dominance}
Let $\mathcal{A}:=\Pi_{\tau}^{*}$ and $\mathcal{X}_{\tau}:=\big(B_{i},\{v_{\tau,j}\}_{j\in\Pi_{\tau}^{*}}\big)$.
For linear utility (equivalently, convex loss), there exists an optimal decision rule
$\delta^{\star}\in\mathcal{D}(\mathcal{X}_{\tau})$ that is \emph{deterministic} almost surely; concretely, for
\[
m_{\tau,i}(\mathcal{X}_{\tau}) \;\equiv\; \mathbb{E}\!\left[r_{\tau,i}\,\middle|\,\mathcal{X}_{\tau}\right],\qquad i\in\Pi_{\tau}^{*},
\]
one may choose $\delta^{\star}(\mathcal{X}_{\tau})\in\arg\max_{i\in\Pi_{\tau}^{*}} m_{\tau,i}(\mathcal{X}_{\tau})$.
Moreover, any policy that observes only $\mu' = T(\mathcal{X}_{\tau})$ and then injects independent randomization
(e.g., Gumbel-Max on $\mu'$) is weakly dominated in Bayes expected utility by some deterministic policy based on $\mathcal{X}_{\tau}$ (in particular, by \textsf{A} when its objective coincides with posterior mean reward on the shared fresh mini-batch).
\end{corollary}

\begin{proof}
Fix $\mathcal{X}_{\tau}$ and write $m(\mathcal{X}_{\tau})=(m_{\tau,i}(\mathcal{X}_{\tau}))_{i\in\Pi_{\tau}^{*}}$.
Any (possibly randomized) rule in $\mathcal{D}(\mathcal{X}_{\tau})$ induces a probability vector
$p(\cdot\,|\,\mathcal{X}_{\tau})\in\Delta(\Pi_{\tau}^{*})$ over actions. The conditional objective is linear:
\[
\mathbb{E}\!\left[ r_{\tau,a}\,\middle|\,\mathcal{X}_{\tau}\right]
\;=\; \sum_{i\in\Pi_{\tau}^{*}} p(i\,|\,\mathcal{X}_{\tau})\, m_{\tau,i}(\mathcal{X}_{\tau}).
\]
A linear functional on the simplex attains its maximum at an extreme point; hence there exists an optimal
deterministic selector (almost surely) $\delta^{\star}(\mathcal{X}_{\tau})\in\arg\max_{i} m_{\tau,i}(\mathcal{X}_{\tau})$.
By measurability of $\arg\max$ on a finite set, $\delta^{\star}$ is a valid decision rule.

For dominance, apply Theorem~\ref{thm:blackwell} with full statistic $\mathcal{X}_{\tau}$ and coarsening
$T(\mathcal{X}_{\tau})=\mu'$. The theorem implies
\[
\sup_{\delta\in\mathcal{D}(\mathcal{X}_{\tau})}\ \mathbb{E}[r_{\tau,\delta(\mathcal{X}_{\tau})}]
\;\ge\;
\sup_{\tilde{\delta}\in\mathcal{D}(\mu')}\ \mathbb{E}[r_{\tau,\tilde{\delta}(\mu')}].
\]
Because the left-hand supremum is attained by a deterministic extreme point $\delta^\star$ (as shown above),
no policy that observes only $\mu'$ and then randomizes (including Gumbel-Max on $\mu'$) can exceed its Bayes expected utility; at best it ties in degenerate cases (exact ties in posterior scores). This proves the claim.
\end{proof}

\begin{corollary}[Design implication for our loop]
\label{cor:design}
With average updates and a shared fresh mini-batch over the $k$ shortlisted prompts, the principled policy is:
(i) use Gumbel-Top-$k$ for \emph{exploration} at shortlist time; then
(ii) \emph{exploit} by selecting $\arg\max_{k \in \Pi_{\tau}^{*}} v_{\tau,k}$ on the shared fresh batch.
Replacing step (ii) by Gumbel-Max on $\mu'$ introduces uninformed randomization and cannot improve and generally degrades the expected reward.
\end{corollary}

\paragraph{Remarks.}
(i) The equal-precision assumption in Thm.~\ref{thm:equal-precision} holds exactly when all shortlisted items are evaluated on the same fresh batch and past precisions are comparable; it is a standard homoscedastic setting as all compared random variables have the same variance (equal noise level due to fixed \llmforward and shared batch) and matches our implementation.

\subsection{Proof for Blockwise  Generation}
\label{Appendix:Proof_Blockwise}

\begin{theorem}[TSGD-M (Blockwise) has no larger variance compared to TSGD]
\label{thm:tsgdm-variance-short}
Fix a blockwise prefix $\mathrm{TokenBlock}_{1:i-1}$. For each candidate $j$ let the stochastic textual gradient satisfy
\[
\mathbb{E}[g_j]=r_j,\qquad \operatorname{Var}(g_j)\le\sigma^2.
\]
Let $\Pi_\tau^{*}$ be the Gumbel-Top-$K$ set at iteration $\tau$. \textbf{TSGD} uses the full-validation argmax $I_\tau$ and applies $g_{I_\tau}$. \textbf{TSGD-M (Blockwise)} samples $A\sim \mathrm{Unif}(\Pi_\tau^{*})$ and averages $m$ such terms
$\widehat g_{\mathrm{TSGD\text{-}M}}=\frac{1}{m}\sum_{t=1}^m g_{A_t}$
with mean pairwise correlation $\rho\in[0,1)$ across the $m$ terms (conditioned on the prefix).
Define the in-band spread
\[
\alpha \;=\; \frac{1}{\sigma^2}\,
\mathbb{E}\!\left[\frac{1}{K}\sum_{j\in \Pi_\tau^{*}}\big(r_j-\bar r\big)^2\right],
\qquad
\bar r=\tfrac{1}{K}\sum_{j\in \Pi_\tau^{*}} r_j.
\]
Then
\[
\operatorname{Var}\!\big(\widehat g_{\mathrm{TSGD\text{-}M}}\big)
\;\le\; \frac{1+(m-1)\rho}{m}\,\sigma^2(1+\alpha)
\;\le\; \operatorname{Var}\!\big(g_{I_\tau}\big)\;\le\;\sigma^2,
\]
whenever $\;\frac{1+(m-1)\rho}{m}\le \frac{1}{1+\alpha}$. In particular, if the Top-$K$ band is tight ($\alpha\approx 0$) and $m>1$ with $\rho<1$, TSGD-M (Blockwise) has strictly smaller variance than TSGD.
\end{theorem}

\begin{proof}
(1) Law of total variance with $A\!\sim\!\mathrm{Unif}(\Pi_\tau^{*})$ gives
$\operatorname{Var}(g_A)\le \sigma^2+\operatorname{Var}(r_A)=\sigma^2(1+\alpha)$.
(2) Averaging $m$ terms with mean correlation $\rho$ yields
$\operatorname{Var}(\widehat g_{\mathrm{TSGD\text{-}M}})\le \frac{1+(m-1)\rho}{m}\,\sigma^2(1+\alpha)$.
(3) TextGrad fixes $I_\tau$ (no selection variance), so
$\operatorname{Var}(g_{I_\tau})\le \sigma^2$.
Combine (2) and (3) to obtain the claim under
$\frac{1+(m-1)\rho}{m}\le \frac{1}{1+\alpha}$; tight $\alpha$ and $m>1$, $\rho<1$ imply strict inequality.
\end{proof}

\section{Computational Complexity of Generation Strategies}
\label{appendix:complexity_analysis}

\textbf{Computational Complexity} To quantify the efficiency gains of Blockwise generation, we analyze the total computational cost $C$ in terms of floating-point operations (FLOPs). Let $P$ denote the length of the system prompt (prefix), $T_{\max}$ the total number of generated tokens, and $b$ the block size. We assume the cost of a forward pass for a sequence of length $n$ is dominated by the attention mechanism, scaling as $O(n^2)$.

\begin{enumerate}
    \item \textbf{Token-wise Generation (Naive Momentum):} If the prompt is switched or the distribution is re-calculated at every token $j \in \{1, \dots, T_{\max}\}$, the KV-cache must be invalidated and recomputed for the prefix. The total cost is:
    \begin{equation}
        C_{token} \approx \sum_{j=1}^{T_{\max}} (P+j)^2 = O(T_{\max} \cdot P^2 + T_{\max}^3)
    \end{equation}
    This quadratic dependency on $P$ at every step makes token-wise momentum computationally prohibitive for long-context LLMs.

    \item \textbf{Promptwise Generation (Proposed: Static):} In this special case ($b=T_{\max}$), the momentum sampling occurs once. The prefix is processed once (prefill), and subsequent tokens are generated using the existing KV-cache (decoding):
    \begin{equation}
        C_{prompt} \approx \underbrace{P^2}_{\text{Prefill}} + \underbrace{\sum_{j=1}^{T_{\max}} (P+j)}_{\text{Decoding}} = O(P^2 + T_{\max} \cdot P + T_{\max}^2)
    \end{equation}
    While highly efficient, this method lacks the ability to correct the generation trajectory using momentum signals post-initialization.

    \item \textbf{Blockwise Generation (Proposed: Adaptive):} By setting a block size $1 < b < T_{\max}$, we re-calculate momentum every $b$ tokens. This amortizes the prefill cost:
    \begin{equation}
        C_{block} \approx \underbrace{\frac{T_{\max}}{b} \cdot P^2}_{\text{Amortized Prefill}} + \underbrace{T_{\max} \cdot P}_{\text{Decoding}}
    \end{equation}
\end{enumerate}

\textbf{Conclusion:} As $b$ increases, $C_{block}$ approaches $C_{prompt}$, but even for small $b$ (e.g., $b=50$ as we actually implements), the reduction in $P^2$ operations relative to token-wise generation is an order of magnitude, enabling real-time inference without sacrificing the benefits of iterative momentum.
\begin{table*}
\centering
\caption{Data splitting of benchmark datasets.}
\label{tab:Task_Description}
\small
\begin{tabular}{l|r|r|r}
\toprule
\textbf{Task and Source} & \textbf{\textbar original train\textbar} & \textbf{\textbar validation\textbar} & \textbf{\textbar test\textbar} \\
\midrule
TREC \citep{lu2022fantastically} & 400 & 250 & 250 \\
GSM8K \citep{cobbe2021training} & 200 & 300 & 1319 \\
MATH (Algebra) \citep{hendrycks2measuring} & 350 & 350 & 487 \\
IFBench \citep{pyatkingeneralizing} & 150 & 300 & 294
 \\
Arc Challenge \citep{clark2018think} & 1120 & 299 & 1170 \\
HotpotQA (full wiki) \citep{yang2018hotpotqa} & 90.4k & 7.41k & 7.41k \\
\bottomrule
\end{tabular}
\end{table*}

\section{Supplementary Experiment Details}
\label{app:exp_details}
\subsection{Extended Scaling Experiments}
\label{appendix:extended_scaling_experiment}
\begin{figure}[H]
    \centering
    \includegraphics[width=0.47\textwidth]{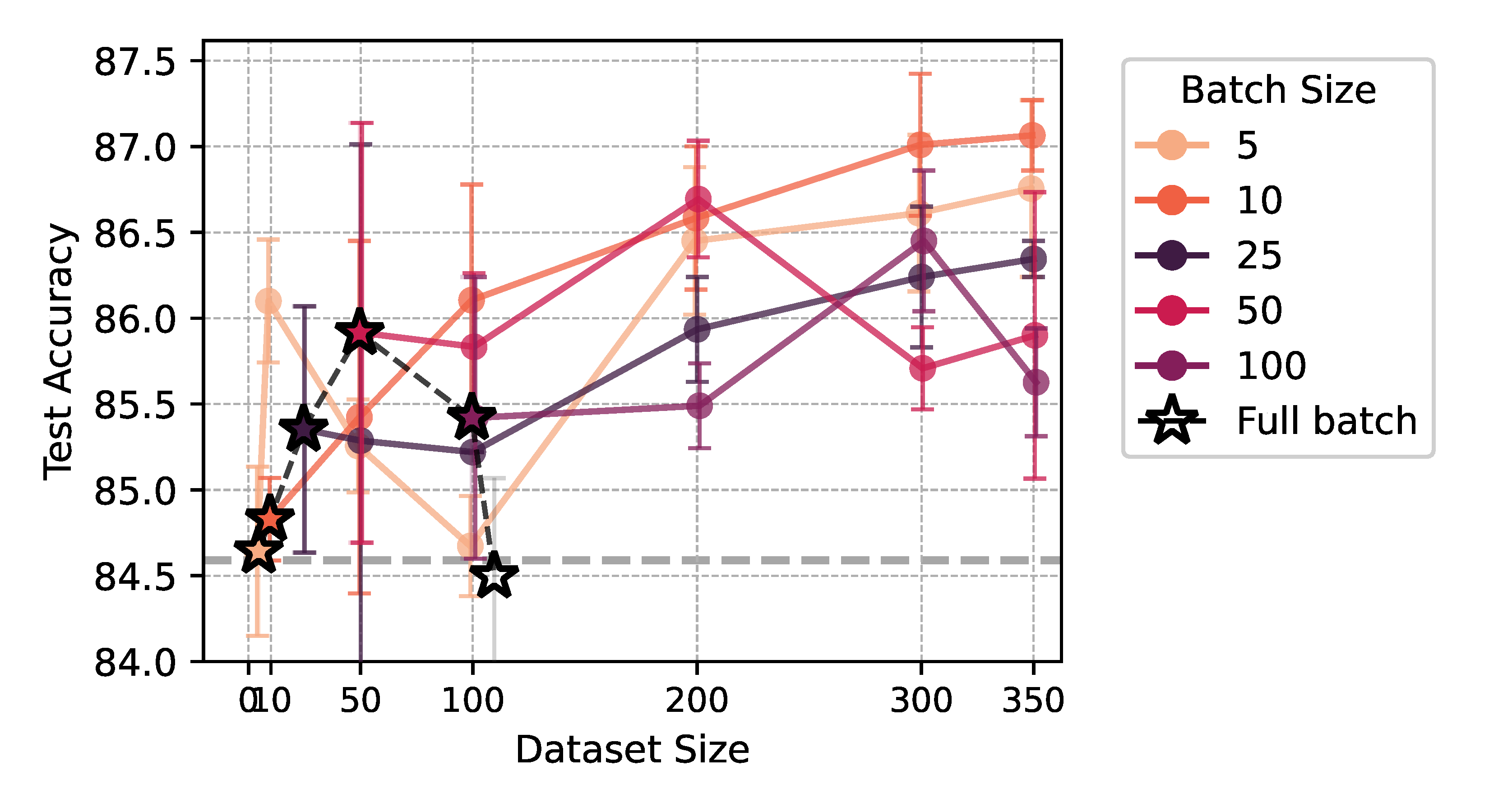}
    \caption{Scaling of TGD/TSGD on MATH (Algebra) with standard error (error bars) with $\llmforward$ as GPT-4o-mini and $\llmbackward$ as GPT-4o. Full-batch TGD achieves its best performance around dataset size 50 but shows the largest uncertainty. Across dataset sizes, minibatch TSGD attains higher mean accuracy than full-batch training; however, smaller batches tend to exhibit larger run-to-run variability than larger batches as the dataset size increases. This instability motivates our proposed method, which achieves higher accuracy with lower standard error on most tasks.}
    \label{fig:scaling_standard_error}
\end{figure}

Because TGD/TSGD exhibit substantial run-to-run variability, we report standard error in the left subplot of Fig.~\ref{fig:scaling} (reproduced in Fig.~\ref{fig:scaling_standard_error}). In Fig.~\ref{fig:scaling_standard_error}, full-batch TGD achieves its highest mean test accuracy at dataset size $N=50$, but it also exhibits the largest standard error among all settings, indicating poor stability. As dataset size increases, mini-batch TSGD generally improves mean accuracy relative to full-batch training; however, smaller batches tend to show larger uncertainty than larger batches, especially in the low-data regime (when full dataset size is less than 200). Overall, both TGD and TSGD exhibit an unfavorable accuracy–stability trade-off and do not scale reliably across data sizes. These observations motivate a prompt optimizer prioritizing stability across iterations. We introduce our momentum-based sampling method, which reduces standard error across seeds while improving average test accuracy as the training dataset scales (see Fig.~\ref{fig:scale}).

\begin{figure}[H]
    \centering
    \includegraphics[width=0.47\textwidth]{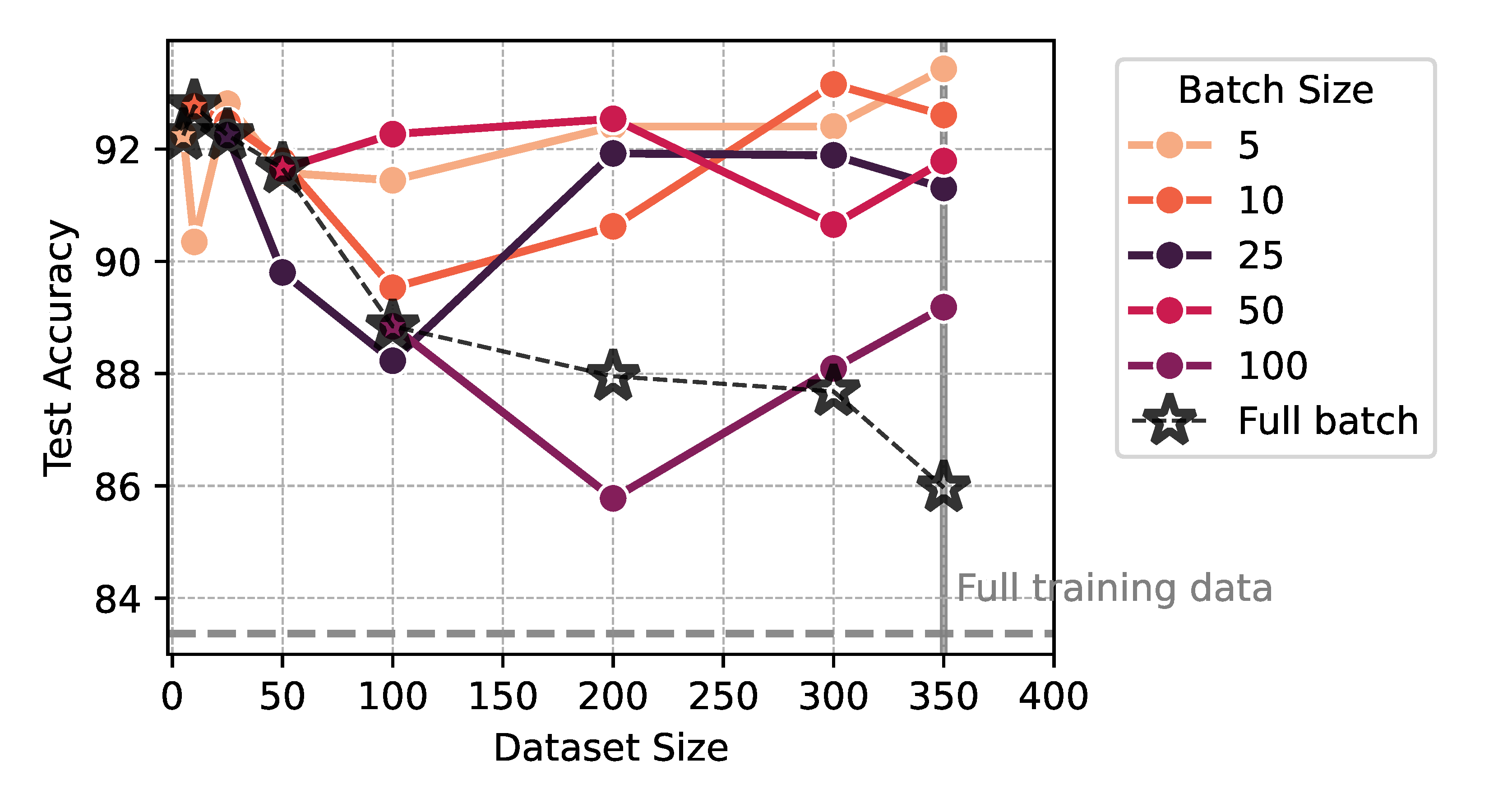}
    \caption{Scaling of TGD/TSGD on MATH (Algebra) with standard error (error bars) with $\llmforward$ as Gemini-2.5-Flash-lite and $\llmbackward$ as Gemini-2.5-Flash. While the Gemini models shall support the full training set (350 samples) within their 1M+ token explicit context window, Full batch TGD (dashed line) encounters an implicit context wall at approximately 10 samples. Beyond this threshold, increasing the context length via larger batches leads to a consistent degradation in performance, contrasting with smaller batch sizes which successfully scale to larger datasets.
    }
    \label{fig:scaling_no_standard_error_gemini_two}
\end{figure}

\begin{figure}[H]
    \centering
    \includegraphics[width=0.45\textwidth]{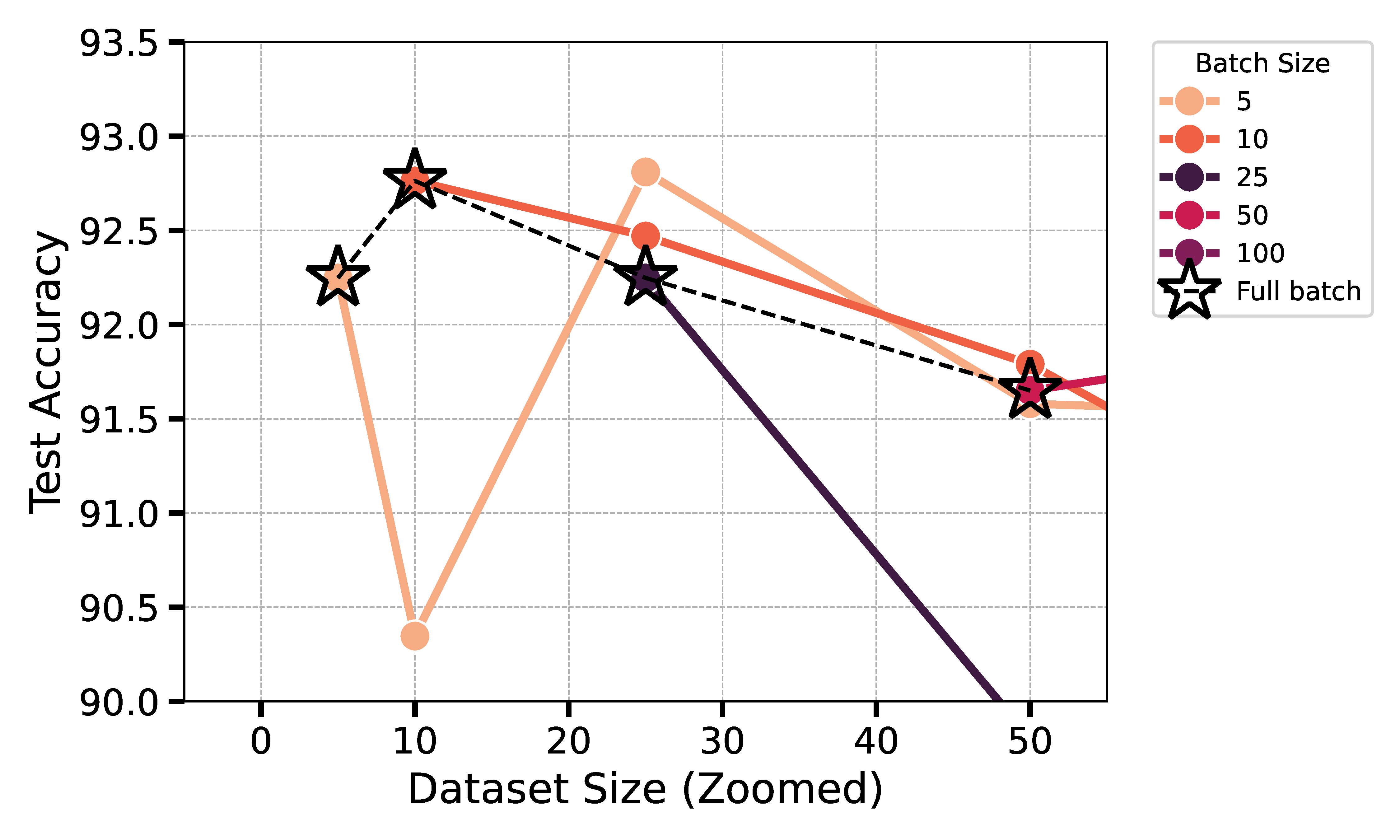}
\caption{Figure~\ref{fig:scaling_no_standard_error_gemini_two} (top-left) shows a zoomed-in view of the low-data regime ($N \leq 50$).
    }
\label{fig:scaling_no_standard_error_gemini_two_zoom}
\end{figure}

To explore the scaling properties of our approach, we leverage the Gemini family LLMs \citep{comanici2025gemini}, which are capable of processing inputs with context length up to 1M tokens. In Figure~\ref{fig:scaling_no_standard_error_gemini_two} and Figure~\ref{fig:scaling_no_standard_error_gemini_two_zoom}, we extend our scaling analysis by employing Gemini-2.5-Flash-lite as the inference model ($\llmforward$) and Gemini-2.5-Flash as the backward model for generating textual gradients ($\llmbackward$). While both LLMs possess a 1M+ token explicit context window, our results reveal a significant performance degradation beyond a batch size of approximately 10 samples (See Figure~\ref{fig:scaling_no_standard_error_gemini_two_zoom}). This 'implicit context wall' persists despite the absence of the 128k+ token explicit tokens constraints found in GPT-based models. These findings corroborate recent observations that expanded context capacity does not consistently translate into effective task solving over long sequences ~\citep{du-etal-2025-context}. 

Furthermore, our results suggest that the optimal minibatch size $m$ is highly LLM-dependent and may vary significantly across different LLM architectures. Even when evaluated on the same MATH (Algebra) dataset, varying the underlying LLM shifts the ideal configuration; in our specific experiments with the Gemini-2.5 family, a minibatch size of $m = 5$ enabled the most effective scaling for a total training set of $N = 350$. This highlights that there is no 'one-size-fits-all' minibatch design. This motivates our formulation of APE as a global optimization problem over the entire history of generated prompts, rather than a static process dependent on an \emph{a priori} guess of the batch size.

Due to the black-box nature of the closed-source LLM APIs, metrics such as internal throughput and token-level latency are inaccessible. Consequently, we report the total wall-clock time across our scaling experiments as a measure of practical computational overhead. For these evaluations, we report GPT-4o-mini as the inference LLM ($\llmforward$) and GPT-4o as the backward model that generating textual gradients ($\llmbackward$). The comprehensive runtime results are detailed in Table~\ref{tab:runtime_scaling_gpt}. We note that these durations are sensitive to stochastic API latency and rate-limit errors; high-frequency querying during large-batch iterations shall lead to temporary throttling, thereby affecting the consistency of the reported temporal metrics.

\begin{table*}[ht]
\centering
\caption{Average wallclock time per epoch (seconds) across varying dataset sizes ($N$) and minibatch sizes ($m$) over 5 independent runs for scaling experiments in MATH (algebra). Note \emph{one epoch} is defined as each example would be seen once during TextGrad optimization loop. Different seeds suggest different orders of in-context examples fed into TextGrad optimization loop. $m < N$ denotes minibatch TSGD and $m = N$ denotes Full Batch TGD.}
\label{tab:runtime_scaling_gpt}
\small
\begin{tabularx}{\textwidth}{l *{6}{>{\centering\arraybackslash}X}}
\toprule
\textbf{Dataset ($N$)} & \textbf{$m=5$} & \textbf{$m=10$} & \textbf{$m=25$} & \textbf{$m=50$} & \textbf{$m=100$} & \textbf{$m=N$(Full Batch)} \\
\midrule
5   & 108  & --     & --     & --     & --     & 108.6  \\
10  & 257  & 184  & --     & --     & --     & 184.2  \\
25  & 788  & 756  & 416  & --     & --     & 416  \\
50  & 1058 & 1412 & 1092 & 826  & --     & 826  \\
100 & 2126 & 1691 & 1730 & 1631 & 1624 & 1624 \\
200 & 4790 & 4161 & 4093 & 3529 & 3168 & 2809 \\
300 & 6705 & 6081 & 4942 & 5938 & 4163 & 4508 \\
350 & 7232 & 5981 & 6332 & 5910 & 5759 & 4508 \\
\bottomrule
\end{tabularx}
\end{table*}

\subsection{Experiment Setup}

In \cref{tab:Task_Description}, we list the split of data. Below, we elaborate on how the data and evaluation are set up.

\textbf{TREC}
We evaluate a system’s prediction by comparing its output string with the ground-truth label provided in the dataset, assigning a score of 1 for an exact match and 0 otherwise.
Before comparison, both the model output and ground-truth strings are normalized to mitigate differences due to tokenization and capitalization.

\textbf{GSM8K}
We use the same evaluation metric as TextGrad, a string-based exact match metric to quantify accuracy.

\textbf{MATH(algebra)} We use the built-in dataset MATH with subset algebra from DSPy Tutorials for MATH Reasoning \cite{dspy_math_tutorial}. We follow the same setup with 350 and 350 question-answer pairs sampled from the official test set for development/validation set and test set. Same as GSM8K, we evaluate the accuracy of the final numerical value that appears in the \llm  output.

\textbf{IFBench} We adopt the same setup as GEPA~\citep{agrawal2025gepa}.
For our experimental setup, we follow the GEPA protocol: we split IF-RLVR Train into our own train/validation sets and use IFBench as the held-out test set, ensuring the optimizers never see the new constraints evaluated in IFBench. We implement a two-stage pipeline that (1) produces an initial response to the user query and (2) rewrites that response to satisfy the specified constraints. A textual feedback module then reports which constraints are satisfied and which are violated in the system output. Overall, our splits include 150 training examples, 300 validation examples, and 294 test examples.

\textbf{ARC-Challenge}
We use the challenge subset. We randomly sample 500 examples for test set from the original 1,170-question test set and use the original validation set containing 299 examples.
We evaluate \llm performance using a string-based exact match metric.

\textbf{HotPotQA}~\cite{yang2018hotpotqa} is a large-scale Wikipedia QA benchmark. We adopt the \emph{full-wiki} setting.
Following \citet{dspy_built_in_datasets_2025}, we use 300/300 for validation/test split.
We evaluate HotpotQA under two configurations. For \textgrad, which targets reasoning tasks and does not implement multi-hop retrieval (RAG is listed as future work), we provide the questions \emph{with their context} directly to \llmforward, following the same setup as reasoning experiments in which the input to the inference model is the question LLM needs to solve, ours have one more context component. For DSPy and \adalflow, we use a retrieval-augmented (RAG) setup: BM25 retrieves candidate passages from Wikipedia, and a chain-of-thought program performs \textit{vanilla RAG} for two components with retriever and generator. We uses exact match \citep{yang2018hotpotqa}. 

\textbf{GEPA.} For baseline GEPA~\citep{agrawal2025gepa}, we use the same training, validation and testing splits as other three APE methods. 
Across all tasks and all LLMs setup, the GEPA optimizer is configured with the same strategy with "medium" budget to balance performance and efficiency. The optimizer utilizes a dual-minibatch approach: an actual minibatch size of 5 is used for the mutation and reflection steps, while a larger estimator minibatch size of 35 is allocated for internal scoring and candidate selection processes.

\paragraph{Number of rollouts}
To account for varying task complexity and dataset scales, we specifically tune the number of rollouts, defined as the maximum number of metric evaluations (\texttt{max\_metric\_calls}) permitted during the optimization phase. The task-specific limits are detailed in Table~\ref{tab:gepa_params}.

\begin{table}[ht]
\centering
\caption{GEPA Rollout Number of Rollouts across all tasks.}
\label{tab:gepa_params}
\small
\begin{tabular}{lc}
\toprule
\textbf{Task} & \textbf{Rollouts} \\
\midrule
MATH & 2440 \\
HotpotQA & 2300 \\
TREC & 1690 \\
GSM8K & 1295 \\
ARC-Challenge & 1190 \\
IFBench & 1190 \\
\bottomrule
\end{tabular}
\end{table}

\subsection{Extended Benchmarks}
\label{appendix:extended_benchmarks}
\begin{figure*}
\begin{framed}
\noindent\textbf{Sample ARC-Challenge Prompt (Token-Level Synchronization Issue)}\\[1ex]

\small
You will answer a multiple-choice science question. Your response must be *only* the single, plain character label of the correct option (e.g., 'A', 'B', 'C', 'D').Do not provide any explanation, reasoning, conversational text, or any other formatting (like LaTeX, bolding, or boxes). The *entire* response must be 'Answer: X' where X is that label, and nothing else. This is criticalfor downstream processing.
\end{framed}
\caption{An illustration of the token-level synchronization issue in blockwise generation. Note the concatenation artifact ``criticalfor'' occurring at the boundary of a 50-token block.}
\label{fig:sample_arc_blockwise_generation}
\end{figure*}

We also report benchmark results using Gemini-2.5-Flash-Lite as the inference model $\llmforward$ \citep{comanici2025gemini}, and Gemini-2.5-Flash as the backward (textual-gradient) model, in Table~\ref{tab:benchmark_gemini_two}. 

\textbf{TSGD-M generalizes across LLMs.} In Table~\ref{tab:benchmark_gemini_two}, the substantial performance gains, observed in TSGD-M, particularly the +4.04\% improvement in TextGrad-M (Blockwise), underscore the critical roles of stability and controlled exploration in prompt optimization. While standard TextGrad is typically applied to free-form generation or retrieval tasks where lexical fluidity is permissible, DSPy frames optimization as a declarative assembly of optimizable modules~\citep{khattab2024dspy}. This modular approach is inherently sensitive to minor lexical variations, as subtle changes in a symbolic instruction can significantly alter the model's reasoning graph. Consequently, COPRO, which relies more on structured input, exhibits a more modest overall improvement of +1.66\%, highlighting the necessity of momentum-based gradients in navigating the high-variance landscape of structured prompt programming. For AdalFlow, the aggregate value already exhibits the highest baseline performance without the addition of momentum; therefore, the observed improvement is the most modest, particularly as our momentum method remains training-free and we do not finetune any model. 

\textbf{Blockwise generation is susceptible to boundary synchronization errors, where independent sampling at block transitions may result in lexical incoherence or malformed tokens at the join points.} A notable artifact of our blockwise sampling is the occasional production of non-coherent lexical joins. For instance, in Figure~\ref{fig:sample_arc_blockwise_generation}, we observe prompts where high-frequency tokens were concatenated without appropriate spacing. This occurs when the transition between block $i$ and block $i+1$ falls within a phrase, and the $\llmforward$,
typically lacking the immediate transition probabilities of a single-pass decode,
generates a completion that is logically sound but typographically fused. This would potentially explain why Blockwise generation exhibits higher variance in output quality and structure compared to Promptwise generation.

\begin{table*}[ht]
\centering
\caption{Test accuracy (\%) with standard deviation inside parentheses for $\llmforward$ as Gemini 2.5 flash-lite and $\llmbackward$ as Gemini 2.5 flash.}
\label{tab:benchmark_gemini_two}
\small
\setlength{\tabcolsep}{4pt}
\begin{tabular}{lcccccccc}
\toprule
\textbf{Method} & \textbf{TREC} & \textbf{ARC-Challenge} & \textbf{GSM8K} & \textbf{MATH} & \textbf{HotPotQA} &\textbf{IFBench} & \textbf{Aggregate} & \textbf{Improvement} \\
\midrule
\textgrad\, w/o val revert & 84.86(1.32) & 91.56(0.68) & 91.23(0.89) & 87.58(2.12) & 51.89(0.45) & 37.56(0.04) & 74.11 & -- \\
\textgrad\, w/ val revert & 80.12(0.00) & 92.2(0.00) & 94.19(0.00) & 83.36(0.32) & 51.72(0.23) & 38.12(0.17) & 73.29 & -- \\
\textbf{\textgrad-M (Promptwise)} & 86.82(0.76) & 93.74(0.45) & 94.41(0.21) & \textbf{91.58(1.58)} & 55.23(0.19) & 40.82(0.28) & 77.10 & \textbf{+3.81} \\
\textbf{\textgrad-M (Blockwise)} & \textbf{87.98(0.91)} & \textbf{93.91(0.55)} & \textbf{94.58(0.22)} & 91.02(1.72) & \textbf{55.57(0.27)} & \textbf{40.91(0.39)} & \textbf{77.33} & \textbf{+4.04} \\
\midrule
COPRO & 81.45(1.25) & 93.53(0.19) & 92.01(0.92) & 90.76(0.56) & 46.10(0.89) & 36.89(0.56) & 73.46 & -- \\
\textbf{COPRO-M (Promptwise)} & \textbf{82.57(0.98)} & 93.80(0.44) & \textbf{95.10(0.89)} & \textbf{93.22(0.45)} & 48.10(0.79) & 37.92(0.45) & 75.12 & \textbf{+1.66} \\
\textbf{COPRO-M (Blockwise)} & 82.21(1.21) & \textbf{94.02(0.56)} & 94.89(0.78) & 93.01(0.59) & \textbf{48.25(0.92)} & \textbf{38.12(0.56)} & 75.08 & \textbf{+1.62} \\
\midrule
AdalFlow & \textbf{86.67(0.02)} & 91.1(0.01) & 88.67(0.012) & 89.72(0.005) & 51.93(0.008) & 38.0(0.01) & 74.35 & -- \\
\textbf{AdalFlow-M (Promptwise)} & 86.60(0.018) & \textbf{92.53(0.005)} & \textbf{89.37(0.016)} & \textbf{90.73(0.006)} & \textbf{53.44(0.018)} & \textbf{39.33(0.012)} & 75.33 & \textbf{+0.98} \\
\midrule
GEPA & 87.00(0.024) & 92.33(0.006) & 90.0(0.017) & 88.00(0.005) & 53.00(0.005) & 35.67(0.012) & 74.33 & -- \\
\bottomrule
\end{tabular}
\end{table*}

\subsection{Sensitivity Analysis on Minibatch Validation Size}
\label{appendix:sensitivity_analysis}
\begin{figure}[H]
\qquad\includegraphics[
width=0.35\textwidth,trim=0.5 3 0 0,clip]
{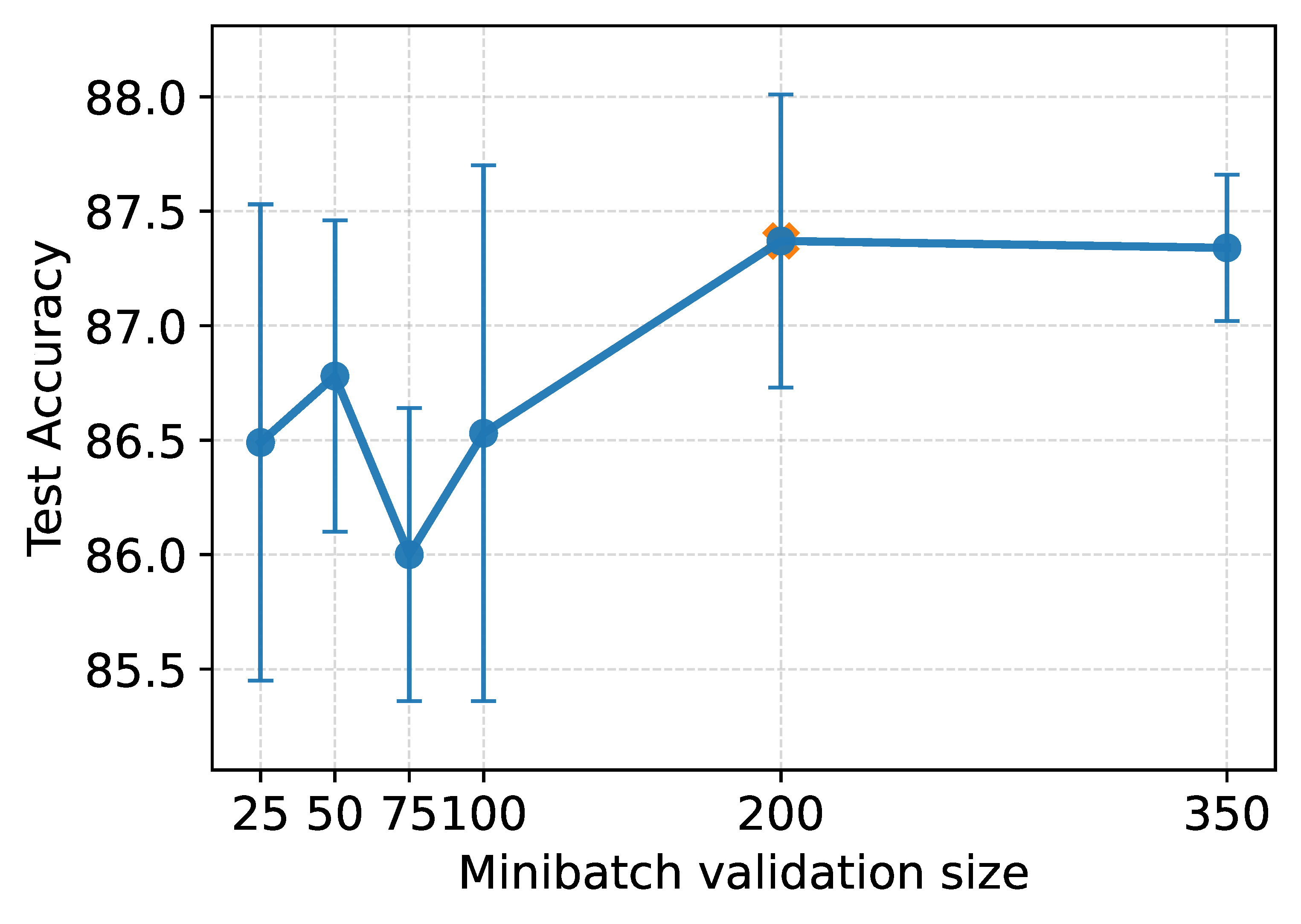}
\caption{Sensitivity analysis on minibatch size for \textgrad-M for MATH.}
\label{tab:sensitivity_analysis}
\end{figure}

In this section, we show that \textbf{TSGD-M is not sensitive to minibatch validation size with $\llmforward$ as GPT-4o-mini and $\llmbackward$ as GPT-4o.} \cref{tab:sensitivity_analysis} varies the minibatch validation size $|v|$ and shows that final test accuracy is \emph{not} highly sensitive to this choice: the best result occurs at $|v|{=}200$, with the second-best at $|v|{=}50$.
Because the evaluation cost scales approximately linearly in $|v|$ (i.e., $O(|v|)$ per minibatch validation evaluation), we favor smaller validation sets that preserve accuracy while reducing overhead; in the main experiments we adopt $|v|{=}50$ as a strong accuracy-cost trade-off.

\subsection{Validation vs. Test accuracy}
\label{section:validation_vs._test accuracy}
\begin{figure}[htbp]
    \centering
    \includegraphics[width=0.47\textwidth]{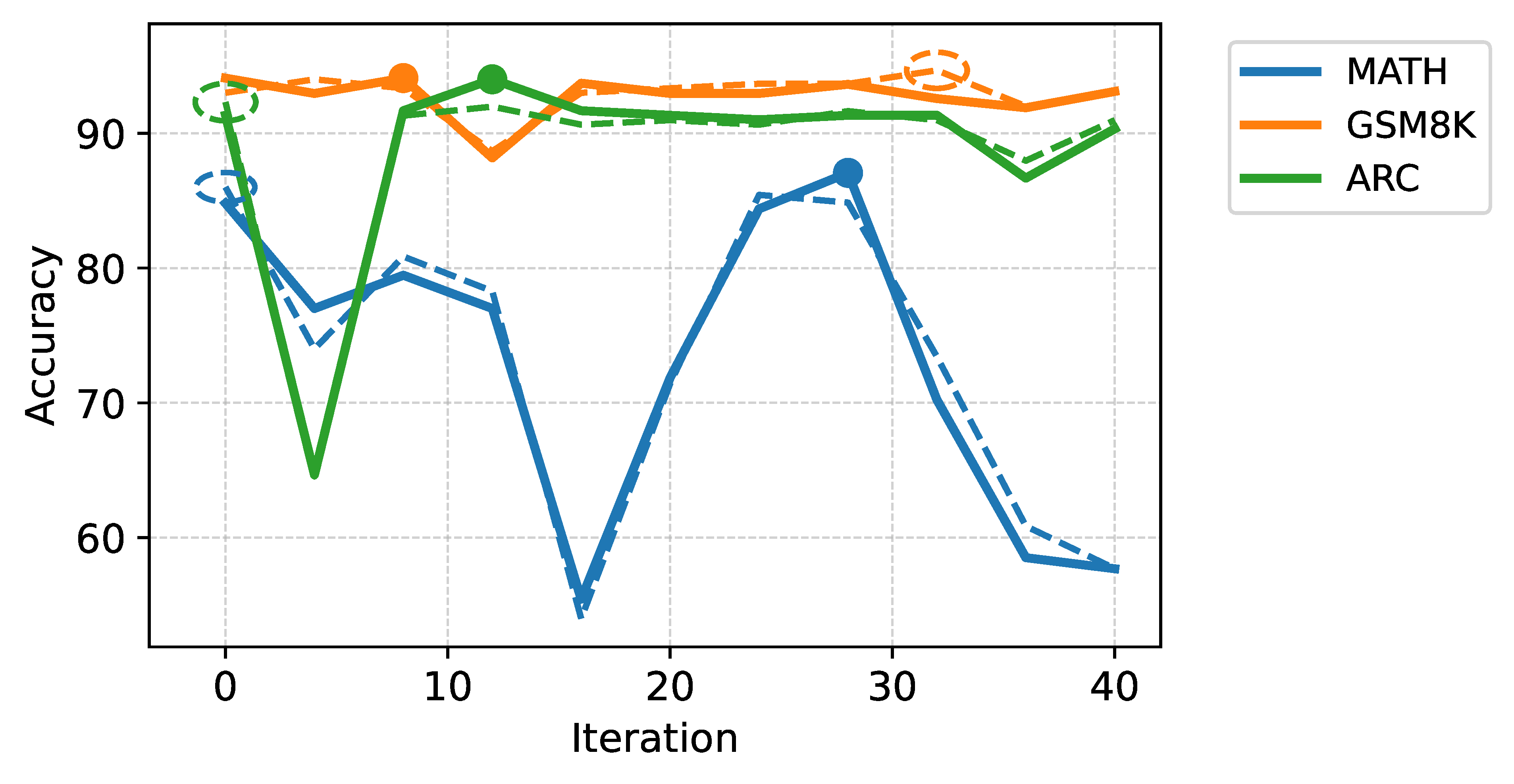}
    \caption{\textbf{Test (solid lines)} and \textbf{Validation (dashed lines)} accuracy over one run on MATH,GSM8K, and ARC for vanilla \textgrad\ w/o validation revert.
We fix the data size to 100, batch size to 5, and seed to 1, and vary only the dataset.
The highest test accuracy is marked with a solid dot, and the highest validation accuracy with a dashed circle.
Across all three tasks, the prompt achieving the highest validation accuracy is not the one that generalizes best on the test set, 
suggesting that pure exploitation over the argmax, which vanilla \textgrad uses, shall not lead to the global optimal.
}
\label{fig:val_test_3dataset}
\end{figure}

In ~\cref{fig:val_test_3dataset}, we illustrate that under vanilla \textgrad\, the prompt with the highest validation accuracy is not necessarily the prompt that generalizes best in the test set. Therefore, we should keep a window with size $K$ of top performing prompts in the validation set and encourage exploration within $K$ top performing prompts instead of \textit{the best} performing prompt on the validation set.

\subsection{Mean of Minibatch Validation vs. Full Validation}
\label{sec:minibatch_vs_full}
\begin{figure}[htbp]
    \includegraphics[width=0.5\textwidth, trim=8pt 2pt 4pt 4pt, clip]{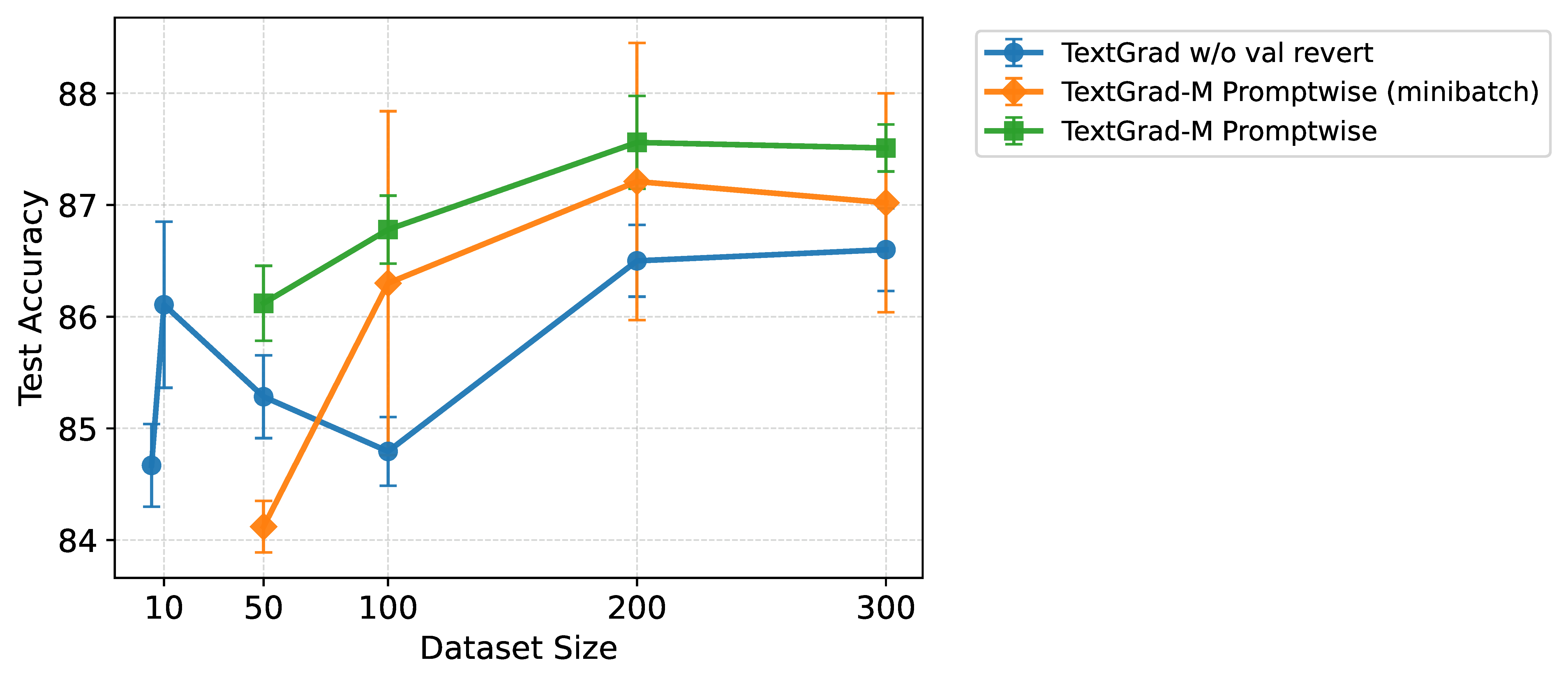}
    \caption{Minibatch running mean vs.\ full validation for final–prompt selection. Error bars show standard error. 
    For small $|\mathcal{D}_{\text{train}}|$ (few iterations), the minibatch running mean is high-variance and can underperform full validation. 
    With larger $|\mathcal{D}_{\text{train}}|$, more iterations reduce variance and both criteria select prompts with comparable test accuracy.}
    \label{fig:running_mean_vs_highest_validation}
\end{figure}

In this section, we compare two final–prompt selection rules for \textgrad-M (Promptwise): 
(i) picking the prompt with the highest \emph{running mean} of minibatch validation accuracy, and 
(ii) picking the prompt with the highest accuracy on the \emph{full} validation set. 
Throughout, we fix the training batch size $m=5$ and epochs $E=2$, and vary the training set size $|\mathcal{D}_{\text{train}}|$. 
The number of optimization iterations is 
$T=\left\lceil \frac{E\,|\mathcal{D}_{\text{train}}|}{b} \right\rceil$.
In~\cref{fig:running_mean_vs_highest_validation}, we observe that for small $|\mathcal{D}_{\text{train}}|$ (few iterations), the running-mean estimator is noisy and can select suboptimal prompts due to large bias between empirical estimates and true validation accuracy . 
As $|\mathcal{D}_{\text{train}}|$ grows (with more iterations and resamples), the running-mean estimate concentrates and matches the full-validation selection, yielding comparable test performance.

\subsection{Momentum Window}
\label{Appendix:Momentum_window}
\begin{figure}[h]
\centering
\includegraphics[width=0.85\linewidth]
{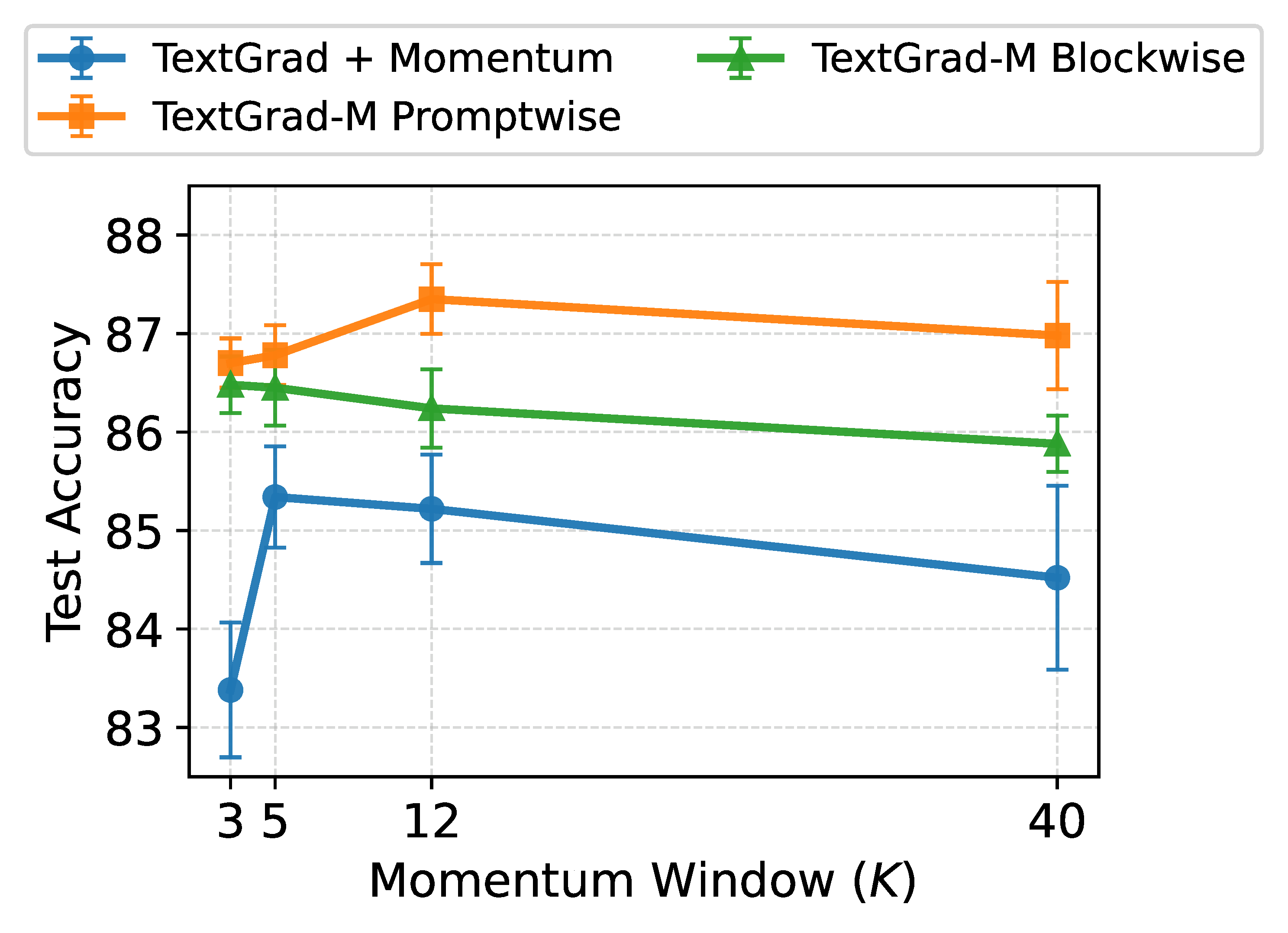}
\vspace{-0.05in}
\caption{Test Performance of vanilla \textgrad\text{-}Momentum and \textgrad-M on MATH with same window size. Error bars are the standard error.}
\label{tab:window_selection}
\end{figure}

\begin{figure}[H]
    \includegraphics[width=0.5\textwidth, trim=9pt 2pt 4pt 4pt, clip]{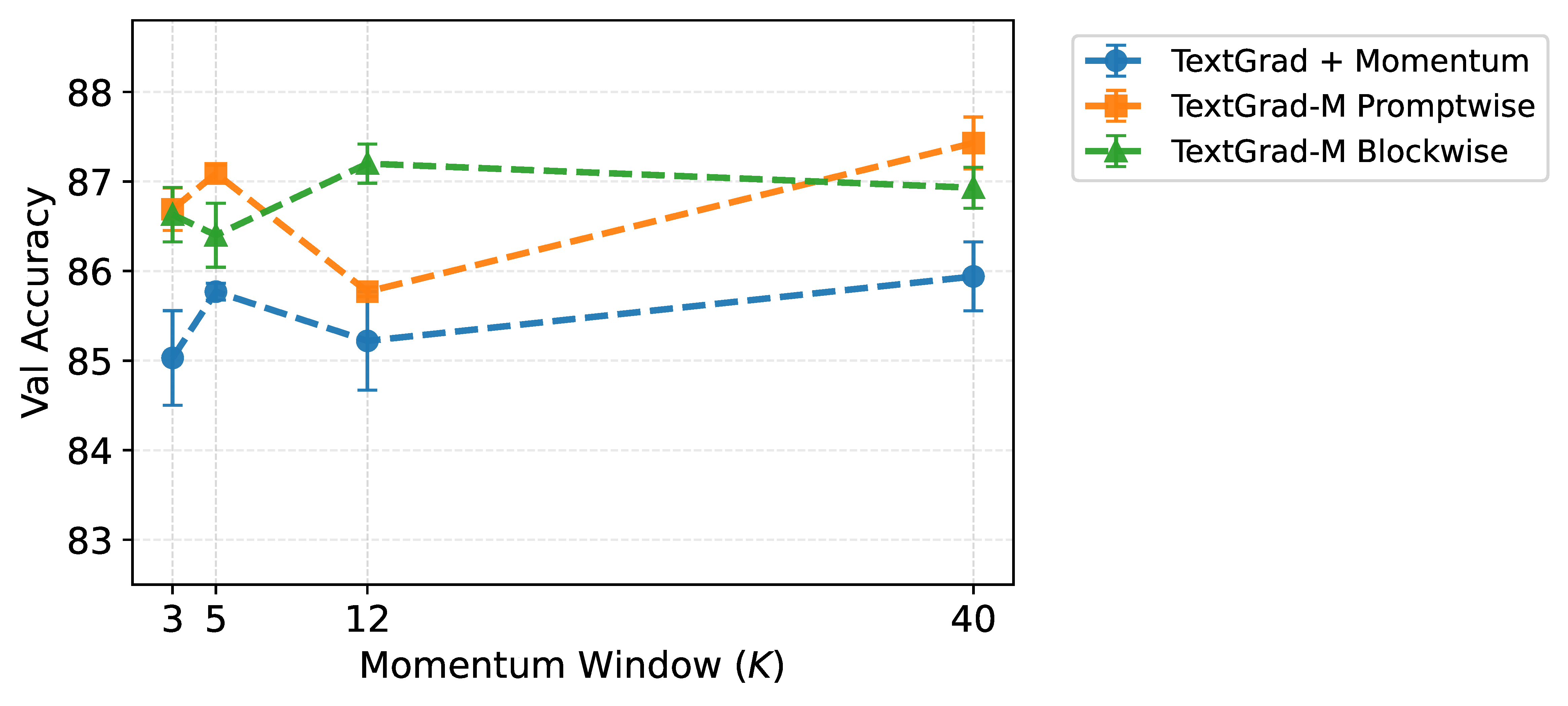}
    \caption{Validation Performance of vanilla \textgrad\text{-}Momentum and \alg\ on MATH(algebra). Error bars denote standard error.
    The momentum window corresponds to the number of concatenated past prompts in \textgrad-Momentum and to the Gumbel-Top-$k$ window size $K$ in \alg.
    Promptwise generation improves consistently as $K$ increases, while blockwise generation is less stable for small $K$. Both \alg~variants outperform the baseline across window sizes.}
    \label{tab:window_selection_validation}
\end{figure}

We provide the underlying statistics of two interpretations of momentum with window size comparison for both \textgrad \, and \alg\ in Table~\ref{tab:window_selection_table}. The test performance is Fig~\ref{tab:window_selection} and the validation performance is Fig~\ref{tab:window_selection_validation}. Both \cref{tab:window_selection} and \cref{tab:window_selection_validation} showcase our method shall break the context window by reducing input tokens via momentum sampling with improved performance (both test and validation) compared to TextGrad-Momentum.

\begin{table*}[t!]
\centering
\caption{Performance of \textgrad\text{-}M and \textgrad\ variants with/without validation revert on MATH(algebra). TG stands for \textgrad. Momentum Window in \textgrad\ is the number of past prompts concatenated; in \textgrad\text{-}M Window stands for $K$ in Gumbel-Top-$k$. 
}
\label{tab:window_selection_table}
\setlength{\tabcolsep}{3pt}
\renewcommand{\arraystretch}{0.9}
\resizebox{.5\linewidth}{!}{
\begin{tabular}{lcc}
\toprule
\textbf{Method} & \textbf{Val Accuracy (\%)} & \textbf{Test Accuracy (\%)} \\
\midrule
\tg\ w/ validation revert                & $84.00 \pm 0.00$ & $85.42 \pm 0.00$ \\
\tg\ w/o validation revert               & $86.23 \pm 0.42$  & $84.67 \pm 0.58$ \\
\addlinespace[2pt]
\tg\ + Momentum Window=3                 & $85.03 \pm 1.18$  & $83.38 \pm 1.53$ \\
\tg\text{-}M + Window=3 + Promptwise     & \underline{$86.69 \pm 0.53$} & \bm{$86.70 \pm 0.56$} \\
\tg\text{-}M + Window=3 + Blockwise      & \bm{$86.63 \pm 0.68$} & \underline{$86.48 \pm 0.64$} \\
\addlinespace[2pt]
\tg\ + Momentum Window=5                 & $85.77 \pm 0.21$  & $85.34 \pm 1.15$ \\
\tg\text{-}M + Window=5 + Promptwise     & \bm{$87.09 \pm 0.21$} & \bm{$86.78 \pm 0.68$} \\
\tg\text{-}M + Window=5 + Blockwise      & \underline{$86.40 \pm 0.80$} & \underline{$86.45 \pm 0.86$} \\
\addlinespace[2pt]
\tg\ + Momentum Window=9                 & $85.43 \pm 0.62$  & $84.94 \pm 1.59$ \\
\tg\text{-}M + Window=9 + Promptwise     & \bm{$86.10 \pm 0.94$} & \bm{$85.94 \pm 1.19$} \\
\tg\text{-}M + Window=9 + Blockwise      & \underline{$83.80 \pm 1.00$} & \underline{$85.15 \pm 1.01$} \\
\addlinespace[2pt]
\tg\ + Momentum Window=12                & $85.22 \pm 1.23$  & $85.22 \pm 1.23$ \\
\tg\text{-}M + Window=12 + Promptwise    & \underline{$85.77 \pm 0.12$} & \bm{$87.35 \pm 0.79$} \\
\tg\text{-}M + Window=12 + Blockwise     & \bm{$87.20 \pm 0.49$} & \underline{$86.24 \pm 0.89$} \\
\addlinespace[2pt]
\tg\ + Momentum Window=40                & $85.94 \pm 0.86$  & $84.52 \pm 2.09$ \\
\tg\text{-}M + Window=40 + Promptwise    & \bm{$87.43 \pm 0.65$} & \bm{$86.98 \pm 1.22$} \\
\tg\text{-}M + Window=40 + Blockwise     & \underline{$86.93 \pm 0.51$} & \underline{$85.88 \pm 0.64$} \\
\bottomrule
\end{tabular}
}
\end{table*}

\subsection{Tasks and Prompt Initialization}
\label{Appendix:Prompt Initialization}
We provide all initial prompts in Table~\ref{Prompt_Initialization} for tasks reported in Table \ref{tab:Task_Description} for TextGrad experiments. 
\begin{table*}[t!]
\centering
\caption{Prompt initializations for all tasks.}
\label{Prompt_Initialization}
\small
\begin{tabularx}{\textwidth}{lX}
\toprule
\textbf{Task} & \textbf{Initialization} \\
\midrule
TREC & Read the following question, then choose whether it is about a description,
entity, expression, human, location or number. \\
GSM8K & You will answer a mathemetical reasoning question. Think step by step. The last line of your response should be of the following format: 'Answer: \$VALUE' where VALUE is a numerical value.
 \\
MATH(Algebra) & You will answer a mathemetical reasoning question. Think step by step. The last line of your response should be of the following format: 'Answer: \$VALUE' where VALUE is a numerical value.
\\
IFBench & You are an expert at following precise instructions and constraints. When given a prompt with specific requirements, you must carefully: 1. Read and identify ALL constraints embedded in the prompt. 2. Plan your response to satisfy every constraint before writing. 3. Address each constraint explicitly in your response. 4. Double-check your response against every requirement before finalizing. Common constraint types include:- Word, sentence, or paragraph count requirements.- Keyword inclusion at specific frequencies.- Formatting requirements (bullets, sections, headings, etc.)- Language or style constraints.- Content structure requirements.- Repetition or reference constraints. Always prioritize constraint satisfaction over response quality or creativity. A response that follows all constraints perfectly is better than an eloquent response that misses any constraint.
\\
Arc Challenge & You will answer a multiple-choice science question. Respond with the label of the correct option (A–D or 1–4). 
The last line must be 'Answer: \$ X' where X is that label.
\\
HotpotQA(full wiki) & You are a precise multi-hop QA assistant. Use ONLY the context to answer the question concisely. Output ONLY the final answer. The last line of your response should be of the following format: 'Answer: \$STRING' where STRING is what the question is EXACTLY asking for.
\\
\bottomrule
\end{tabularx}
\end{table*}

\subsection{Templates}
\label{Appendix:Templates}
\newcommand{\tpl}[1]{\texttt{#1}}
For \emph{Promptwise} generation, we use the original DSPy instruction generation template unchanged. 
In iteration $\tau$, we shortlist seeding prompts $K$ with Gumbel–Top-$k$ and feed them to the template.

For \emph{Blockwise} generation, we employ a dedicated template that instructs \llmbackward\ to continue the current instruction by exactly $b$ tokens per block. 
Concretely, after applying Gumbel–Top-$k$ to obtain $\Pi_{\tau}^{\star}=\{\prompt_{\tau,1},\ldots,\prompt_{\tau,K}\}$, we \emph{uniformly sample} the number of width prompts from $\Pi_{\tau}^{\star}$ to generate each block $i$ of length $b$ (referred to as \texttt{more\_tokens} in the template). 
Thus, every block is produced by sampled prompts, and the template explicitly asks the model to “generate $b$ more tokens” continuing from the current text.
\begin{figure*}[htbp]
\begin{framed}
\noindent\textbf{Instruction Generation Prompt Template (Blockwise)}\\[1ex]

\small
You are an instruction optimizer for large language models. I will give some task instructions I've tried, along with their corresponding validation scores. The instructions are arranged in increasing order based on their scores, where higher scores indicate better quality.
Your goal is to CONTINUE the instruction from the exact last character of more\_tokens that will lead a good language model to perform the task even better. Don't be afraid to be creative. 
Do not restart, repeat, or reformat.
Only continue writing additional tokens that extend more\_tokens. In practice, we set more\_tokens as 50.

\vspace{0.3em}
\textbf{Attempted\_instructions:} \tpl{\{\{ attempted\_instructions = dspy.InputField() \}\}}

\vspace{0.3em}
\textbf{More\_tokens:} \tpl{\{\{ More\_tokens = dspy.InputField() \}\}}

\vspace{0.3em}
\textbf{Instruction:} \tpl{\{\{ proposed\_instruction = dspy.OutputField(desc="The improved instructions for the language model. One-paragraph instruction. Less than 120 words.") \}\}}

\vspace{0.3em}
\textbf{Prefix:} \tpl{\{\{ proposed\_prefix\_for\_output\_field = dspy.OutputField(desc="The string at the end of the prompt, which will help the model start solving the task. A short prefix. Less than 10 words.") \}\}}
\end{framed}

\newpage
For reference, we also provide the original instruction-generation template below.

\begin{framed}
\noindent\textbf{Instruction Generation Prompt Template (Original DSPy)}\\[1ex]

\small
You are an instruction optimizer for large language models. I will give some task
instructions I've tried, along with their corresponding validation scores. The
instructions are arranged in increasing order based on their scores, where higher scores
indicate better quality. Your task is to propose a new instruction that will lead a good
language model to perform the task even better. Don't be afraid to be creative.

\vspace{0.6em}
\textbf{Attempted\_instructions:} \tpl{\{\{ attempted\_instructions = dspy.InputField() \}\}}

\vspace{0.4em}
\textbf{Instruction:} \tpl{\{\{ proposed\_instruction = dspy.OutputField(desc="The improved instructions for the language model") \}\}}

\vspace{0.4em}
\textbf{Prefix:} \tpl{\{\{ proposed\_prefix\_for\_output\_field = dspy.OutputField(desc="The string at the end of the prompt, which will help the model start solving the task") \}\}}
\end{framed}
\end{figure*}
\clearpage

\subsection{EXAMPLES OF BEST PROMPTS FOR BENCHMARK}
\label{subsection:best_prompts}
Below we show best prompts selected by the highest validation set accuracy for \textgrad (with and without validation revert). 
Note here we fix seed as 1 and report the validation accuracy and test accuracy for all tasks and optimizers for fair comparison.
We report even for the same prompt the validation accuracy and test accuracy might vary by ~1\% difference with temperature as 0 due to nondeterminism in LLM inference \cite{tm-defeating-nondeterminism}. 
The original setup in \citet{yuksekgonul2024textgrad} is run with validation revert and we observe that on all tasks, the best prompt selected by the highest validation accuracy would stay the same as the initial prompt always has the highest validation accuracy (i.e. pure exploitation).  
\begin{framed}
\noindent\textbf{\strut\parbox{\linewidth}{\textbf{Task:} \textgrad \, Trec \hfill \textbf{Model:} GPT-4o-mini + GPT-4o}}\\[1ex]

\textbf{Sample Question:} \emph{x: How did serfdom develop in and then leave Russia ?
y: description,}\\[2pt]
\par\smallskip

\Base{\textbf{Base Template instruction (with validation accuracy 74.22 and test accuracy 77.0 and 35 tokens):}} 
Read the following question, then choose whether it is about a description, entity, expression, human, location or number.
\par\smallskip

\TGwith{\textbf{TextGrad (w/ revert) (with validation accuracy 74.22 and test accuracy 77.0 and 35 tokens):}} Read the following question, then choose whether it is about a description, entity, expression, human, location or number.
\par\smallskip

\TGwithout{\textbf{TextGrad (w/o revert) (with validation accuracy 82.03 and test accuracy 82.8 and 760 tokens):} } Identify the main category of the question from the given options: description, entity, expression, human, location, or number. Provide the category directly without additional context or explanation. - **Task Objective**: Choose the category that best describes the main subject of the question.- **Category Definitions**: - **Description**: Questions seeking detailed explanations or historical context. *Example: \"Why is the sky blue?\"* Edge Case: Questions about functions or roles, e.g., \"What function does homeostasis have on the existence of an organism?\"  - **Entity**: Questions about specific non-human objects, concepts, or works of art, including contact information. *Example: \"What is the capital of France?\"* Edge Case: Questions about books or artworks, e.g., \"What John Steinbeck novel portrays the Joad family?\"  - **Expression**: Questions about phrases or idioms. *Example: \"What does 'break the ice' mean?\"*  - **Human**: Questions about people, individuals, roles, or characters. *Example: \"Who is the current president of the United States?\"*  - **Location**: Questions about places, cities, or regions. *Example: \"Where is the Eiffel Tower located?\"*  - **Number**: Questions involving numerical data, quantities, or specific dates. *Example: \"How many continents are there?\"* Include dates and birth years, e.g., \"When was Babe Ruth born?\"- **Handling Ambiguity**: In cases of ambiguity, prioritize the category that most directly relates to the question's main subject. Use a detailed decision tree approach to resolve ambiguities, such as prioritizing \"human\" for roles or names, and \"entity\" for contact information. Example: If a question involves both a person and a date, prioritize \"number\" for birthdates. Provide a step-by-step example or flowchart for constructing this decision tree.- **Consistency and Synonyms**: Ensure consistency by recognizing synonyms and related terms that align with the chosen category. Maintain and dynamically update a list of synonyms, such as \"amount\" and \"quantity\" for \"number\", and \"name\" and \"character\" for \"human\", based on feedback. Implement a mechanism for suggesting new synonyms during categorization, using recent categorizations and user feedback.- **Feedback Loop**: Use past categorizations to refine your understanding and improve decision-making. Log categorizations and review them periodically to identify patterns or errors, especially in numerical and ambiguous contexts. Implement specific metrics for evaluating past performance, such as accuracy rates for each category or error rates in ambiguous contexts, and set improvement goals.- **Output Format**: Output the category as a single word string. Reinforce this format with examples of correct and incorrect outputs, especially in edge cases.- **Minimalism with Clarity**: Use minimal text to convey the category, ensuring clarity and precision. Verify categorizations internally before outputting. Introduce a verification step to ensure alignment with the question's main subject, specifying checks or criteria for verification.- **Numerical Precision**: Emphasize accuracy in numerical contexts, considering numerical reasoning and calculations when applicable. For questions involving dates or quantities, prioritize numerical reasoning. Suggest using external tools or APIs for complex calculations, and provide examples of when and how to use these tools.- **Error Analysis**: Reflect on decision-making processes and adjust based on past errors to improve future performance. Categorize errors by type and implement targeted strategies for each, such as misclassification or synonym recognition. Provide a framework for analyzing common misclassifications, detailing how errors are categorized and addressed.- **Contextual Awareness**: Consider the broader context of the question, including surrounding text or previous questions, to better understand the main subject. Use contextual cues to guide categorization, such as recognizing names or characters as \"human\". Specify techniques for identifying relevant context, such as keyword extraction or context windowing.- **Iterative Improvement**: Encourage iterative improvement by setting specific goals for accuracy and incorporating user feedback into the learning process. Establish a formal process for setting and reviewing these goals, such as regular performance reviews or user feedback sessions. Use SMART criteria for goal-setting, providing examples of specific, measurable, achievable, relevant, and time-bound goals.- **Robustness to Query Variations**: Ensure robustness to variations in query phrasing by training on diverse phrasings and contexts. Introduce a mechanism for asking clarifying questions when faced with ambiguous inputs, specifying when and how to generate these questions, along with examples.
\par\smallskip
\TSGDPromptwise{\textbf{TSGD-M (Promptwise)
(with validation accuracy 84.77 and test accuracy 86.0 and 515 tokens):}}
Read the question and classify it into one of the following categories: description, entity, expression, human, location, or number. Follow these streamlined steps to ensure accurate classification:1. **Identify Key Elements**: Focus on the main subject, action, and specific details in the question. Consider any implicit assumptions or common knowledge that might influence the category.2. **Contextual Analysis**: Analyze the broader context and any implicit clues within the question. Recognize synonyms or related terms that might imply the same category.3. **Category Definitions**: Use the following refined definitions to guide your classification:   - **Description**: Questions seeking detailed explanations or characteristics. Example: "What is a virtual IP address?\"   - **Entity**: Questions about specific objects, people, or concepts that can be named or identified. Example: \"What was the first domesticated bird?\"   - **Expression**: Questions involving phrases or idiomatic expressions. Example: "What does 'break the ice' mean?"   - **Human**: Questions specifically about individuals or groups of people, especially in roles or titles. Example: "Who created 'The Muppets'?"   - **Location**: Questions about places or geographical areas. Look for terms like "city," "state," or "country."   - **Number**: Questions involving numerical values or calculations, including dates or quantities. Pay attention to temporal indicators like "when."4. **Ambiguity Handling**: If the question is ambiguous, note the ambiguity and suggest possible clarifications. Provide a ranked list of potential categories if necessary.5. **Efficiency and Precision**: Provide a direct and concise response with the category label only. Ensure your response aligns with the expected category label exactly as provided in the options.6. **Feedback and Learning**: Implement a feedback mechanism to learn from past classification errors. Review previous misclassifications and the correct category to refine decision-making.7. **Example-Driven Learning**: Include more examples and analogies to illustrate the reasoning process for each category. This will help the model better understand the nuances of each classification and apply them more effectively.8. **Interactive Learning Component**: Engage in a dialogue to clarify uncertainties, asking follow-up questions or seeking additional information to ensure accurate classification.9. **Explicit Numerical Focus**: Pay special attention to numerical indicators and prioritize numerical reasoning when classifying questions related to numbers. Verify numerical classifications by considering alternative interpretations or performing a quick mental check.10. **Synonym Recognition and Flexibility**: Actively search for synonyms or related terms during classification, using a dynamic thesaurus or external linguistic resources to enhance understanding.11. **Incorporate Specificity in Responses**: When the question requires a specific answer, ensure to name the exact entity or fact. Provide a concise explanation for your choice, highlighting the key elements that led to this decision.12. **Cultural and Contextual Nuances**: Consider cultural or contextual nuances that may influence the interpretation of the question, ensuring your response is complete and coherent. By following these guidelines, you can enhance the accuracy and reliability of your classification decisions.
\par\smallskip
\TSGDBlockwise{\textbf{TSGD-M (Blockwise) (with validation accuracy 83.59 and test accuracy 85.4 and 270 tokens):}} Classify the given question into one of the following categories: description, entity, expression, human, location, or number. - A question about a \"description\" seeks to explain or characterize a general concept or thing, focusing on its nature or function. Example: \"What is Tyvek?\" asks for a description of Tyvek's characteristics and uses. - A question about an \"entity\" seeks to identify a specific non-human person, place, organization, or thing, such as a historical landmark or company. Example: \"What is the rarest coin?\" seeks to identify a specific coin, making it an entity.- A question about an \"expression\" involves evaluating or calculating a mathematical or logical expression. Example: \"What is HDLC?\" might require understanding technical details or calculations.- A question about a \"human\" asks to identify or provide information about a person, including their name or role. Example: \"What are the first names of the famous husband-and-wife acting team of Lunt and Fontanne?\" - A question about a \"location\" asks for information about a place, geographic area, or virtual space, such as an internet site. Consider the geographical or spatial context. Example: \"What is the most useful site on the Internet?\" focuses on the virtual space of the internet, classifying it as a location. - A question about a \"number\" asks for a specific numerical value or count. Carefully read the question, consider any technical, descriptive, entity-specific, or spatial aspects, and determine which category it belongs to. Provide your reasoning and then give the final classification. 
\par\smallskip
\GEPA{\textbf{GEPA (with validation accuracy 86.0 and test accuracy 84.0 and 450 tokens):}} You are tasked with classifying questions into one of the specified TREC categories: description, entity, expression, human, location, or number. Your objective is to accurately identify the category most appropriate for the question based on its content and the type of information it seeks.

\#\#\# Task Description:

For each question provided, follow these steps to classify it accurately:

1. **Description:**
   - Target questions that seek explanations, definitions, descriptions of concepts, things, processes, or phenomena.
   - Example: "What causes the body to shiver in cold temperatures?"

2. **Entity:**
   - Choose this category for questions identifying specific entities such as people, places, organizations, or things. 
   - The question should refer to a specific name or noun that identifies an entity.
   - Example: "What is the capital of France?" (Answer: Paris)

3. **Expression:**
   - Apply this category to questions asking to evaluate, calculate, or solve mathematical or logical expressions.
   - Example: "What is the result of 2 + 3?"

4. **Human:**
   - Use this category for questions directly inquiring about individuals, including names or roles.
   - Example: "Who was the inventor of silly putty?"

5. **Location:**
   - Select this category for questions asking about geographical areas or places.
   - Example: "Where is the Sahara Desert?"

6. **Number:**
   - Reserve this category for questions seeking specific numerical information such as dates, quantities, and measurements.
   - Example: "What is the population of Japan?"

\#\#\# Guidelines for Classification:

1. **Read Carefully:** Thoroughly understand the question before categorizing it.

2. **Identify Information Type:** Determine the type of answer the question seeks (description, name, number, etc.).

3. **Consider Definitions:** Match the question against each category definition to find the best fit.

4. **Logical Reasoning:** Justify your classification with a clear rationale based on the definitions.

5. **Provide the Category Name:** Respond only with the category name. Ensure the last line states the chosen category in the format: 'Answer: X', where X is one of the category names (description, entity, expression, human, location, number).

6. **Avoid Overlap:** Carefully consider questions that may fit multiple categories, selecting the most appropriate and specific category.
\end{framed}

\begin{framed}
\noindent\textbf{\strut\parbox{\linewidth}{\textbf{Task:} \textgrad \, ARC-Challenge \hfill \textbf{Model:} GPT-4o-mini + GPT-4o}}\\[1ex]

Sample Question: \emph{x: {George wants to warm his hands quickly by rubbing them. Which skin surface will produce the most heat? "text": [ "dry palms", "wet palms", "palms covered with oil", "palms covered with lotion" ], "label": [ "A", "B", "C", "D" ] }
y: A}\\[2pt]

\Base{\textbf{Base Template instruction (with validation accuracy 92.31 and test accuracy 91.33 and 50 tokens):}} You will answer a multiple-choice science question. Respond with the label of the correct option (A–D or 1–4). The last line must be 'Answer: X' where X is that label.\\
\TGwith{\textbf{TextGrad (w/ revert) (with validation accuracy 92.31 and test accuracy 91.33 and 50 tokens):}} You will answer a multiple-choice science question. Respond with the label of the correct option (A–D or 1–4). The last line must be 'Answer: X' where X is that label.\\
\TGwithout{\textbf{TextGrad (w/o revert) (with validation accuracy 92.31 and test accuracy 91.33 and 480 tokens):}} You will answer a multiple-choice science question. These questions are based on general science knowledge and reasoning. Follow these steps:\\1. **Evaluation**: Begin by evaluating each option thoroughly, prioritizing the elimination of those that contradict fundamental scientific principles. Select the option that aligns best with scientific principles or common knowledge. Provide a brief summary of the reasoning process for each option evaluated, even if the option is incorrect.\\2. **Response Format**: After evaluation, respond with the label of the correct option (A–D or 1–4). Ensure your answer matches the format of the choices provided. The last line must be: 'Answer: X'.\\3. **Confidence Assessment**: Internally evaluate your confidence in the chosen answer using a confidence scale (high, medium, low). If confidence is low, provide a brief note and flag the response for review.\\4. **Handling Ambiguity**: Proactively identify and address any potential ambiguities in the question or options. Provide a brief explanation of the ambiguity and proceed with the most likely answer. Prioritize addressing ambiguities that most affect the correctness of the answer.\\5. **Critical Thinking**: Engage in critical thinking by considering alternative interpretations of the question or options. Document and evaluate potential alternatives internally.\\6. **Cross-Verification**: Cross-verify your reasoning with known scientific principles or facts. Reference reliable scientific sources or databases if needed. Clearly reference specific criteria or data points used in the analysis to ensure precision and reduce ambiguity.\\7. **Feedback Mechanism**: Be open to user feedback and acknowledge any corrections or suggestions for refining future responses. Incorporate feedback into future responses through a brief internal review process. Review past responses and incorporate lessons learned to continuously improve the accuracy and clarity of future answers.\\8. **Conciseness and Clarity**: Use complete sentences and avoid ambiguity if an explanation is required. Encourage conciseness in explanations to ensure clear and direct answers. Use a conversational tone and break down complex information into simpler terms to make the response more engaging and accessible.\\By following these steps, you will consistently provide accurate, clear, and well-reasoned answers to multiple-choice science questions.\\
\TSGDPromptwise{\textbf{TSGD-M (Promptwise) (with validation accuracy 92.31 and test accuracy 93.20 and 370 tokens):}} You will answer a multiple-choice science question. These questions are based on general science knowledge and reasoning. Follow these guidelines to ensure clarity, accuracy, and educational value: 1. **Evaluate Options and Justify**: Carefully consider each option and select the one that best fits the question based on scientific reasoning. Provide a concise explanation for your choice, directly linking it to key scientific principles. Ensure explanations are succinct and directly support the answer. 2. **Handle Uncertainty with Confidence**: If uncertain, choose the option that is most logically consistent with the question's context. Clearly state any assumptions made and provide a confidence score or rationale to indicate the certainty of your response. 3. **Maintain Response Format**: Adhere strictly to the format: 'Answer: X'. Ensure consistency in this format across all responses to maintain clarity and uniformity. 4. **Use Clear and Engaging Language**: Use clear and precise language to avoid ambiguity. Avoid technical jargon unless necessary, and consider using examples or analogies to make explanations more relatable and engaging. Prioritize brevity and clarity to ensure the main point is communicated effectively. 5. **Adapt to Question Formats**: Adapt your approach if the question format varies slightly, focusing on the core scientific principles involved. Provide examples of how to adapt to different question formats when necessary. 6. **Implement Error Handling**: Identify and address potential errors or ambiguities in the question. Provide examples of common errors and how to address them. Review your response for ambiguous terms and provide clarifications to prevent misinterpretation. 7. **Incorporate Feedback and Learn**: Implement a feedback loop to reflect on your reasoning process and learn from any mistakes. Use user feedback to refine predictions and improve accuracy. Consider alternative answers and why they were not chosen to enhance future decision-making.8. **Enhance Educational Value**: When applicable, provide additional context or interesting facts related to the question to enhance the educational value of your response. By following these guidelines, you will ensure that your responses are accurate, clear, and contextually relevant, aligning more closely with the objective function. 
\par\smallskip
\TSGDBlockwise{\textbf{TSGD-M (Blockwise) (with validation accuracy: 92.64 and test accuracy: 92.60 and 460 tokens):}} Answer a multiple-choice science question by providing the correct option label (A–D or 1–4). Ensure your response is concise and accurate, formatted as 'Answer: X'. Use logical reasoning to support your answers. identify areas for improvement and incorporate feedback to refine your approach. Verify calculations and scientific principles to ensure accuracy. If you detect an error in the question or choices, provide the best possible answer and note the error. Maintain a balance between brevity and informat for improvement in conciseness and formatting. Strive to make your responses educational by explaining underlying principles or concepts, thereby enhancing the user's understanding. Encourage continuous improvement by reflecting on feedback and adapting your responses over time. By following these guidelines, you willUse simple language to ensure clarity and accessibility for all audiences. yo counter ambiguous or poorly worded questionshos the most likely answer based on abinformation and provid a brief note on the ambiguity Reularly review past responses to identify areas for improvement and incorporate feedback to refine your approachTstyour responses aainst a range of similar questions to ensure reliability and consistencyand consistency. Organize your approach into clear sections: "Strtegy," "Confidence Indication," "Error Handling," and "Continuous Irovment." This structure will help you systematically apply the instructions. If the answer is not selfvd, prvide aief explanatin to enhance understanding. Indicate your confidence le using  simpe scalee.g.,lw,medium, highh gauge the certainty of your response Regularly test your responses across a variety ofUse relatable analogies or simple language to make complex concepts more accessible and engaging. If you encounter ambiguous or poorly worded questions, choose the most likely answer based on available information and provide a brief note on the ambiguity. Regularly review past responses to identify areas for improvement in conciseness and formatting. Strive to make your responses educational by explaining underlying principles or concepts, thereby enhancing the user's understanding. cntinuous improvement by re improve your approach by focusing on clarity, accuracy, and engagement. This approach will help you adapt to new challenges and enhance your overall effectiveness. Use logical reasoning to support your answers and cross-verify with available data. By focusing on these strategies, you will enhance your ability to provide clear, concise, and accurate answers. Encourage continuous improvement by reflecting on feedback and adapting your responses over time. By following these guidelines, you will enhance your ability to make complex concepts more understandable and ensure your responses are both informative and engaging for the user.
\par\smallskip
\GEPA{\textbf{GEPA (with validation accuracy 93.0 and test accuracy 91.0 and 420 tokens):}} You will answer a multiple-choice science question using a detailed task-oriented approach. Carefully adhere to the following instructions to provide accurate and well-reasoned responses:

1. **Read Comprehensively:** Begin by carefully reading the entire question and all provided answer choices. Ensure you understand the question being asked and the context of each option.

2. **Identify the Core Inquiry:** Determine exactly what the question is addressing. Pay attention to keywords and contextual clues to understand the specific scientific concept involved.

3. **Evaluate Options Thoroughly:** Analyze each option:
   - **For Environmental and Laboratory Safety**: Consider the implications of hygiene, contamination, and proper disposal methods. Prioritize safe handling and disposal to prevent contamination.
   - **For Biological Systems**: Recognize the hierarchical organization of biological structures, such as cells, tissues, organs, and systems, to determine the level of organization being referenced.
   - **For Earth and Space Science**: Distinguish between temporal phenomena on different scales, such as daily, monthly, and seasonal changes, understanding the rotation and revolution of celestial bodies.

4. **Apply Scientific Principles:** Use relevant scientific principles to guide your reasoning:
   - **Genetics**: Consider dominance patterns, incomplete dominance, and the implications for trait inheritance.
   - **Graphical Data Representation**: Match data types to appropriate graph types, such as pie charts for parts of a whole.
   - **Physics and Vectors**: Analyze vector components and resultant forces or movements.

5. **Reason and Conclude:** Provide a concise explanation of your reasoning, focusing on the correct application of scientific principles to determine the correct answer. Ensure your reasoning aligns with the question's context and logical deductions.

6. **Respond Succinctly:** Conclude your response with only the letter of the correct choice (A, B, C, or D). Format your final answer as follows: 'Answer: X', where X is the letter corresponding to the correct option based on your analysis.

By following these step-by-step instructions, ensure your responses are accurate, precise, and demonstrate a clear understanding of the scientific concepts involved.

\end{framed}

\begin{framed}
\noindent\textbf{\strut\parbox{\linewidth}{\textbf{Task:} \textgrad \, GSM8K \hfill \textbf{Model:} GPT-4o-mini + GPT-4o}}\\[1ex]

Sample Question: \emph{x: Natalia sold clips to 48 of her friends in April, and then she sold half as many clips in May. How many clips did Natalia sell altogether in April and May?
y: 72}\\[2pt]

\Base{\textbf{Base Template instruction  (with validation accuracy 93.00 and test accuracy 93.86 and 55 tokens):}} You will answer a mathemetical reasoning question. Think step by step. The last line of your response should be of the following format: 'Answer: \$VALUE' where VALUE is a numerical value.\\
\TGwith{\textbf{TextGrad (w/ revert) (with validation accuracy 93.00 and test accuracy 93.86 and 55 tokens):}} You will answer a mathemetical reasoning question. Think step by step. The last line of your response should be of the following format: 'Answer: \$VALUE' where VALUE is a numerical value.\\
\TGwithout{\textbf{TextGrad (w/o revert) (with validation accuracy 94.67 and test accuracy 92.57 and 960 tokens):}} You will answer a mathematical reasoning question. Focus on delivering the final numerical answer succinctly, formatted as 'Final Answer: VALUE' with the numerical value clearly visible. Prioritize brevity and directness, especially for straightforward problems, by providing the final answer directly without additional context or breakdowns. Use bullet points or numbered lists to maintain clarity for multi-step problems, and break down complex problems into smaller, manageable parts, explaining each step logically with necessary assumptions and formulas. Clearly define all numerical values and their significance within the problem context to prevent ambiguity. Use simple mathematical notation in plain text to ensure accessibility and maintain consistency in units. Include units in the final answer only if explicitly required by the question. Eliminate non-essential details or steps that do not directly contribute to deriving the final answer. Implement a primary verification mechanism for the final answer, using a secondary method only if needed, and perform a verification step after completing calculations to ensure accuracy. Specify what constitutes a \"secondary method\" for verification, providing examples or guidelines, such as cross-verifying with a different formula or using estimation techniques. Identify and mention common pitfalls or errors, providing strategies to avoid them, and incorporate examples of such pitfalls, like misinterpreting units or overlooking edge cases. Offer a general approach or formula when applicable, to enable users to apply the solution to similar problems. Engage the user for clarifications only when the problem context is ambiguous, and include strategies for identifying ambiguity, such as looking for missing information or unclear terms. Ensure the response is contextually relevant and accurately reflects the specific scenario presented. Emphasize accuracy in calculations by double-checking each step and ensuring the final result aligns with expected outcomes. Specify the level of precision required for the final answer, including rounding rules if necessary, and provide guidelines on determining the appropriate level of precision, with examples. Use consistent terminology and formatting throughout the response, aligning with the expected ground truth, and provide specific examples or templates for common mathematical terms and formats. Implement a final verification step to ensure the output format and numerical value align with the expected ground truth, correcting any discrepancies before finalizing the response. Focus on delivering precise and concise information, omitting additional context unless it directly contributes to the accuracy of the final answer. Explicitly declare all assumptions made during the problem-solving process, emphasizing their importance. Engage in a feedback loop to learn from past errors and adjust strategies for continuous improvement, maintaining a log of common errors and successful strategies for future reference, and specify how to prioritize learning from these logs, focusing on the most frequent errors or those with the highest impact on accuracy. Ensure the final answer is clearly stated at the end of the response. If the problem is simple and the final answer is evident, provide the answer directly without additional explanation. Use a consistent format for the final answer, such as \"Answer:\" followed by the numerical value. Implement intermediate checks after each calculation step to catch potential errors early. Consider potential edge cases or variations in the problem and prepare to handle them, offering examples of common edge cases in mathematical reasoning and strategies for addressing them. Provide a brief summary or conclusion that reiterates the final answer and its significance, focusing on the most challenging aspects of the problem or the most significant takeaways. Adapt the complexity of your response based on the problem's difficulty, providing direct answers for simple problems and detailed breakdowns for complex ones. Use illustrative examples to clarify complex steps and ensure all variables and terms are explicitly defined to eliminate ambiguity. Provide a brief introduction that contextualizes the problem and outlines the steps to be taken. Use analogies or relatable examples to make explanations more engaging. Maintain a log of common errors and successful strategies, focusing on learning from the most frequent or impactful mistakes. Begin with a brief contextual overview of the problem scenario to enhance user comprehension. Conclude with a reflective summary that emphasizes the most challenging aspects of the problem and the key takeaways. Prioritize brevity and directness for simple problems, providing only the final numerical answer without additional context or breakdown. Ensure the final answer is presented as a standalone number to facilitate direct comparison and reduce parsing requirements. Include a brief contextual introduction to the problem, summarizing key elements such as initial conditions and relevant details. Conclude with a sentence that not only states the final answer but also reiterates the key findings and the significance of the solution. Implement a mechanism for error detection and correction by comparing the final answer with expected values or using alternative methods for verification. Ensure clarity and conciseness by eliminating unnecessary steps and providing self-contained responses. Adapt responses based on problem complexity, and incorporate a feedback loop for continuous improvement. Emphasize the importance of providing complete and explicit reasoning for each step in the problem-solving process, avoiding ellipses or shorthand. Maintain a clear and logical flow with transitions between ideas, and ensure consistent variable usage. Include a verification or justification step for the final answer, and avoid ambiguous language by clearly stating assumptions and external information. Encourage learning from past interactions by maintaining a log of feedback and incorporating it into future responses. Ensure the final answer format aligns with the expected ground truth, omitting currency symbols or decimal places unless specified. Explicitly state each step in the calculation process, including the context and significance of each numerical value. Identify the primary focus of the question and tailor the response to address that aspect directly. Use language that directly corresponds to the question's focus, such as 'Total cost' or 'Total number of stickers,' to ensure clarity and alignment with the expected answer format. Implement a verification step to compare the final answer with the expected format and value, and adjust if discrepancies are found. Maintain a log of feedback and incorporate it into future responses to continuously improve accuracy and alignment with expected outcomes. \\
\TSGDPromptwise{\textbf{TSGD-M (Promptwise) (with validation accuracy 94.67 and test accuracy 94.39 and 265 tokens):}} You will answer a mathematical reasoning question. Begin by carefully reading and interpreting the question, ensuring you fully understand the problem statement. Explicitly introduce and define any variables used in your calculations. Provide a concise explanation, focusing on key steps, and ensure calculations are simplified. Use clear and consistent mathematical notation, breaking down complex equations into simpler parts. Present the final answer directly in the format: 'Answer: VALUE' without any additional text or symbols. Avoid redundancy and ensure each step adds unique value. Verify your calculations and logic before finalizing the answer, performing a final verification check to ensure accuracy and consistency with the input data. If the question involves multiple steps or data points, offer a brief explanation of the calculation process to enhance understanding, but prioritize conciseness. Implement checks to identify potential errors in input data or calculations, using alternative methods or estimation techniques for cross-verification. If an error is detected, address it before providing the final answer. Ensure the final answer matches the expected format, such as presenting numbers without additional symbols unless specified. If the input query is ambiguous, request clarification or make reasonable assumptions, clearly stating and justifying them in the response. Pay attention to contextual clues such as discounts or special conditions that might affect the calculation. Example: For the question 'What is 2 + 2?', respond with 'Answer: 4'. \\
\TSGDBlockwise{\textbf{TSGD-M (Blockwise) (with validation accuracy: 93.67 and test accuracy: 93.71 and 330 tokens):}} You will answer a mathematical reasoning question. Follow these guidelines to ensure clarity, accuracy, and completeness:\\1. **Step-by-Step Reasoning**: Break down the problem into clear, logical steps. Articulate each step fullythat ensue yur response includes all necessary details.\\2. **Verification and Reflection**: After dering th answer, verify it by substituting back into the problem to ensure consistency. Reflect on the reasoning process and consider potential errors or alternative approaches before finalizing the answer.\\3. **Conciseness and Clarity**: Provide clear andtt format: 'Answer: VALUE', where VALUE is a plain numerical result. Ensure it matches the format of the ground truth, avoiding unnecessary symbols or decimal precision.\\5. **Error Handling**: Identify and address potential ambiguities or errors in the inputContxtual Adaptability**: Tailor your explanation to the contexttqueston, ensuring that your response is relevant and understandheuseof Visual Aids**: If applicable, use diagrams or visual aids to simplify complex problems and provide additional clarity.\\8. ** and Interaction**: Engage with the user by providing additional context or explanations when necessary to enhancethat ensue yur response includes all necessary details.\\2. **Verification and Reflection**: After dering th answer, verify it by substituting back into the problem to ensure consistency. Reflect on the reasoning process and consider potential errors or alternative approaches beforeSimplification and Language Alignment**: Simplify language to align with the user's level of understanding, ensuring clarity and preventing misunderstandings. dcta quick error analysis to ensure that it reains effective in diverse scenarios.\\10. **Continuous Learning**: Encourage the model to learn from siteactinsby incorratigfeedckan refining strategies over time. This will enhance the model's alyo dliver accurate and contextually relevant answers. This structured approach will help the language model deliver precise, concise, and contextually relevant answers, ultimately improving the accuracy and engagement of its responses to mathematical reasoning questions.
\par\smallskip
\GEPA{\textbf{GEPA (with validation accuracy 93.67 and test accuracy 91.33 and 500 tokens):}}
Your task is to solve math problems by following a detailed, systematic, and meticulous approach to ensure clarity and accuracy in your response. The problems presented will involve arithmetic or algebra, typically framed within a real-life scenario. Therefore, your solution must not only address numerical accuracy but also contextual relevance.

1. **Comprehensive Problem Reading and Understanding**:
   - Carefully read the problem to determine what is being asked.
   - Identify key details and quantities provided in the problem statement, paying close attention to order, operations, and any specified relationships between quantities.

2. **Methodical Problem Solving**:
   - Break down the problem into smaller, manageable steps.
   - Start by identifying known values and relate them to the unknowns, using algebraic expressions or straightforward arithmetic as necessary.
   - Clearly document your mathematical reasoning as you progress from one step to the next.

3. **Logical Structure and Calculation**:
   - Use algebraic manipulations where necessary, ensuring you carefully perform arithmetic operations.
   - Execute calculations in a coherent sequence to avoid logical errors.
   - If applicable, ensure you respect relationships or multiples described in the problem (e.g., "twice the amount", "triple the value").

4. **Precision and Accuracy Assurance**:
   - Double-check all calculations, considering both intermediate steps and the final answer.
   - Ensure that units of measure or other constraints are correctly applied and adjusted throughout the calculations if relevant.

5. **Clear and Structured Presentation**:
   - Present each step distinctly to reflect the logical progression of your solution.
   - Format your final answer by writing it on a new line, prefaced by 'Answer: $VALUE', ensuring that $VALUE is a singular numerical outcome of the solved problem.

6. **Validation Against Problem Requirements**:
   - Post-process the problem statement to verify all components have been satisfactorily addressed.
   - Ensure the resulting solution matches the problem's requirements and meets any specified conditions or constraints.

7. **Application of Context-Specific Knowledge**:
   - Recognize the context or scenario being described and apply relevant domain knowledge as needed.
   - Adjust your approach considering scenario-specific facts such as division of quantities among people, proportions, or frequency of actions.

Remember, each problem might appear unique, but the fundamental principles of calculation accuracy, logical reasoning, and context consideration remain constant. Utilize this structured approach to provide comprehensive and precise solutions.

\end{framed}

\begin{framed}
\noindent\textbf{\strut\parbox{\linewidth}{\textbf{Task:} \textgrad \, MATH(algebra) \hfill \textbf{Model:} GPT-4o-mini + GPT-4o}}\\[1ex]

Sample Question: \emph{x: Let \[f(x) = \left\{ \begin{array}{cl} ax+3, &\text{ if }x>2, \\ x-5 &\text{ if } -2 \le x \le 2, \\ 2x-b &\text{ if } x <-2. \end{array} \right.\]Find $a+b$ if the piecewise function is continuous (which means that its graph can be drawn without lifting your pencil from the paper).
y: 0}\\[2pt]

\Base{\textbf{Base Template instruction  (with validation accuracy 84.00 and test accuracy 85.42 and 45 tokens):}} You will answer a mathematical reasoning question. Think step by step. The last line of your response should be of the following format: 'Answer: \$VALUE' where VALUE is a numerical value.\\
\TGwith{\textbf{TextGrad (w/ revert) (with validation accuracy 84.00 and test accuracy 85.42 and 45 tokens):}} You will answer a mathematical reasoning question. Think step by step. The last line of your response should be of the following format: 'Answer: \$VALUE' where VALUE is a numerical value.\\
\TGwithout{\textbf{TextGrad (w/o revert) (with validation accuracy 86.00 and test accuracy 84.80 and 45 tokens):}} You will answer a mathematical reasoning question. Think step by step. The last line of your response should be of the following format: 'Answer: \$VALUE' where VALUE is a numerical value.\\
\TSGDPromptwise{\textbf{TSGD-M (Promptwise) (with validation accuracy 85.42 and test accuracy 86.65 and 215 tokens):}} You will answer a mathematical reasoning question. Follow these steps to ensure clarity and correctness:\\ 1. **Restate the Problem**: Begin by summarizing the question in your own words to ensure understanding.\\2. **Identify Variables and Conditions**: Clearly define all variables and note any specific conditions or constraints mentioned in the problem.\\3. **Step-by-Step Solution**: Break down the problem into smaller, manageable parts. For each step:\\   - Identify applicable mathematical properties or rules.\\   - Explain the reasoning and calculations clearly.\\   - Use consistent and clear mathematical notation.\\4. **Verification**: After finding the solution, verify it by substituting the value back into the original equations to ensure it satisfies all conditions.\\5. **Final Answer**: Present the final answer in the following format: 'Answer: VALUE' where VALUE is a numerical value. Ensure the answer is clear, concise, and directly related to the original question.\\6. **Avoid Ambiguity**: Do not use ambiguous symbols or language. Provide specific details and avoid redundancy in your explanation.\\ By following these guidelines, you will produce a response that is robust, clear, and reliable. \\
\TSGDBlockwise{\textbf{TSGD-M (Blockwise) (with validation accuracy: 85.71 and test accuracy: 86.65 and 115 tokens):}} You will answer a mathematical reasoning question. Start by restating the problem concisely to capture key information and assumptions. Define all variables and express relationships using equations. Solve the problem step-by-step, focusing on essential steps and avoiding unnecessary details. Use clear and consistent mathematical notation, and ensure all calculations are precise. After finding the to facilitate easy parsing and verification is a numerical value. Ensure the output is in plain text without additional formatting or symbols.
\par\smallskip
\GEPA{\textbf{GEPA (with validation accuracy 85.5 and test accuracy 82.5 and 560 tokens):}} Solve algebraic and mathematical problems step by step by following these detailed guidelines. For each task:
1. **Understand the Problem**: 
   - Carefully read the problem statement to ensure a clear understanding of what is being asked.
   - Analyze the objective of the problem (e.g., finding minimum/maximum values, solving equations).
2. **Identify Problem Components**:
   - Determine the type of problem (e.g., algebra, arithmetic, unit conversion).
   - Recognize any given formulas or expressions and note what needs to be calculated.
   - Identify necessary information such as coefficients in equations, constants, variable interactions, or terms in polynomial equations.
3. **Set Up Calculations**:
   - Write out any equations clearly, identifying the known and unknown quantities.
   - For algebra problems, consider various algebraic techniques like factoring, expanding, and using formulas like the quadratic formula when appropriate.
   - For unit conversion or cost analysis problems, clearly delineate the relations and conversions necessary.
4. **Execute Calculations**:
   - Perform each calculation step by step, maintaining a logical and organized approach.
   - Use precise calculations, especially for intermediate steps to minimize rounding errors.
   - Check for alternative methods to simplify or solve equations, and consider verifying by substituting back into original conditions where applicable.
5. **Verification and Consistency**:
   - Verify all steps against the conditions provided in the problem.
   - Ensure your mathematical manipulations align with the problem's requirements.
   - Double-check results for accuracy, particularly for calculations that require specific forms, like fractions or decimals.
6. **Answer Validation**:
   - Check each potential solution in the context of the original problem conditions.
   - Identify any constraints or extraneous solutions that might invalidate an algebraic manipulation.
7. **Formatting**:
   - Format the final answer in LaTeX if applicable, using '\texttt{\textbackslash frac\{numerator\}\{denominator\}}' for fractions when required.
   - Use decimals to the appropriate precision as asked in the problem (e.g., nearest tenth).
8. **Conclusion**:
   - Clearly summarize the final result, ensuring it meets the problem requirements.
   - Order the solutions from least to greatest when multiple solutions exist and as dictated by the problem statement.
   - Express integers precisely and round off results where specified (e.g., nearest tenth for decimal expressions).
By following these instructions, you can ensure a comprehensive and accurate response to algebraic and mathematical problems.
\end{framed}

\begin{framed}
\noindent\textbf{\strut\parbox{\linewidth}{\textbf{Task:} \textgrad \, HotPotQA \hfill \textbf{Model:} GPT-4o-mini + GPT-4o}}\\[1ex]

Sample Question: \emph{x: Which magazine was started first Arthur's Magazine or First for Women? Context: { "title": [ "Radio City (Indian radio station)", "History of Albanian football", "Echosmith", "Women's colleges in the Southern United States", "First Arthur County Courthouse and Jail", "Arthur's Magazine", "2014–15 Ukrainian Hockey Championship", "First for Women", "Freeway Complex Fire", "William Rast" ], "sentences": [ [ "Radio City is India's first private FM radio station and was started on 3 July 2001.", " It broadcasts on 91.1 (earlier 91.0 in most cities) megahertz from Mumbai (where it was started in 2004), Bengaluru (started first in 2001), Lucknow and New Delhi (since 2003).", ...} Details omitted
y: "Arthur's Magazine"}\\[2pt]

\Base{\textbf{Base Template instruction  (with validation accuracy 48.67 and test accuracy 53.00 and 45 tokens):}} 
You are a precise multi-hop QA assistant.  Use ONLY the context to answer the question concisely.Output ONLY the final answer. The last line of your response must be: 'Answer: \$STRING'.\\
\TGwith{\textbf{TextGrad (w/ revert) (with validation accuracy 48.67 and test accuracy 53.00 and 45 tokens):}} 
You are a precise multi-hop QA assistant.  Use ONLY the context to answer the question concisely.Output ONLY the final answer. The last line of your response must be: 'Answer: \$STRING'.\\
\TGwithout{\textbf{TextGrad (w/o revert) (with validation accuracy 49.33 and test accuracy 49.33 and 45 tokens):}} You will answer a reasoning question. Think step by step. The last line of your response should be of the following format: 'Answer: \$VALUE' where VALUE is a numerical value.\\
\TSGDPromptwise{\textbf{TSGD-M (Promptwise) (with validation accuracy 50.33 and test accuracy 51.67 and 190 tokens):}} "You are a precise and concise multi-hop QA assistant. \\ Use ONLY the context to answer the question, prioritizing context that directly relates to the entities in the question. \\ Identify key entities and perform keyword matching to filter relevant information. \\ Use multi-hop reasoning to connect different pieces of context that relate to the question. \\ Provide the shortest possible answer that directly addresses the question. \\ Do not include any additional context or explanations beyond what is necessary to answer the question. \\ Avoid repeating information unnecessarily; mention each relevant detail only once. \\ State the answer directly and succinctly, without elaboration. \\ Ensure the final answer is accurate and directly supported by the context. \\ The last line of your response must be: 'Answer: \$STRING'. \\
\TSGDBlockwise{\textbf{TSGD-M (Blockwise) (with validation accuracy 50.00 and test accuracy 54.00 and 390 tokens):}} You are a precise and concise multi-hop QA assistant. Use \textbf{only} the provided context to derive your answer. Provide a direct answer that matches the expected format.
\textbf{Core task.} Prioritize the most relevant information, analyze past errors, and adjust strategies. Allow for synonymous terms and alternative expressions. Ensure numerical data are interpreted correctly and match the expected format.
\textbf{Structured organization.} Organize instructions into distinct sections: ``Context Utilization'', ``Verification'', and ``Final Answer''. When you form the answer, ensure it aligns with the ground truth; check corroborating details within the context.
\textbf{Confidence scoring.} Assign a confidence score to indicate reliability. If the answer does not match the ground truth, learn from the error and improve.
\textbf{Disambiguation and ambiguity handling.} Use robust string matching to handle language variation. When ambiguous, provide short clarification or request more specific information. Pay attention to case, punctuation, and format so the output exactly matches the expected answer.
\textbf{Feedback loop.} Learn from feedback and prior errors; adjust strategies accordingly.
\textbf{Verification mechanism.} Before finalizing, verify with corroborating evidence in the context to ensure consistency and accuracy.
\textbf{Training and adaptation.} Use diverse examples to handle varied contexts effectively.
\textbf{Error analysis.} Analyze mistakes and refine strategies. Briefly explain discrepancies when helpful.
\textbf{Exact matches and synonyms.} Recognize synonyms, but prioritize exact matches when the ground truth requires a specific format. Keep responses consistent in format.
\textbf{Clarity and conciseness.} Remove redundancy; consolidate similar directives. Highlight key words or phrases tied to the question.
\textbf{Human review.} For low-confidence answers, suggest human review.
\textbf{Output.} Provide the shortest possible answer that directly addresses the question.
\textbf{Contextual reference.} When helpful, cite the specific context snippet that supports your answer.
\textbf{Handling multiple candidates.} If multiple possible answers appear, select the one that best matches the question’s criteria and briefly justify.
\textbf{Error logging.} Note cases where your prediction diverges from the ground truth and why.
\textbf{Final verification.} Re-check the answer against the context before finishing.
\par\smallskip
\GEPA{\textbf{GEPA (with validation accuracy 53.00 and test accuracy 48.00 and 550 tokens):}}
Task: Given the fields `question`, `summary\_1`, and `summary\_2`, produce the field `answer`.
Instructions:
1. **Understand the Question**: Carefully read the `question` field to determine what specific information is being asked. This could be a nickname, a full name, a population figure, or any other specific detail.
2. **Analyze the Summaries**: Examine both `summary\_1` and `summary\_2` to extract relevant information that directly addresses the question. Pay attention to any discrepancies or additional context provided in the summaries.
3. **Cross-Verification**: If the summaries provide conflicting information, prioritize the summary that offers more specific or detailed data. If both summaries agree, use the information as is.
4. **Factual Accuracy**: Ensure that the answer is factually accurate. If the summaries do not provide enough information to answer the question accurately, acknowledge the need for further research.
5. **Formatting the Answer**: 
   - If the question asks for a name, ensure the full and correct name is provided, including middle names or initials if specified in the summaries.
   - If the question asks for a nickname, provide it in the exact format as commonly recognized or as specified in the summaries.
   - If the question asks for a numerical value, such as a population, provide the number without additional text unless specified otherwise.
6. **Generalizable Strategy**: Use logical reasoning to deduce the answer from the given summaries. If the summaries lack specific information, indicate the need for additional research or context.
7. **Domain-Specific Knowledge**: 
   - For questions related to historical figures or events, ensure the answer reflects the most widely accepted historical facts.
   - For questions about geographical data, such as population, ensure the answer is based on the most recent and reliable data available in the summaries.
8. **Example Contexts**:
   - For a question about a nickname, ensure the answer matches the commonly known or documented nickname.
   - For a question about a person's full name, ensure the answer includes all known parts of the name as documented.
   - For a question about a population, provide the exact figure as stated in the summaries.
By following these instructions, you will be able to generate accurate and contextually appropriate answers based on the provided summaries.
\end{framed}

Below prompts are \dspy \, for DSPy modules. Note here we fix seed 1 and only report sample prompts from all 4 modules per one run.
\begin{framed}
\noindent\textbf{\strut\parbox{\linewidth}{\textbf{Task:} \dspy \, Trec \hfill \textbf{Model:} GPT-4o-mini + GPT-4o}}\\[1ex]

Sample Question: \emph{x: How did serfdom develop in and then leave Russia ?
y: description,}\\[2pt]

\Base{\textbf{Base Template instruction (with validation accuracy 77.34 and test accuracy 79.8 and 30 tokens):}} Read the following question, then choose whether it is about a description, entity, expression, human, location or number.
\par\smallskip
\TGwith{\textbf{\dspy \, (with validation accuracy 80.25 and test accuracy 78.8 and 115 tokens):}} Review each question thoroughly to determine its core subject and function. Assign to each the most fitting category, choosing among a description, entity, expression, human, location, or number. Evaluate the context beyond words analyzing not just what kind of query it is but also what is implied or explicitly requires attention, whether qualitative, identifying references, quoting sources, referencing individuals and roles, highlighting geographical places, or pursuing factual, quantitative information. Maintain precisive insight as you categorize each inquiry to ensure keen accuracy across thematic dimensions.
\par\smallskip
Prefix:
\par\smallskip
Identified Category: \par\smallskip
\TSGDPromptwise{\textbf{TSGD-M (Promptwise) (with validation accuracy 82.42 and test accuracy 83.0 and 118 tokens):}} Propose a user-centric approach to classify questions into six distinct categories: Description, Entity, Expression, Human Information, Location Identification, and Numeric Details. Begin by sparking natural comprehension of the question's intent by analyzing themes, syntactic structures, and emblematic senaitory fields. Develop structured mental maps to identify how emotional palette variances inoculate this progression. Next, innovate with technologies that automate reasoning behind human-centric path categorization, carving intellectual conduits for determination replay, repeatedly framing challenging readiness for scalability emotions media-consuming agriculture underscore commentary.
Prefix:Classification Class: [Autonomous Labeling Choice] with Intellectual Cognition process Sustainability: \par\smallskip
\TSGDBlockwise{\textbf{TSGD-M (Blockwise) (with validation accuracy 79.69 and test accuracy 80.6 and 160 tokens):}} Categorize each question into primary categories: description, entity, expression, human-being, location, or numerical detail. Examine them for keywords and underlying messages that reveal their true intent. Clearly articulate your classification by employing critical thinkingand analytical skills, while systematically reducing ambiguity. Aim for clarity and precision throughout the classification process, ensuring each determination increases understanding and captures the intricacies of language. Your classifications should encourage deeper engagement with the context behind each inquiry presented.
[[ \#\# proposedunderstanding and captures the intricacies of language. Ensure you highlight key aspects of each question, positioning your insights to be both relatable and informative. Strive for consistency through defined categorization while focusing on providing interpretations that genuinely mirror the intent behind queries acrosseach context. This classification should reinforce understanding not just for yourself, but also for those who may refer to your insights.</p> \par\smallskip
Prefix: Categorization and analysis of queries.
\par\smallskip
\end{framed}

\begin{framed}
\noindent\textbf{\strut\parbox{\linewidth}{\textbf{Task:} \dspy \, ARC-Challenge \hfill \textbf{Model:} GPT-4o-mini + GPT-4o}}\\[1ex]

Sample Question: \emph{x: {George wants to warm his hands quickly by rubbing them. Which skin surface will produce the most heat? "text": [ "dry palms", "wet palms", "palms covered with oil", "palms covered with lotion" ], "label": [ "A", "B", "C", "D" ] }
y: A}\\[2pt]

\Base{\textbf{Base Template instruction (with validation accuracy 91.97 and test accuracy 93.77 and 30 tokens):}} Given the fields "question", produce the fields "answer". Reasoning: Let's think step by step in order to\\
\TGwith{\textbf{\dspy \, (with validation accuracy 93.65 and test accuracy 93.85 and 160 tokens):}}  Refine and craft exceptionally articulate answers to a broad spectrum of questions, including factual, conceptual, and open-ended types. Begin by thoroughly analyzing both explicit and implicit aspects of the questions provided. Deliver your insights in precise detail, ensuring clarity with relevant context to enhance the audience's understanding. Strategically leverage existing knowledge to approach even the most complex inquiries with insightful, logic-supported answers. Aim for balance in fluency and depth in your explanations, effectively utilizing partial information when needed, and frequently refining your approach for optimal outcomes. Prioritize seamless and intuitive comprehension while fostering the incrementally enhanced learning of readers, focusing on maintaining their engagement with concise, rich context-building solutions.\\ Prefix:
Start the Response:\\
\TSGDPromptwise{\textbf{TSGD-M (Promptwise) (with validation accuracy 94.31 and test accuracy 94.37 and 100 tokens):}} Formulate a comprehensive and articulate response to the derived question, focusing on an extensive analysis and cohesive integration of pertinent information. Evaluate the question nuances, delving into diverse perspectives and assumptions, and uphold your arguments with context-relevant examples. Aim for an engaging response that harmonizes detail, sophistication, and clarity to thoroughly inform and captivate the reader. Enhance accessibility through memorable insights and examples that ensure precision and sustained reader interest.
\\
Prefix: Comprehensive and articulate response: \\
\TSGDBlockwise{\textbf{TSGD-M (Blockwise) (with validation accuracy 93.98 and test accuracy 94.97 and 190 tokens):}} Imagine you are an expert guide dedicated to nurturing understanding and curiosity in users. Your primary role is to intuitively address their queries by weaving elaborate cross-references and metaphors, providing creatively structured clarifications with comprehensive detail.Incorporate relatable examples that resonate with various experience levels and foster a collaborative atmosphere. Conclude your answers by inviting further exploration and suggesting actionable takeaways that promote independent learning and engagement. Your mission is to illuminate the topic at hand while empowering users withpractical knowledge and inspiring them to pursue their inquiries beyond your response. Remember to advocate for ongoing curiosity and present resources for users to deepen their understanding further.\\ 
Prefix: Foster inquiry and guide exploration. practical knowledge and inspiring them to pursue their inquiries beyond your response. Remember to advocate for ongoing curiosity and present resources for users to deepen their understanding further. Provide prompts or thought-provoking questions that encourage users to think critically about the topic...
\end{framed}

\begin{framed}
\noindent\textbf{\strut\parbox{\linewidth}{\textbf{Task:} \dspy \, GSM8K \hfill \textbf{Model:} GPT-4o-mini + GPT-4o}}\\[1ex]

Sample Question: \emph{x: Natalia sold clips to 48 of her friends in April, and then she sold half as many clips in May. How many clips did Natalia sell altogether in April and May?
y: 72}\\[2pt]

\Base{\textbf{Base Template instruction (with validation accuracy 86.67 and test accuracy 90.37 and 30 tokens):}} Given the fields "question", produce the fields "answer". Reasoning: Let's think step by step in order to\\
\TGwith{\textbf{\dspy \, (with validation accuracy 80.25 and test accuracy 87.79 and 145 tokens):}} Instruction:Aimoto create an answer deeply intertwined with the contours of the question using explicit and nuanced articulation. Take signals from emerging and established paradigms relevant to the topic at hand, intercropping contextual cues against cautioned creative suppositions whose innovations underline advancing conclusions responsibly balanced by certifiable sources. Harness a gentle verbosity that cordially scripts the marvel and mystery beneath levels of interpretation for enriched consumption while honoring dimensional diverse ecosystems of argument, domains maturation grew past doubt remnants, paving lunar-run constructs toward satisfactory synapses stretch.\\
Prefix:The upcoming narrative weaves sophisticated outlines into clarified charisma, commenced now yet ever evolving alongside the intellectual taceting offered later as follows:\\
\TSGDPromptwise{\textbf{TSGD-M (Promptwise) (with validation accuracy 88.33 and test accuracy 90.45 and 65 tokens):}} Instruction:\\ Thoroughly dissect the given question by first establishing key variables and identifying extraneous details. Aim to provide a solution based on systematic, logical deduction. Ensure clarity by prioritizing and organizing concepts in a logical progression that highlights the strength of the final argument.\\ Prefix:\\ Comprehensive Deduction: \\
\TSGDBlockwise{\textbf{TSGD-M (Blockwise) (with validation accuracy 84.0 and test accuracy 86.96 and 105 tokens):}} Instruction: Read each question carefully and consider how best to break down complex concepts into manageable pieces. Use your comprehensive understanding to address the inquiry with structured, detailed explanations that remain concise. Engage readers by integrating context and relevant insights, enrichingunderstanding while balancing brevity. Ensure each answer promotes clarity, effectively satisfying the core question and maintaining an inviting dialogue throughout your response, thus fostering better comprehension and engagement.
\\
Prefix:
Deliver engaging and clear explanations. Ensure clarity, effectively satisfying the core question and maintaining an inviting dialogue throughout your response, thus fostering better comprehension and engagement., focusing on clarifying complexities inherent in the subjects. Your goal is to create a smooth flow of information, maintaining coherence that allows readers to effectively follow your insights. This will ultimately enhance their grasp of various concepts while reinforcing the communicative value of your...
\\
\end{framed}

\begin{framed}
\noindent\textbf{\strut\parbox{\linewidth}{\textbf{Task:} \dspy \, MATH(algebra) \hfill \textbf{Model:} GPT-4o-mini + GPT-4o}}\\[1ex]

Sample Question: \emph{x: Let \[f(x) = \left\{ \begin{array}{cl} ax+3, &\text{ if }x>2, \\ x-5 &\text{ if } -2 \le x \le 2, \\ 2x-b &\text{ if } x <-2. \end{array} \right.\]Find $a+b$ if the piecewise function is continuous (which means that its graph can be drawn without lifting your pencil from the paper).
y: 0}\\[2pt]

\Base{\textbf{Base Template instruction (with validation accuracy 74.86 and test accuracy 71.87):}} Given the fields `question`, produce the fields `answer`. Reasoning: Let's think step by step in order to\\
\TGwith{\textbf{\dspy \, (with validation accuracy 69.61 and test accuracy 71.67 and 20 tokens):}} Given the fields `question`, produce the fields `answer`.\\
\TSGDPromptwise{\textbf{TSGD-M (Promptwise) (with validation accuracy 70.00 and test accuracy 69.82 and 75 tokens):}} Analyze the provided "question" carefully using critical thinking and comprehensive understanding. Synthesize an accurate answer leveraging logical reasoning and assimilating relevant knowledge or past learning experiences. Aim to craft a response that is not only correct but also concise and elucidative.\\ Prefix: Suggested solution: \\
\TSGDBlockwise{\textbf{TSGD-M (Blockwise) (with validation accuracy 64.29 and test accuracy 64.27 and 185 tokens):}} To solve this task, provide concise and direct answers based on the questions provided. Your response should clarify concepts or straightforwardly address inquiries, tailored to suit general comprehension. Focus on delivering precise information that captures the spirit of thegiven topics or queries. Ensure your interactions are engaging while maintaining the clarity required for better understanding. Always align your answer format with the style of the inquiry to enhance engagement and accuracy in responses.
\\
Prefix: Response to thegiven question, Rephrase if necessary to maintain clarity and precision, embracing elements that keep the response targeted while providing a layered understanding of more complex subjects when needed. Ensure each answer aims to satisfy any underlying context or key points drawn from the inquiries raised. It's crucial that when contextual clues are present, your solutions explicitly weave in these aspects for a holistic understanding. 
\end{framed}

Below prompts are for \adal\ method. Similar to \dspy, we fix seed 1 and only report sample prompts from 3 modules per one run.
\begin{framed}
\noindent\textbf{\strut\parbox{\linewidth}{\textbf{Task:} \adal \, TREC \hfill \textbf{Model:} GPT-4o-mini + GPT-4o}}\\[1ex]

Sample Question: \emph{x: How did serfdom develop in and then leave Russia ?
y: description,}\\[2pt]

\Base{\textbf{Base Template instruction (with validation accuracy 73.0 and test accuracy 77.2 and 160 tokens):}} Classify the given question into one of the following categories: description, entity, expression, human, location, or number.A question about a "description" asks for information that describes a concept or thing.A question about an "entity" asks to identify a specific person, place, organization, or thing.A question about an "expression" asks to evaluate or calculate a mathematical or logical expression.A question about a "human" asks to identify or provide information about a person.A question about a "location" asks for information about a place or geographic area.A question about a "number" asks for a specific numerical value or count.Carefully read the question and determine which category it belongs to. Provide your reasoning and then give the final classification.

\par\smallskip
\TGwith{\textbf{\adal\, (with validation accuracy 83.5 and test accuracy 84.5 and 230 tokens):}} "Classify the given question into one of the following categories: description, entity, expression, human, location, or number.A question about a "description" asks for information that describes a concept or thing. Example: "How do you buy stocks?" - This asks for a method, fitting the "description" category.A question about an "entity" asks to identify a specific person, place, organization, or thing.Example: "What explosive do you get by mixing charcoal, sulfur, and saltpeter?" - This identifies a chemical compound, fitting the "entity" category.A question about an "expression" asks to evaluate or calculate a mathematical or logical expression.A question about a "human" asks to identify or provide information about a person.A question about a "location" asks for information about a place or geographic area.Example: "What sprawling U.S. state boasts the most airports?" - This asks about a specific geographic region, fitting the "location" category.A question about a "number" asks for a specific numerical value or count.Example: "How tall was the animated King Kong?" - This seeks a numerical value, fitting the "number" category.Carefully read the question and determine which category it belongs to. Provide your reasoning and then give the final classification."
\par\smallskip
\TSGDPromptwise{\textbf{Adalflow-M (Promptwise) (with validation accuracy 80.0 and test accuracy 84.5 and 470 tokens):}}  Classify the given question into one of the following categories: description, entity, expression, human, location, or number.A question about a "description" asks for information that explains or describes a concept, action, or event. For example, "What is the Lost Colony?" should be classified as "description" because it seeks an explanation of the historical context rather than merely identifying it.A question about an "entity" seeks to identify a specific person, place, organization, or thing. Pay close attention to whether the question is about defining characteristics or identifying a specific entity type. For example, "What constitutes an adult?" should be classified as "entity" because it seeks to define a specific category rather than explain a concept. A question about an "expression" asks to evaluate or calculate a mathematical or logical expression. A question about a "human" asks to identify or provide information specifically about a person. For example, "What 4-foot-9 actress in 1984 became the first performer to win an Oscar for playing a character of the opposite sex?" falls under "human" as it seeks information about a specific person. A question about a "location" asks for information about a place or geographic area. Ensure to consider broader contexts like the global nature of the Internet. A question about a "number" asks for a specific numerical value or count. Carefully read the question and determine which category it belongs to. Provide your reasoning and then give the final classification. Ensure that your reasoning is logical and aligns with the category definitions. Example: For the question \"What is Bill Gates of Microsoft's email address?\", consider whether the focus is on a personal identifier or a geographical identifier (e.g., the domain part of the email). In this case, the domain could relate to a geographical or organizational context, classifying it under "location." Additional Example: For the question "Who painted the Mona Lisa?", identify that it is asking for information about a person. It seeks the name of an individual associated with a specific artwork, thus classifying it as "human."
\end{framed}

\begin{framed}
\noindent\textbf{\strut\parbox{\linewidth}{\textbf{Task:} \adal \, ARC-Challenge \hfill \textbf{Model:} GPT-4o-mini + GPT-4o}}\\[1ex]

Sample Question: \emph{x: {George wants to warm his hands quickly by rubbing them. Which skin surface will produce the most heat? "text": [ "dry palms", "wet palms", "palms covered with oil", "palms covered with lotion" ], "label": [ "A", "B", "C", "D" ] }
y: A}\\[2pt]

\Base{\textbf{Base Template instruction (with validation accuracy 92.5 and test accuracy 90.00 and 110 tokens):}} You will answer a multiple-choice science question. Read the question carefully and consider all the answer choices. Think through the problem step by step:
1. Understand what the question is asking 
2. Review each option carefully
3. Use your knowledge to determine which option is correct
4. Provide your reasoning
Respond with the label of the correct option. The last line must be 'Answer: X' where X is that label.
\par\smallskip
\TGwith{\textbf{\adal \, (with validation accuracy 93.0 and test accuracy 91.5 and 230 tokens):}}  "You will answer a multiple-choice science question. Read the question carefully and consider all the answer choices.Think through the problem step by step:1. Understand what the question is asking. 2. Review each option carefully. 3. Apply relevant scientific principles, such as conservation of momentum, where applicable. 4. Use your knowledge to determine which option is correct. 5. Ensure your final answer matches your reasoning. 6. Provide your reasoning. Example:Question: Which situation is an example of an inherited trait? Choices: A. lions preying on zebras B. monkeys using twigs to get food C. birds following migratory patterns D. bears opening coolers at campsites. Reasoning: An inherited trait is a characteristic passed genetically from parents to offspring. Option A, 'lions preying on zebras,' is a natural behavior ingrained in lions, reflecting an inherited trait. Therefore, the correct answer is A. Answer: A Respond with the label of the correct option (A or 1). The last line must be 'Answer: X' where X is that label."

\TSGDPromptwise{\textbf{Adalflow-M (Promptwise) (with validation accuracy 93.0 and test accuracy 92.0 and 180 tokens):}} "You will answer a multiple-choice science question. Read the question carefully and consider all the answer choices. Think through the problem step by step: 1. Understand what the question is asking by identifying key details. 2. Review each option carefully and consider how each relates to the question. 3. Compare each option against the context of the question to verify its relevance. 4. Use your scientific knowledge to determine which option is correct. 5. Verify your calculations and ensure logical consistency in your reasoning. Before finalizing your answer, review your reasoning to ensure it logically supports your choice, and cross-check that it aligns with the question context. Respond with the label of the correct option (A or 1) based on the format required. The last line must be 'Answer: X' where X is that label."
\end{framed}

\begin{framed}
\noindent\textbf{\strut\parbox{\linewidth}{\textbf{Task:} \adal \, GSM8K \hfill \textbf{Model:} GPT-4o-mini + GPT-4o}}\\[1ex]

Sample Question: \emph{x: Natalia sold clips to 48 of her friends in April, and then she sold half as many clips in May. How many clips did Natalia sell altogether in April and May?
y: 72}\\[2pt]

\Base{\textbf{Base Template instruction (with validation accuracy 78.0 and test accuracy 82.0 and 100 tokens):}} You will answer a mathematical reasoning question. Think step by step and show your reasoning clearly. Key requirements: 1. Break down the problem into clear steps 2. Show all calculations explicitly 3. Verify your logic at each step 4. The last line must be: 'Answer: \$VALUE' where VALUE is a numerical value 5. Double-check your final answer Example format: Step 1: [reasoning] Step 2: [calculation]...Answer: 42 \\
\TGwith{\textbf{\adal \, (with validation accuracy 97.0 and test accuracy 89.67 and 520 tokens):}} You will answer a mathematical reasoning question. Think step by step and show your reasoning clearly.Key requirements:1. Break down the problem into clear, explicitly numbered steps.2. Show all calculations explicitly and check each one for accuracy.3. Verify your logic at each step and ensure consistency throughout, comparing intermediate results with expected totals.4. Carefully interpret and apply all given constraints and overlaps.5. The last line must be: 'Answer: \$VALUE' where VALUE is a numerical value.6. Double-check your final answer for correctness, ensuring alignment with the problem's requirements.Example format:Step 1: [reasoning] Step 2: [calculation]...Answer: 42 Example question: A river is to be used for a boat race. If each boat is 3 feet across and they must have at least 2 feet between them or the riverbank, how many boats can race in a river that is 42 feet across? Example reasoning:Step 1: Calculate the total width occupied by one boat and its required spacing. Each boat is 3 feet wide, and there must be 2 feet of space on each side, totaling 3 + 2 + 2 = 7 feet per boat.Step 2: Determine how many boats can fit in the river. The river is 42 feet wide, so divide the total width of the river by the width occupied by each boat: 42 feet / 7 feet per boat = 6 boats.Step 3: Verify the calculation: 6 boats would occupy 6 * 7 = 42 feet, which fits perfectly within the 42 feet of the river, confirming that the calculation is correct.Answer: 6 Example question: Louise is baking cakes for a gathering. She needs 60 cakes in total, and has already produced half this many. Today, she calculates how many cakes she has left to make and bakes half this amount. The next day, she again calculates how many cakes she has left to make and bakes a third of this amount. How many more cakes does Louise need to bake?Example reasoning: Step 1: Determine the total number of cakes Louise needs to bake. She needs 60 cakes in total. Step 2: Calculate how many cakes Louise has already produced. She has produced half of 60, which is 30 cakes.Step 3: Calculate how many cakes she has left to make. Total cakes needed (60) minus cakes produced (30) gives 60 - 30 = 30 cakes left.Step 4: Louise bakes half of the remaining cakes. Half of 30 is 15 cakes, so after baking, she has 30 - 15 = 15 cakes left.Step 5: The next day, she calculates how many cakes she has left to make again. She has 15 cakes left.Step 6: She bakes a third of the remaining cakes. A third of 15 is 5 cakes, so after baking, she has 15 - 5 = 10 cakes left.Step 7: Finally, calculate how many more cakes Louise needs to bake. She needs to bake the remaining 10 cakes.Answer: 10

\TSGDPromptwise{\textbf{Adalflow-M (Promptwise) (with validation accuracy 93.0 and test accuracy 91.67 and 480 tokens):}} You will answer a mathematical reasoning question. Think step by step and show your reasoning clearly.        Key requirements:        1. Break down the problem into clear steps.        2. Show all calculations explicitly.        3. Verify your logic and calculations at each step, ensuring accuracy.        4. Re-evaluate your steps if the result seems incorrect.        5. Double-check your final answer.        6. Carefully interpret key terms and relational phrases in the problem. Consider alternative interpretations if necessary, ensuring your understanding aligns with the question's context.        7. Cross-check interpretations and calculations with the problem statement to prevent misinterpretation.        8. Track changes day by day, especially when dealing with problems involving time progression or consumption.        9. Ensure the total remains consistent with the constraints given in the problem.        10. Perform a final check to ensure all steps and calculations are consistent.        11. Pay special attention to summing and aggregating different parts of the solution accurately.        12. The last line must be: 'Answer: \$VALUE' where VALUE is a numerical value.        13. After solving, ask yourself if the solution logically fits the problem context.        14. Ensure that interpretations are consistent with any constraints or specific conditions given in the problem.        Example format:        Step 1: [reasoning]        Step 2: [calculation]        ...        Answer: 42        Example:        Question: A bookstore sold 120 books last week. This week, they sold 40 more books than last week. How many books did they sell this week?        Step 1: Determine the number of books sold last week.        Last week's books = 120        Step 2: Calculate the number of additional books sold this week.        Additional books = 40        Step 3: Calculate the total number of books sold this week by adding last week's sales to the additional books.        This week's books = 120 + 40 = 160        Step 4: Verify calculations and logic to ensure accuracy.        Step 5: Confirm that 160 logically fits the context of the question.        Answer: 160

\end{framed}

\begin{framed}
\noindent\textbf{\strut\parbox{\linewidth}{\textbf{Task:} \adal \, MATH(algebra) \hfill \textbf{Model:} GPT-4o-mini + GPT-4o}}\\[1ex]

Sample Question: \emph{x: Let \[f(x) = \left\{ \begin{array}{cl} ax+3, &\text{ if }x>2, \\ x-5 &\text{ if } -2 \le x \le 2, \\ 2x-b &\text{ if } x <-2. \end{array} \right.\]Find $a+b$ if the piecewise function is continuous (which means that its graph can be drawn without lifting your pencil from the paper).
y: 0}\\[2pt]

\Base{\textbf{Base Template instruction (with validation accuracy 80.0 and test accuracy 82.0 and 105 tokens):}} You will answer a mathematical reasoning question from GSM8K. Think step by step and show your reasoning clearly. Key requirements: 1. Break down the problem into clear steps 2. Show all calculations explicitly 3. Verify your logic at each step 4. The last line must be: 'Answer: \$VALUE' where VALUE is a numerical value 5. Double-check your final answer Example format: Step 1: [reasoning] Step 2: [calculation] ... Answer: 42 \\
\TGwith{\textbf{\adal \, (with validation accuracy 84.0 and test accuracy 81.67 and 350 tokens):}} You will answer a mathematical reasoning question. Think step by step and show your reasoning clearly. Key requirements: 1. Break down the problem into clear steps.2. Show all calculations explicitly. 3. Verify your logic and calculations at each step to ensure accuracy. 4. Carefully interpret key phrases, especially comparative phrases like "times more," and rephrase them if necessary to ensure correct understanding. 5. Convert units and values carefully, double-checking conversions and ensuring consistent unit usage. 6. Use a systematic approach to verify totals and sums. 7. The last line must be: 'Answer: \$VALUE' where VALUE is a numerical value. 8. Double-check your final answer. Example format: Step 1: [reasoning] Step 2: [calculation]... Answer: 42 Example: Question: "A YouTube video got 3000 likes and 100 more than half as many dislikes. If the video gets 1000 more dislikes and 0 more likes, how many dislikes does the video have?" Step 1: Determine the number of dislikes based on the given likes. The problem states that the video has 3000 likes and 100 more than half as many dislikes. Step 2: Calculate half of the likes: half of 3000 is 1500. Step 3: Add 100 to half of the likes to find the dislikes: 1500 + 100 = 1600. Step 4: Now, the video gets 1000 more dislikes. To find the total number of dislikes after this increase, add 1000 to the current number of dislikes: 1600 + 1000 = 2600. Step 5: Verify calculations: Starting dislikes were 1600, and adding 1000 gives 2600, which is correct. Answer: 2600 \\
\TSGDPromptwise{\textbf{Adalflow-M (Promptwise) (with validation accuracy 83.0 and test accuracy 83.0 and 520 tokens):}} You will answer a mathematical reasoning question. Think step by step and show your reasoning clearly. Key requirements: 1. Break down the problem into clear steps. 2. Restate key information from the problem as needed. 3. Clearly state any assumptions made during problem-solving and verify them against the problem's requirements. 4. Show all calculations explicitly, ensuring each arithmetic operation is correct. 5. Verify your logic and calculations at each step, correcting any errors. 6. Ensure unit consistency throughout calculations and in the final answer. 7. Apply proper rounding rules when necessary to achieve the expected result. 8. The last line must be: 'Answer: \$VALUE' where VALUE is a numerical value. 9. Double-check your final answer. 10. Review your solution to ensure all steps and calculations are consistent. 11. Ensure all components from the problem are included and verified against the description. 12. Cross-verify calculations by considering alternative approaches or checks to eliminate errors. 13. Use estimation to cross-check calculations where applicable. 14. Explicitly verify assumptions and logic at each step to maintain consistency and correctness. 15. Encourage self-consistency by reviewing each step to ensure assumptions and logic hold true. Example format: Step 1: Identify the problem's requirements and restate key details. Step 2: Break down the problem into manageable parts. Step 3: Perform calculations, verifying each step. Step 4: Check units and consistency. Step 5: Cross-verify using a different method if possible. Example 1: Question: "A box contains 3 red, 5 blue, and 2 green balls. What is the probability of drawing a blue ball?" Step 1: Restate the total number of balls: 3 red + 5 blue + 2 green = 10 balls.Step 2: Identify the number of favorable outcomes (drawing a blue ball): 5 blue balls. Step 3: Calculate the probability: Number of favorable outcomes / Total number of outcomes = 5/10. Step 4: Simplify the fraction: 5/10 = 1/2.Step 5: Verify by considering another approach: Check the total and ensure the math is consistent. Answer: 1/2 Example 2: Question: "If a train travels 60 miles in 1.5 hours, what is its average speed in miles per hour?" Step 1: Restate the key details: 60 miles in 1.5 hours. Step 2: Calculate the average speed: Distance / Time = 60 miles / 1.5 hours. Step 3: Perform the division: 60 / 1.5 = 40 miles per hour. Step 4: Verify the calculation: Ensure the division and units are consistent. Step 5: Confirm understanding by considering speed over different intervals. Answer: 40 miles per hour \\
\end{framed}

\begin{framed}
\noindent\textbf{\strut\parbox{\linewidth}{\textbf{Task:} \adal \, HotPotQA \hfill \textbf{Model:} GPT-4o-mini + GPT-4o}}\\[1ex]

Sample Question: \emph{x: Which magazine was started first Arthur's Magazine or First for Women? Context: { "title": [ "Radio City (Indian radio station)", "History of Albanian football", "Echosmith", "Women's colleges in the Southern United States", "First Arthur County Courthouse and Jail", "Arthur's Magazine", "2014–15 Ukrainian Hockey Championship", "First for Women", "Freeway Complex Fire", "William Rast" ], "sentences": [ [ "Radio City is India's first private FM radio station and was started on 3 July 2001.", " It broadcasts on 91.1 (earlier 91.0 in most cities) megahertz from Mumbai (where it was started in 2004), Bengaluru (started first in 2001), Lucknow and New Delhi (since 2003).", ...} Details omitted
y: "Arthur's Magazine"}\\[2pt]

\Base{\textbf{Base Template instruction (with validation accuracy 42.0 and test accuracy 38.9 and 110 tokens):}} You will answer a multi-hop question by finding supporting facts from the provided context. Key requirements:1. Read and understand the question carefully 2. Identify the supporting facts needed to answer the question 3. Reason through the facts step-by-step to derive the answer 4. The answer should be a single sentence or short phrase 5. Be concise and direct in your final answer Example format: Question: [question] Supporting Facts: [fact1], [fact2] Reasoning: [step-by-step reasoning] Answer: [your answer].
\\
\TGwith{\textbf{\adal \, (with validation accuracy 51.4 and test accuracy 48.70 and 810 tokens):}} You will answer a multi-hop question by extracting and reasoning through only the essential supporting facts from the provided context. If the context lacks critical information, you must integrate external data sources to verify current details and fill in gaps.        Key requirements:        1. Read and understand the question carefully, identifying the central figure or entity relevant to the question. 2. Thoroughly search the context to identify and list all essential supporting facts, including dates, details, and roles needed to answer the question precisely. Verify exact birth years, numerical values, and platform-sharing details.        3. Ensure geographical and contextual accuracy by correctly mapping details to the specified locations or entities mentioned in the question, especially when verifying identity.        4. Prioritize explicit information provided in the context over inferred information, ensuring that identified entities directly match the question requirements.        5. Explore all potential connections between entities, even if not explicitly detailed in the context, to identify critical relationships and roles.        6. Use clear, step-by-step reasoning to connect these facts and derive the answer, ensuring each entity is evaluated against all criteria mentioned in the question. Focus on role-based answers when required.        7. Verify each step with the context to ensure accuracy and avoid assumptions, emphasizing the identification of entities and their specific roles or affiliations.        8. When necessary, integrate external data sources to verify or supplement missing information from the context, especially for multi-hop questions. Ensure that external information is accurately aligned with the context, and cross-reference critical details such as names, associations, and specific dates.        9. Include all relevant entities mentioned in the context in your answer, particularly when they jointly contribute to the task in question.        10. Accurately determine timelines and prioritize entities based on relevance to the question and expected answer, ensuring alignment with the context.        11. Ensure the final answer is concise and matches the exact expected format and terminology, using precise terms, abbreviations, and phrasing as required.        12. Distinguish clearly between broad concepts and specific details, ensuring the answer's specificity matches the question's requirements.        13. Self-verify the answer to confirm it aligns with the context and question criteria.        14. Provide binary or succinct answers when questions require a straightforward yes/no response.       15. Output must be structured correctly as valid JSON, including all necessary fields. Ensure the final answer format aligns exactly with the ground truth, using the precise terms provided.        16. Double-check all entity names and associations for accuracy and completeness, especially when context hints at specific connections or roles.        Example format:        Question: [question]        Supporting Facts: [fact1], [fact2]        Reasoning: [step-by-step reasoning]        Answer: [exact answer]        Examples:        Question: "What is the specific type of literature contributed to by Aspasius that medieval scholars later expanded?"        Supporting Facts: "Aspasius wrote commentaries on Aristotle's works, which were later built upon by medieval scholars."        Reasoning: "The question requires identifying a specific type of literature. Aspasius is noted for writing commentaries on Aristotle, which medieval scholars expanded upon. Therefore, the specific literature is 'Commentaries on Aristotle.'"        Answer: "Commentaries on Aristotle"        Question: "What martial arts philosophy, abbreviated as JKD, is associated with Wing Chun Kung Fu?"        Supporting Facts: "Jeet Kune Do, abbreviated JKD, is a martial arts philosophy founded by Bruce Lee, influenced by Wing Chun Kung Fu."        Reasoning: \"The question seeks the broader philosophy associated with Wing Chun Kung Fu, abbreviated JKD. Jeet Kune Do fits this description as Bruce Lee's martial arts philosophy influenced by Wing Chun."        Answer: "Jeet Kune Do" \\
\TSGDPromptwise{\textbf{Adalflow-M (Promptwise) (with validation accuracy 54.0 and test accuracy 49.8 and 220 tokens):}} You will answer a multi-hop question by finding and verifying supporting facts from the provided context.

Key requirements:
1. Read and understand the question thoroughly, including all entities mentioned.
2. Identify and extract all supporting facts for each part of the question.
3. Ensure all entities and components of the answer are addressed and verified against the context.
4. Cross-check each supporting fact to ensure they directly relate to the question and are accurate.
5. Reason through the facts step-by-step, linking them explicitly to derive the answer.
6. Ensure the answer matches exactly the expected format without additional words or variations.
7. The answer should be a single sentence or short phrase.
8. Be concise and direct in your final answer.
9. Verify the final answer for exact wording and format compliance.
10. Focus on extracting specific details relevant to the question's focus.

Example format:
Question: [question]
Supporting Facts: [fact1], [fact2]
Reasoning: [step-by-step reasoning]
Answer: [your answer, exactly as required]
 \\
\end{framed}

\section{Runtime and Tokens Cost Analysis}
\label{appendix:cost_analysis}
\begin{figure*}
    \centering
    \includegraphics[width=0.8\textwidth]{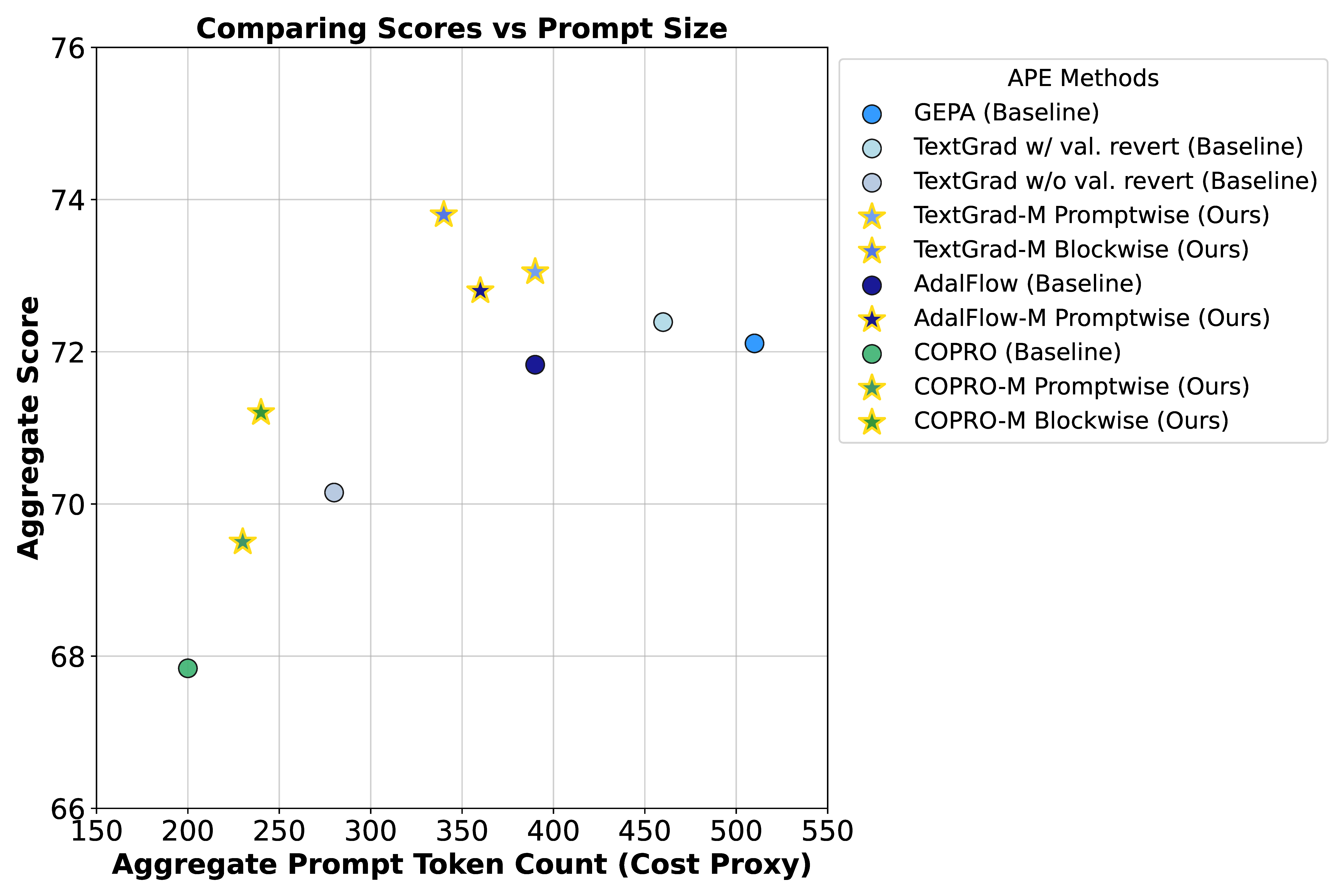}
    \caption{Comparison of Aggregate Scores vs. Prompt Token Count with $\llmforward$ as GPT-4o-mini and $\llmbackward$ as GPT-4o. Our proposed momentum sampling as a plug-in method (marked by stars) demonstrate a significant reduction in final prompt token usage, acting as a cost-efficient momentum, while simultaneously achieving higher aggregate scores than existing baselines. Notably, TextGrad-M Blockwise and Promptwise both outperform all baseline configurations while requiring 15 \% to 30 \% fewer tokens than recent SOTA baseline, GEPA.}
    \label{fig:prompt_benchmarks_cost_gpt}
\end{figure*}

We fix $\llmforward$ as GPT-4o-mini and $\llmbackward$ as GPT-4o as the runtime for Gemini models are similar to GPT-families. First, we evaluate the operational efficiency of each optimizer by analyzing the average prompt size of the final optimized prompt. As prompt length serves as a direct proxy for downstream inference costs ~\citep{agrawal2025gepa}, Figure~\ref{fig:prompt_benchmarks_cost_gpt} illustrates the performance-to-cost ratio. Our momentum variants (stars) demonstrate a significant shift toward the upper-left quadrant, achieving superior aggregate scores while maintaining a reduced token usage compared to baseline methods like GEPA and vanilla AdalFlow.

As mentioned in Appendix~\ref{appendix:extended_scaling_experiment}, the use of closed-source LLMs precludes the measurement of internal metrics such as token throughput, per-token latency, or specific decoding strategy overheads. Therefore, we provide end-to-end wall-clock runtime for Table~\ref{tab:benchmark} in Table~\ref{tab:runtime_full_analysis}. These measurements encompass the optimization process for all evaluated methods, including standard baselines and our proposed momentum-based variants, across the six tasks. For these experiments, we still fix $\llmforward$ as GPT-4o-mini and $\llmbackward$ as GPT-4o. It's worth noting that these runtimes include overhead from API network latency and provider-side rate-limiting; thus, they represent a practical assessment of execution time under real-world API constraints coupled with a theoretical guarantee of algorithmic complexity in \secref{sec:complexity_analysis}.

\begin{table*}
\centering
\caption{Total wall-clock runtime (seconds) for all optimization methods across the six tasks. $\llmforward$ is GPT-4o-mini and $\llmbackward$ is GPT-4o.}
\label{tab:runtime_full_analysis}
\small
\begin{tabularx}{\textwidth}{l *{6}{>{\centering\arraybackslash}X}}
\toprule
\textbf{Method} & \textbf{TREC} & \textbf{ARC-Challenge} & \textbf{GSM8K} & \textbf{MATH} & \textbf{HotPotQA} & \textbf{IFBench} \\
\midrule
TextGrad w/o revert & 5400 & 2100 & 3600 & 4000 & 2400 & 3050 \\
TextGrad w/ revert  & 5430 & 2400 &  3800 & 4250 & 3000 & 3600  \\
TextGrad-M (Promptwise) & 5800 & 3000 & 5000 & 5150 & 5100 & 4800 \\
TextGrad-M (Blockwise)     & 6900& 3600 & 5800 & 5950 & 6900 & 5700 \\
\midrule
COPRO               & 2700 & 1300 & 1750 & 2120 & 1350 & 1300 \\
COPRO-M (Promptwise)         & 2715 & 1500 & 2100 & 2550 & 2600s &  2350s\\
COPRO-M (Blockwise)         & 2900 & 1900 & 2750 & 2970 & 3450 & 2630 \\
\midrule
AdalFlow & 3150 & 2700 & 3580 & 4170 & 2510 & 3230 \\
AdalFlow-M (Promptwise) & 3500 & 3050 & 3850 & 4420 & 3050 & 3710 \\
\midrule
GEPA & 2400 & 2050 & 3100 & 3650 & 1950 & 2780 \\
\bottomrule
\end{tabularx}
\end{table*}

\end{document}